\documentclass[preprint,12pt]{elsarticle}

\usepackage{amssymb}
\usepackage{amsmath}

\usepackage{algorithm}
\usepackage[rightComments=true]{algpseudocodex}

\usepackage{array}           
\usepackage{booktabs}        
\usepackage{makecell}        
\usepackage{multirow}        

\usepackage{tabularx}        
\usepackage{longtable}       
\usepackage{threeparttable}  
\usepackage{arydshln}        

\usepackage{caption}
\usepackage{subcaption}

\newcommand{\norm}[1]{\left\Vert#1\right\Vert}
\newcommand{\abs}[1]{\left\vert#1\right\vert}

\newtheorem{theorem}{Theorem}[section]

\usepackage{hyperref}
\hypersetup{
    colorlinks=true,
    linkcolor=blue,
    urlcolor=blue,
    citecolor=blue}
\hypersetup{pdfauthor={Name}}

\begin{document}

\begin{frontmatter}



    \title{Benchmarking Continuous Dynamic Multi-Objective Optimization: Survey and Generalized Test Suite}

    \author[label_UTS,label_SUSTech]{Chang Shao}
    \author[label_SUSTech]{Qi Zhao}
    \author[label_CDTU]{Nana Pu}
    \author[label_SNNU]{Shi Cheng}
    \author[label_UTS]{Jing Jiang\corref{cor1}}
    \author[label_SUSTech]{Yuhui Shi\corref{cor1}}

    \affiliation[label_UTS]{organization={Australian Artificial Intelligence Institute (AAII), Faculty of Engineering and Information Technology, University of Technology Sydney},
        addressline={15 Broadway, Ultimo},
        city={Sydney},
        postcode={2007},
        state={NSW},
        country={Australia}
    }

    \affiliation[label_CDTU]{organization={School of Big Data and Artificial Intelligence, Chengdu Technological University},
        city={Chengdu},
        postcode={611730},
        state={Sichuan},
        country={China}
    }

    \affiliation[label_SNNU]{organization={School of Artificial Intelligence and Computer Science, Shaanxi Normal University},
        city={Xi’an},
        postcode={710119},
        state={Shaanxi},
        country={China}
    }

    \affiliation[label_SUSTech]{organization={Department of Computer Science and Engineering, Southern University of Science and Technology},
        city={Shenzhen},
        postcode={518055},
        state={Guangdong},
        country={China}
    }

    \cortext[cor1]{Corresponding authors.}
    \ead{shiyh@sustech.edu.cn}

    \begin{abstract}
        The field of Dynamic Multi-Objective Optimization (DMOO) has witnessed a surge of interest from both academia and industry, as numerous time-evolving real-world applications can be naturally formulated as Dynamic Multi-Objective Optimization Problems (DMOPs). This growing demand thus necessitates advanced benchmarks to rigorously evaluate optimization algorithms under realistic conditions. This paper introduces a comprehensive and principled framework for constructing highly realistic and challenging DMOO benchmarks. The proposed framework incorporates several novel components, including: a generalized formulation that allows the Pareto-optimal Set (PS) to change on hypersurfaces; a mechanism for creating controlled variable contribution imbalances to generate heterogeneous landscapes; and dynamic rotation matrices for inducing time-varying variable interactions and non-separability. Furthermore, we incorporate a temporal perturbation mechanism to simulate irregular environmental changes and propose a generalized time-linkage mechanism that systematically embeds historical solution quality into future problems, thereby capturing critical real-world phenomena such as error accumulation and time-deception. Extensive experimental results validate the effectiveness of the proposed framework, demonstrating its superiority over conventional benchmarks in terms of realism, complexity, and its capability for discriminating state-of-the-art algorithmic performance. Thus, this work establishes a new standard for dynamic multi-objective optimization benchmarking and provides a powerful tool for the development and evaluation of next-generation algorithms capable of addressing the complexities of real-world dynamic systems.
    \end{abstract}




\end{frontmatter}



\section{Introduction}

Many practical problems can be modeled as continuous dynamic multi-objective optimization problems (DMOPs), such as
scheduling~\cite{J_KBS_shen2025reinforcement, J_SWEVO_yang2025deep, J_TCC_xie2024transfer, J_TITS_eaton2017ant}, planning~\cite{J_TPDS_shovan2025parallel, J_SWEVO_herring2024comparative, J_JSYST_ghasemkhani2017optimal}, design~\cite{C_CEC_martins2009dynamic, J_TMAG_dibarba2008dynamic}, control~\cite{J_EAAI_liu2025dynamic, J_ACCESS_ma2025power}, resource allocation~\cite{J_SWEVO_dong2025adaptive,C_SECON_palaniappan2001dynamic}, etc. These problems have multiple (two or three) conflicting objectives, where the objective functions, constraints, and even the number of decision variables or objectives themselves~\cite{J_TEVC_ruan2025coping, J_TEVC_chen2018dynamic} are dynamic and evolve over time.

Dynamic multi-objective optimization (DMOO) has witnessed significant progress over the past two decades, attracting growing interest from both academic researchers and engineering practitioners. The development of dynamic multi-objective optimization algorithms (DMOAs) has progressed from simpler mechanisms, such as diversity introduction~\cite{C_CEC_liu2014adaptive, J_TEVC_goh2009competitivecooperative, C_EMO_deb2007dynamic}, diversity maintenance mechanisms~\cite{J_TCYB_hu2023handling, C_EMO_deb2007dynamic}, and memory-based approaches~\cite{C_CEC_sahmoud2020memoryassisted, C_CEC_wang2009investigation} for periodic or quasi-periodic environments, toward the current mainstream paradigm: prediction-based methods. These methods initially employed classical time-series forecasting methods, such as forward-looking approach~\cite{C_GECCO_hatzakis2006dynamic}, Kalman filter~\cite{J_TCYB_muruganantham2016evolutionary, J_PROCS_muruganantham2013dynamic}, regression~\cite{J_TCYB_zhou2014population}, support vector machine~\cite{C_CEC_hu2019solving} or support vector regression~\cite{J_TEVC_cao2020evolutionary}, later incorporating deep learning techniques like transfer learning~\cite{J_TCYB_jiang2021individualbased, J_TCYB_jiang2021fast, J_TEVC_jiang2018transfer}, recurrent neural network (RNN)~\cite{J_TEVC_hu2025dynamica}, long short-term memory (LSTM)~\cite{J_TEVC_hu2025dynamic, J_ESWA_xu2025prediction, C_CEC_rang2023long}, ensemble learning~\cite{J_ASOC_wang2020ensemble}, manifold learning~\cite{J_TCYB_jiang2021fast}, reinforcement learning (RL)~\cite{J_SWEVO_yan2025dualpopulation, J_ISCI_zou2021reinforcement, J_ISCI_zou2021dynamic}, and so on. Another important category of DMOAs is knowledge-based methods, which mine useful information from accumulated historical knowledge to accurately predict changing patterns in Pareto-optimal Set (PS) or Pareto-optimal Front (PF)~\cite{J_TEVC_zou2025knowledge, J_SWEVO_xie2025weighted, J_KBS_ye2022knowledge}.

It is noteworthy that, in essence, any multivariate time series forecasting model can be integrated with multi-objective optimization algorithms (MOAs) to solve DMOPs. However, a critical prerequisite is that the time required for the optimization process incorporating these forecasting algorithms should not exceed that of the base algorithm to obtain solutions of comparable quality. Existing experimental results~\cite{J_TEVC_jiang2025dynamic} demonstrated that the prediction process in~\cite{J_TEVC_cao2020evolutionary} dominated the majority of the total optimization time (requiring 100 minutes per run, compared to less than 2 minutes for the base algorithm). The primary purpose of incorporating machine learning-based forecasting methods is to achieve faster predictions, thereby enabling rapid response to changes and resource conservation. Moreover, the design of such integrated approaches must be logically sound and avoid introducing bias towards specific types of dynamics. In the literature, some existing methods~\cite{C_NCAA_chen2023lstm, C_CEC_rang2023long} first use classical DMOAs to obtain a set of solutions, then spend additional time training a model for subsequent direct use. A key issue with this approach is that, in practice, it is difficult to determine whether the environment changes slowly enough to allow sufficient time for model training. The environment may change before model training is completed. Therefore, when a forecasting model is incorporated, the time required for both prediction and model training must be considered as an essential evaluation metric.

With the advent of large language models (LLMs), their integration into DMOO is becoming increasingly feasible. This trend is supported by a growing body of literature that focuses on time‑series forecasting by leveraging the capabilities of LLMs, such as~\cite{C_ICLR_hu2025contextalignment, C_AAAI_liu2025timecma, C_ICLR_jin2024timellm, C_IJCAI_zhang2024large}. However, a critical challenge is to ensure that the computational cost of prediction does not dominate the optimization process. Additionally, the development of benchmark problems is essential for the fair comparison and advancement of next-generation DMOAs. Currently, most existing test suites, such as the DF benchmark~\cite{R_jiang2018benchmark} (designed in 2018) and FDA~\cite{J_TEVC_farina2004dynamic} (designed in 2004), are widely used to evaluate newly developed DMOAs due to their simplicity of structure and mathematical tractability. However, these benchmarks exhibit the property that PS changes on a hyperplane, introducing a potential bias that may favor certain algorithms (see Section~\ref{sec:gts_friedman_rank_test} for more details) and limit their generalizability to real-world scenarios.

To address this limitation, this article conducts a comprehensive review of existing benchmarks, analyzes their limitations, and introduces a novel configurable test suite (the Generalized Test Suite), which provides a more challenging and unbiased testbed for rigorous evaluation of DMOAs. Our main contributions are summarized as follows:

\begin{itemize}
    \item[1)] We propose a simple method to construct diverse PS dynamics on hypersurfaces. Such dynamics are more general than those confined to hyperplanes, as commonly seen in existing benchmarks, and thus provide a more effective means of evaluating the tracking ability of DMOP solvers in dynamic environments.
    \item[2)] We introduce controlled imbalances in variable contributions via diagonal adjustments of symmetric positive semidefinite matrices, and incorporate dynamic rotation matrices to induce partly-separable interactions, capturing realistic complexity.
    \item[3)] We introduce temporally irregular dynamics using pseudo-random perturbations based on the digits of $\pi$, capturing the unpredictable nature of real-world systems and guaranteeing numerical experimental repeatability.
    \item[4)] We mathematically formulate the time-linkage property in DMOPs to investigate solver robustness. The proposed formulation simultaneously emulates the time-deception and error accumulation characteristics of time-linkage DMOPs.
\end{itemize}

The rest of the paper is organized as follows: Section~\ref{sec:gts_survey_benchmarks} reviews existing DMOO benchmarks; Section~\ref{sec:gts_gts_philosophy} details the design philosophy of the proposed benchmark; Section~\ref{sec:gts_gts_instances} presents specific problem instances; Section~\ref{sec:gts_experiments} conducts comparative experiments; and Section~\ref{sec:gts_conclusion} concludes the work.

\section{Survey on DMOPs Benchmarks}%
\label{sec:gts_survey_benchmarks}

Suitable benchmark problems are crucial for evaluating the performance of DMOP solvers. Studies on DMOO benchmarks can be divided into four categories. Early work introduces dynamics to some well-known static multi-objective optimization problems. For example, Farina et al. developed the FDA test suite~\cite{J_TEVC_farina2004dynamic} based on the static ZDT~\cite{J_ECJ_zitzler2000comparison} and DTLZ~\cite{C_CEC_deb2002scalable} problems. Jin and Sendhoff~\cite{C_EvoWS_jin2004constructing} aggregated multiple static objectives and simulated dynamics by changing the weight of each objective. The ZJZ~\cite{C_EMO_zhou2007predictionbased} benchmark improved the FDA by considering nonlinear correlation between decision variables and Mehnen~\cite{R_mehnen2006evolutionary} proposed a new generic scheme DTF, which is a generalized FDA function.

To diversify problem properties, subsequent studies consider special and difficult test scenarios. For instance, the cases with time-varying numbers of decision variables and objectives were investigated in~\cite{J_ISCI_huang2011dynamic}. The test suites in~\cite{C_CIDUE_helbig2013benchmarks, J_CSUR_helbig2014benchmarks, J_TCYB_zhou2014population, C_CEC_biswas2014evolutionary} simulated various complicated PS and disconnected PF. A common feature of these benchmarks is that each dimension of the PS has a separate geometry. This kind of PS geometry is not intuitive and makes the static problem difficult to solve.

To emphasize the effect of dynamics on DMOO benchmarks, in the third category, the test suites simplify the static form of problems and focus on emulating different PF features. Gee et al.~\cite{J_TCYB_gee2017benchmark} suggested two forms of DMOO benchmarks, i.e., the additive form and multiplicative form. PFs of their problems have diverse characteristics, e.g., modality, connectedness, and degeneracy. Jiang et al. proposed the JY~\cite{J_TCYB_jiang2017evolutionary}, DF~\cite{R_jiang2018benchmark}, and SDP~\cite{J_TCYB_jiang2020scalable} test suites by combining the merits of previous benchmarks. These test suites mainly focus on PF properties, such as connectivity, degeneration, PF shrinkage/expansion, search favorability, etc. The DF~\cite{R_jiang2018benchmark} and SDP~\cite{J_TCYB_jiang2020scalable} benchmarks described various PS/PF characteristics and have gained great popularity in studies~\cite{J_TEVC_rong2020multimodel, J_TEVC_zhang2020novel, J_TEVC_cao2020evolutionary, J_TCYB_rambabu2020mixtureofexperts}.

Recently, RDF~\cite{J_TEVC_zou2025knowledge} and RDP~\cite{J_SWEVO_ruan2021random} leveraged the DF~\cite{R_jiang2018benchmark} and FDA~\cite{J_TEVC_farina2004dynamic} test suites to model irregular changes. A common technique involves generating a predefined random sequence $Q \in [0, l]$ by shuffling the integers from $0$ to $l$ using a fixed random number generator seed (e.g., rng(42) in Matlab). It is worth noting that this implementation confines the sequence to a specific computational environment and random seed. Consequently, results may become non-reproducible if the experiment is replicated on a different platform or with a different seed, potentially affecting the fairness of comparative studies. The work in~\cite{J_ASOC_sun2024scalable} primarily focuses on the scalability of test problems, with a specific aim of designing DMOPs.

\section{Design Principles for Benchmarking}%
\label{sec:gts_gts_philosophy}

To ensure compatibility, we retain the widely-used formulation as the base framework of DMOPs, which is given as follows~\cite{J_TEVC_farina2004dynamic}:
\begin{equation}\label{eq:gts_dmop}
    \begin{aligned}
        \min \ \mathbf{f}(\mathbf{x},t) = & \left( f_1(\mathbf{x},t), f_2(\mathbf{x},t), \cdots, f_M(\mathbf{x},t) \right)^T, \\
                                          & \mathrm{s.t.} \ \mathbf{x} \in \Omega
    \end{aligned}
\end{equation}
with
\begin{equation}\label{eq:gts_func_f}
    f_{i=1:M}(\mathbf{x},t) = g(\mathbf{x},t)\mu_i(\mathbf{x},t).
\end{equation}

In Equation~\eqref{eq:gts_dmop}, $\mathbf{x}=(x_1,x_2,\ldots,x_D)^T$ is a candidate solution, $\mathbf{f}:\Omega\rightarrow\mathbb{R}^M$ contains $M$ objective functions to be minimized, $\Omega\subseteq\mathbb{R}^D$ is the decision space, $\mathbb{R}^M$ is the objective space, and $t$ is a discrete time parameter controlling dynamics. In Equation~\eqref{eq:gts_func_f}, commonly $g(\mathbf{x}, t)>0$, the solution $(\mathbf{x}^*, t)$ to minimize $g(\mathbf{x}, t)$, characterizes PS properties, e.g., line, curve and surface; the PF is obtained through $\mu(\mathbf{x}^*, t) = \min (\mu_1(\mathbf{x}, t), \cdots, \mu_M(\mathbf{x}, t))$, and $\mu(\mathbf{x}^*,t)$ describes PF geometric shapes, i.e., convexity and concavity. $\mathbf{x}=(x_1,x_2,\ldots,x_D)^T$ is usually divided into two sub-vectors, i.e., $\mathbf{x}_I = (x_1, \cdots, x_m)^T$, ${M-1} \leq m < D$, for $\mu (\mathbf{x}, t)$ and $ \mathbf{x}_{II} = (x_{m+1}, \cdots, x_D)^T$ for $g(\mathbf{x}, t)$. These sub-vectors are used for separately designing the PS and PF, respectively. Following~\cite{J_TEVC_farina2004dynamic}, $t$ controls the dynamics via
\begin{equation}\label{eq:gts_time_regular}
    t=\frac{1}{n_t} \left\lfloor \frac{\tau}{\tau_t} \right\rfloor, \ \tau = 0, 1, 2, \dots
\end{equation}
where $n_t$ governs the severity of change, $\tau$ is the generation counter, and $\tau_t$ is the number of generations available at each time.

In this section, we present the design philosophy of our framework by focusing on the $g(\mathbf{x}, t)$ function, which enables the construction of benchmarks with configurable properties. The details are as follows:
(1) General dynamic components – to simulate fundamental changes in the decision or objective space over time;
(2) Imbalanced variable components – to create heterogeneous contributions across decision variables;
(3) Variable‑interaction components – implemented through time-varying rotation matrices, introduce non-separable or partly-separable correlations among variables;
(4) Irregular dynamic components – to emulate non-smooth and abrupt changes in the problem landscape;
(5) Time‑linkage components – in which future landscape variations depend on historical solutions, thereby challenging algorithms to track the evolving PS/PF accurately.

The modular design of these components enables flexible integration and the systematic generation of test instances with configurable complexity and a wide spectrum of dynamic features.

\subsection{Components with General Dynamics}

The $g(\mathbf{x}, t)$ function in Equation~\eqref{eq:gts_func_f} is the key component for constructing the PS/PF shapes and understanding the PS/PF changing patterns. To emulate general PS changing patterns, we divide $\mathbf{x}_{II}$ into $P$ parts $\mathbf{x}_{II,p}$, where $p \in \{1,2,\ldots,P\}$, and $1 \leq P \leq {D - M}$.

Obviously, $P=1$ and $P = D - M$ represent two extreme cases. The former has been widely used in the FDA~\cite{J_TEVC_farina2004dynamic}, ZJZ~\cite{C_EMO_zhou2007predictionbased}, dMOPs~\cite{J_TEVC_goh2009competitivecooperative}, DF~\cite{R_jiang2018benchmark}, JY~\cite{J_TCYB_jiang2017evolutionary} and SDP~\cite{J_TCYB_jiang2020scalable, J_TCYB_gee2017benchmark} test suites. In such cases, all $x_j$ in $\mathbf{x}_{II}$ are the same, resulting in PS varying on a static hyperplane, which is not always the case in practice. The latter has been considered in~\cite{J_ISCI_huang2011dynamic, C_CIDUE_helbig2013benchmarks, J_CSUR_helbig2014benchmarks, J_TCYB_zhou2014population, C_CEC_biswas2014evolutionary}. In such cases, each dimension of the PS holds a separate geometry, which makes the static problem difficult to solve. As a result, it would be unclear whether algorithms' performance on these benchmarks is attributed to the dynamics or the static properties of problems.

To address the above drawbacks and to achieve a balance between the generality of dynamics and the complexity of static problem-solving, we propose a generalized $g(\mathbf{x}, t)$ function with $1<P<{D - M}$. In this paper, we present the $g(\mathbf{x}, t)$ with $P = 2$ to construct the new benchmarks, as shown in Equation~\eqref{eq:gts_func_g}. The benchmark problems can be easily extended to any $2<P<{D - M}$ cases following our proposal.

\begin{equation}\label{eq:gts_func_g}
    \begin{split}
        g(\mathbf{x},t) = 1
         & + \Bigl(\bigl(\mathbf{x}_{II,1} - h_1(\mathbf{x}_I)\bigr)^T \mathbf{R}_{II,1}(t) \bigl(\mathbf{x}_{II,1} - h_1(\mathbf{x}_I)\bigr)\Bigr)^{\frac{1}{p}} \\
         & + \Bigl(\bigl(\mathbf{x}_{II,2} - h_2(\mathbf{x}_I)\bigr)^T \mathbf{R}_{II,2}(t) \bigl(\mathbf{x}_{II,2} - h_2(\mathbf{x}_I)\bigr)\Bigr)^{\frac{1}{p}}
    \end{split}
\end{equation}
where $p\geq 1$, $h_1(\mathbf{x}_I, t)$, $h_2(\mathbf{x}_I, t)$ will be explicitly provided for different problems, and $\mathbf{R}_{II,1}(t)$ and $\mathbf{R}_{II,2}(t)$ are positive semidefinite matrices in the $t$-th environment.

\begin{figure}[!t]
    \centering
    \begin{minipage}[t]{0.4\textwidth}
        \centering
        \includegraphics[width=\linewidth]{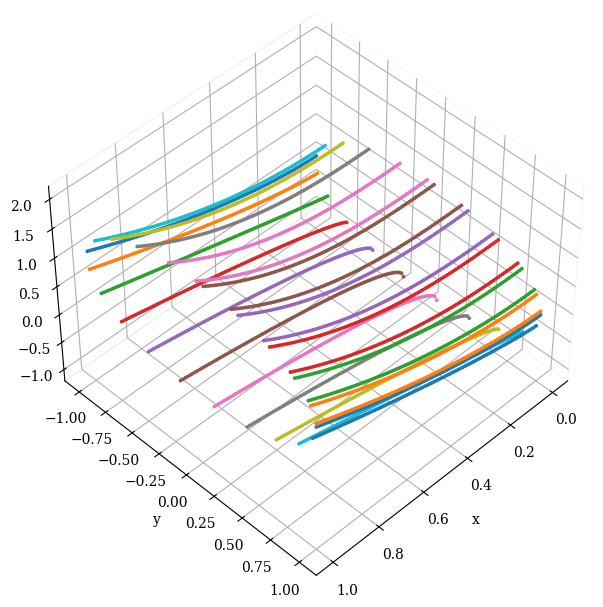}
        \caption{PS of GTS1}%
        \label{fig:gts_GTS1_ps}
    \end{minipage}
    \hfill
    \begin{minipage}[t]{0.4\textwidth}
        \centering
        \includegraphics[width=\linewidth]{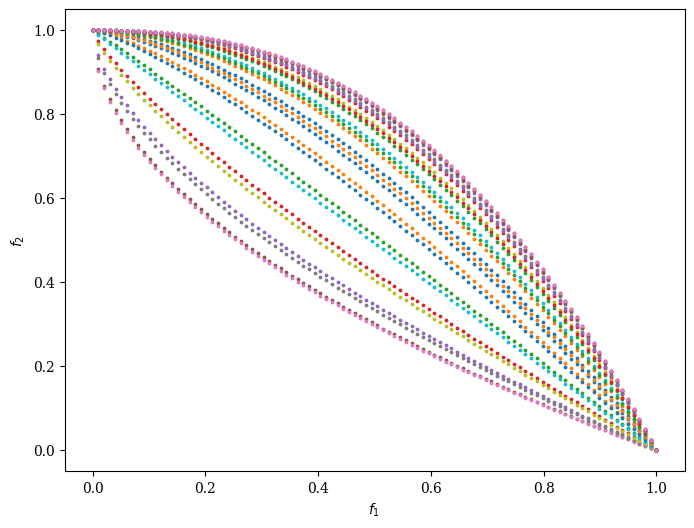}
        \caption{PF of GTS1}%
        \label{fig:gts_GTS1_pf}
    \end{minipage}
\end{figure}

\begin{figure}[!t]
    \centering
    \begin{minipage}[t]{0.4\textwidth}
        \centering
        \includegraphics[width=\linewidth]{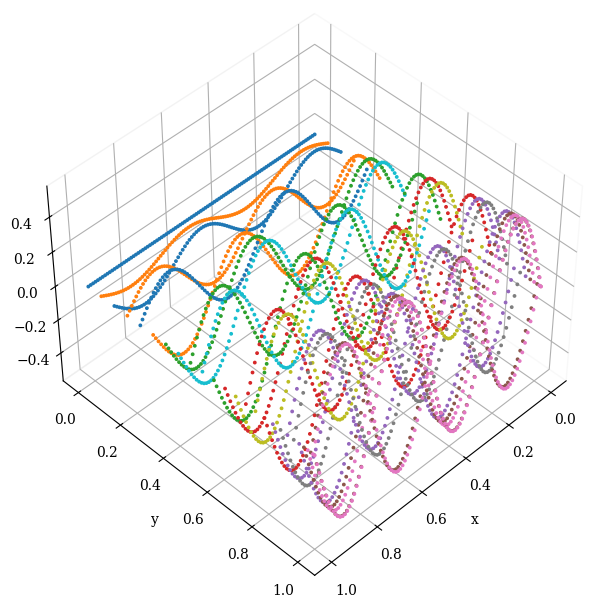}
        \caption{PS of GTS4}%
        \label{fig:gts_GTS4_ps}
    \end{minipage}
    \hfill
    \begin{minipage}[t]{0.4\textwidth}
        \centering
        \includegraphics[width=\linewidth]{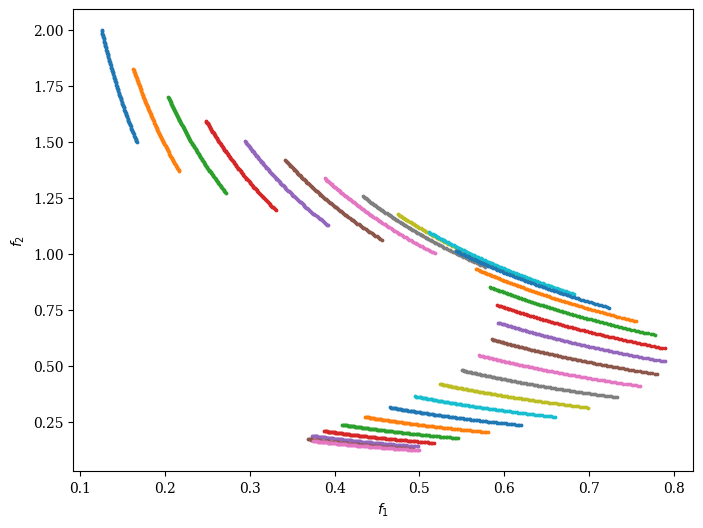}
        \caption{PF of GTS4}%
        \label{fig:gts_GTS4_pf}
    \end{minipage}
\end{figure}

The proposed component function~\eqref{eq:gts_func_g} emulates the dynamics of PS over hypersurfaces through controlling $h_1(\mathbf{x}_I, t)$ and $h_2(\mathbf{x}_I, t)$ as well as $\mathbf{R}_{II,1}(t)$ and $\mathbf{R}_{II,2}(t)$. These dynamics are more general cases with respect to the dynamics on hyperplanes. Figures~\ref{fig:gts_GTS1_ps}, and~\ref{fig:gts_GTS4_ps} show two representative PS dynamics in our proposed benchmarks. Furthermore, variables in $\mathbf{x}_{II,1}$ and $\mathbf{x}_{II,2}$ are independent of each other, reducing the difficulty of static problem-solving. Therefore, the proposed benchmarks can focus on evaluating the performance of algorithms on dynamic-handling.

\subsection{Components with Imbalanced Variable Contributions}

In most of the existing benchmark functions~\cite{R_jiang2018benchmark,J_TCYB_jiang2017evolutionary,J_TCYB_jiang2020scalable}, $\mathbf{R}_{II,1}(t)$ and $\mathbf{R}_{II,2}(t)$ in Equation~\eqref{eq:gts_func_g} are set as identity matrices $\mathbf{I}$. This conventional approach results in all variables having equal influence on the objective function values, thereby creating a symmetric fitness landscape. While such symmetry offers mathematical convenience and tractability, it often fails to capture the heterogeneous parameter sensitivities that are characteristic of real-world optimization problems.

To address this limitation, we propose a systematic methodology for introducing controlled imbalances into the benchmark functions through the adjustment of the diagonal elements of $\mathbf{R}_{II,1}(t)$ and $\mathbf{R}_{II,2}(t)$ matrices, which have the following form:
\begin{equation}
    \mathbf{R}_{II}(t) = \begin{pmatrix}
        r_1(t) & 0      & \cdots & 0      \\
        0      & r_2(t) & \cdots & 0      \\
        \vdots & \vdots & \ddots & \vdots \\
        0      & 0      & \cdots & r_n(t)
    \end{pmatrix}, \ r_i(t) > 0, \ i = 1, \cdots, n
\end{equation}
This approach enables the creation of more realistic test functions, where certain variables can exert a stronger influence on the objective function values than others. For instance, in the optimal control of a biological network~\cite{J_BMCSB_nimmegeers2016dynamic}, which is formulated as a dynamic multi-objective problem, variables are defined as $(x, \mu, \theta, t)$, representing the state, control, uncertain parameter vectors, and time, respectively. Here, the state ($x$) and control ($\mu$) variables generally have a more dominant influence on system behavior and objectives than the uncertain parameters ($\theta$).

The magnitude of these diagonal elements in $\mathbf{R}_{II,1}(t)$ and $\mathbf{R}_{II,2}(t)$ plays a crucial role in determining the contribution of each decision variable $x_i$ to the overall objective function landscape. By tuning the diagonal entries, we can precisely control the imbalance ratio $\rho(t) = \frac{\max(r_i(t))}{\min(r_i(t))}$ for $\mathbf{R}_{II,1}(t)$ and $\mathbf{R}_{II,2}(t)$, respectively. Our framework bridges the gap between idealized test functions and complex real-world applications by generating formulations that realistically emulate challenging scenarios. Consequently, our approach not only enables more meaningful assessments of algorithmic robustness, but also directly informs and accelerates the development of next-generation optimization techniques.

\subsection{Components with Variable Interaction}

Since $\mathbf{R}_{II,1}(t)$ and $\mathbf{R}_{II,2}(t)$ in Equation~\eqref{eq:gts_func_g} are typically set to identical matrices in most existing benchmarks, the resulting problems exhibit a separable structure, which significantly reduces optimization complexity (e.g., enabling the use of divide-and-conquer methods). However, this separability does not adequately reflect the complexities of real-world problems.

To address this limitation, we introduce time-varying positive semidefinite rotation matrices $\mathbf{R}_{II,1}(t)$ and $\mathbf{R}_{II,2}(t)$, which dynamically alter interactions among variables. This design transforms originally separable components into partially separable or even fully non-separable formulations through coordinated spatial transformations, thereby increasing optimization difficulty to better reflect the complexity of real-world problems. The construction method for these matrices is specified in Theorem~\ref{theorem:gts_pos_sem_matrix} below.

\begin{theorem}\label{theorem:gts_pos_sem_matrix}
    Let $A_n$ be an $n \times n$ real symmetric matrix such that:
    \begin{enumerate}
        \item Its diagonal elements are consecutive increasing natural numbers: $d_i = k + i - 1$ for $i = 1, 2, \ldots, n$, where $k \geq n$ is the starting number.
        \item Its off-diagonal elements are either $0$ or $1$.
    \end{enumerate}
    Then $A_n$ is positive definite (and hence positive semidefinite).
\end{theorem}
\textbf{Proof:} We prove the theorem by induction on $n$. Specifically, we show that for all $n \geq 1$ and all $k \geq n$, $A_n$ is positive definite by verifying that all leading principal minors are strictly positive.

\textbf{Case $n = 1$:} The matrix is $A_1 = [k]$ with $k \geq 1$. The only leading principal minor is $k \geq 1 > 0$. Thus, $A_1$ is positive definite.

\textbf{Case $n = 2$:} The matrix has the form:
\begin{equation*}
    A_2 = \begin{pmatrix}
        k & a   \\
        a & k+1
    \end{pmatrix}, \ a \in \{0, 1\}.
\end{equation*}
The leading principal minors are:
\begin{itemize}
    \item First-order: $k \geq 2 > 0$.
    \item Second-order:
          $$
              \det(A_2) = k(k+1) - a^2.
          $$
          If $a = 0$, then $\det(A_2) = k(k+1) \geq 2 \cdot 3 = 6 > 0$. \\
          If $a = 1$, then $\det(A_2) = k(k+1) - 1 \geq 2 \cdot 3 - 1 = 5 > 0$ (since $k \geq n$).
\end{itemize}
All leading principal minors are positive, so $A_2$ is positive definite.

\textbf{Case $n \geq 3$:} Assume the theorem holds for all $m < n$ (i.e., any $m \times m$ matrix satisfying the conditions is positive definite). Consider an $n \times n$ matrix $A_n$:
\begin{equation*}
    A_n = \begin{pmatrix}
        A_{n-1} & b   \\
        b^T     & d_n
    \end{pmatrix},
\end{equation*}
where, $A_{n-1}$ is the $(n-1) \times (n-1)$ leading principal submatrix with diagonal elements $d_i = k + i - 1$ for $i = 1, 2, \ldots, n-1$ and off-diagonal elements in $\{0, 1\}$. $b$ is an $(n-1) \times 1$ vector with entries $b_i \in \{0, 1\}$. $d_n = k + n - 1 \geq 2n - 1$ (since $k \geq n$). By the inductive hypothesis, $A_{n-1}$ is positive definite. Thus, $A_{n-1}$ is invertible, and $A_{n-1}^{-1}$ exists and is positive definite. All leading principal minors of $A_{n-1}$ are positive.

The determinant of $A_n$ is given by the block matrix formula:
\begin{equation*}
    \det(A_n) = \det(A_{n-1}) \cdot (d_n - b^T A_{n-1}^{-1} b).
\end{equation*}
Since $A_{n-1}$ is positive definite, $\det(A_{n-1}) > 0$. To prove $\det(A_n) > 0$, it suffices to show:
\begin{equation*}
    d_n - b^T A_{n-1}^{-1} b > 0 \ \iff \ b^T A_{n-1}^{-1} b < d_n.
\end{equation*}

Let $S \subseteq \{1, 2, \ldots, n-1\}$ be the set of indices where $b_i = 1$, then $|S| \leq n-1$,
\begin{equation*}
    b^T A_{n-1}^{-1} b \leq \mathbf{1}^T A_{n-1}^{-1} \mathbf{1} \leq \norm{\mathbf{1}} * \norm{A_{n-1}} \leq \frac{n-1}{\sigma_{min}(A_{n-1})},
\end{equation*}
by Gershgorin Circle Theorem~\cite{B_golub2013matrix},
\begin{equation*}
    \sigma_{min}(A_{n-1}) \geq \min_i \left( |a_{ii}| - \sum_{j \neq i} |a_{ij}| \right) \geq \min_i |a_{ii}| - \max_i \sum_{j \neq i} |a_{ij}| \geq 1,
\end{equation*}
so
\begin{equation*}
    b^T A_{n-1}^{-1} b \leq n - 1 < 2n -1 \leq d_n,
\end{equation*}
Thus, $d_n - b^T A_{n-1}^{-1} b > 0$, so $\det(A_n) > 0$.

For any leading principal minor of size $m < n$, it corresponds to a submatrix $A_m$ satisfying the same conditions (diagonal elements $k, k+1, \ldots, k+m-1$, off-diagonals $0$ or $1$). By the inductive hypothesis ($m < n$), $A_m$ is positive definite, so its determinant (the minor) is positive. Combined with $\det(A_n) > 0$, all leading principal minors of $A_n$ are positive. By the principle of mathematical induction, for all $n \geq 1$ and $k \geq n$, $A_n$ is positive definite. Hence, it is positive semi-definite.

This mechanism forces optimization algorithms to simultaneously adapt to changing variable dependencies while tracking moving PS/PF. Empirical studies confirm that the spectral properties of $\mathbf{R}_{II,1}(t)$ and $\mathbf{R}_{II,2}(t)$ directly correlate with problem difficulty, providing a theoretically grounded means to control benchmark complexity. This advancement not only refines current benchmarking practices but also facilitates the development of more robust optimization algorithms capable of addressing the challenges posed by real-world dynamic systems.

\subsection{Components with Temporally Irregular Dynamics}

Conventional approaches often rely on rigid temporal patterns controlled by a uniform step size $\Delta t = 1/n_t$, as illustrated in Equation~\eqref{eq:gts_time_regular}. While such regularity facilitates theoretical analysis, it fails to capture the irregular temporal changes that are inherent to many dynamic systems, such as the optimal control problem of a biological network~\cite{J_BMCSB_nimmegeers2016dynamic}, dynamic multi-objective route planning~\cite{J_TCBB_guo2018robust}, and discovery of changing communities in dynamic networks~\cite{J_TEVC_ma2024higher}, among others.

To address this limitation, we propose a novel method to introduce controlled irregularity, bridging the gap between artificial test problems and real-world applications. Our approach leverages the mathematical constant $\pi$ to generate pseudo-random perturbations. At each iteration $\tau$, we extract the $\left\lfloor \frac{\tau}{\tau_t} \right\rfloor$-th digit after the decimal point of $\pi$ (denoted as $\pi_{\tau}$), scale it by $0.5 \times \frac{1}{9}$, and add $\frac{1}{n_t} \times \left(0.5 \times \frac{\pi_{\tau}}{9}\right)$ to the original time variation magnitude (see Equation~\eqref{eq:gts_time_regular}). The resulting temporal update mechanism is defined as follows:
\begin{equation}\label{eq:gts_time_irregular}
    t = \frac{1}{n_t} \times \left\lfloor \frac{\tau}{\tau_t} \right\rfloor + \frac{1}{n_t} \times \left(0.5 \times \frac{\pi_{\tau}}{9}\right),
\end{equation}
where $\pi_{\tau}$ is given by:
\begin{equation}
    \pi_{\tau} =
    \begin{cases}
        0,                                                                               & \text{if } \left\lfloor \frac{\tau}{\tau_t} \right\rfloor=0, \\
        \left\lfloor \frac{\tau}{\tau_t} \right\rfloor-\text{-th} \text{ decimal digit of } \pi, & \text{otherwise.}
    \end{cases}
\end{equation}
When $\pi_{\tau} = 0$, the time variation degenerates to the form shown in Equation~\eqref{eq:gts_time_regular}. This design possesses two notable properties:
\begin{itemize}
    \item \textit{Reproducibility}: The deterministic digit sequence of $\pi$ guarantees numerical experimental repeatability.
    \item \textit{Compatibility}: The mechanism naturally reduces to the conventional form when $\pi_{\tau} = 0$.
\end{itemize}

Compared with existing approaches~\cite{J_TEVC_zou2025knowledge} that rely on non-deterministic elements such as the MATLAB random number generator (e.g., rng(42)), our method provides a controlled and reproducible framework for simulating such scenarios. By using the digits of $\pi$ as the perturbation source, we avoid the potential biases inherent in conventional pseudo‑random number generators~\cite{J_TEVC_zou2025knowledge}, while ensuring fair algorithm comparisons and repeatable experimental results.

\subsection{Components with Time-Linkage Property}

Many practical DMOPs exhibit a time-linkage~\cite{C_RIVF_nguyen2012dynamic, C_EvoWS_nguyen2009dynamic} phenomenon, i.e., historical solutions can influence landscapes of future problems. Consequently, an improper current decision or inaccurate solutions may cause modeling error accumulation over time, making the problems inherently time-deceptive.

While time-linkage has been considered in~\cite{J_ISCI_huang2011dynamic, J_TCYB_jiang2020scalable}, their PSs are limited to simple geometric forms, such as straight line segments or rectangular planes. Such forms represent a very special case and are far from sufficient to capture the general time-linkage property. To overcome this limitation, we propose a generalized component function that constructs time-linkage DMOPs by introducing a time-varying parameter $\phi(t)$ into the $g(\mathbf{x},t)$ term of Equation~\eqref{eq:gts_func_g} (see GTS6, GTS7, and GTS8 in Table~\ref{tab:gts_benchmark}). The resulting generalized function is defined as follows:
\begin{equation}
    \begin{split}
        g_{\text{time-linkage}}(\mathbf{x},t) = 1
         & + \Bigl(\bigl(\mathbf{x}_{II,1} - \phi(t) h_1(\mathbf{x}_I)\bigr)^T \mathbf{R}_{II,1}(t) \bigl(\mathbf{x}_{II,1} - \phi(t) h_1(\mathbf{x}_I)\bigr)\Bigr)^{\frac{1}{p}} \\
         & + \Bigl(\bigl(\mathbf{x}_{II,2} - \phi(t) h_2(\mathbf{x}_I)\bigr)^T \mathbf{R}_{II,2}(t) \bigl(\mathbf{x}_{II,2} - \phi(t) h_2(\mathbf{x}_I)\bigr)\Bigr)^{\frac{1}{p}}
    \end{split}
\end{equation}
with
\begin{equation}\label{eq:gts_time_linkage}
    \phi(t) =
    \begin{cases}
        1,                                                                                                  & \text{if } t=1,   \\
        1 + \left\Vert f(\mathbf{x}_{\text{knee}},t-1) - f(\hat{\mathbf{x}}_{\text{knee}},t-1) \right\Vert, & \text{otherwise},
    \end{cases}
\end{equation}
where $f(\mathbf{x}_{knee},t-1)$ and $f(\hat{\mathbf{x}_{knee}},t-1)$ are the objective values of the knee point of the true PF and estimated PF at time $t-1$, respectively.

According to Equation~\eqref{eq:gts_time_linkage}, if the solutions found by an algorithm approximate the true PF, $\phi(t)$ will approach $1$. In contrast, if the algorithm fails to converge at time $t-1$, $\phi(t)$ will be larger than 1. The larger $\phi(t)$ value, the greater model error at time $t-1$, and the error will accumulate over time. The above settings effectively emulate the time-deception and error accumulation characteristics of time-linkage. Furthermore, to ensure a fair comparison, all algorithms are evaluated against the same predefined baseline problem where $\phi(t)=1$.

Certainly, there are numerous alternative approaches to emulate time-linkage characteristics. For instance, the degree of temporal dependency $\phi(t)$ can be automatically adjusted based on algorithm performance metric values. After appropriate adjustment and normalization to the range $[0, 1]$, metrics such as the Inverted Generational Distance (IGD)~\cite{C_MICAI_coellocoello2004study} or Hypervolume (HV)~\cite{T_D_zitzler1999evolutionary} can be employed to define $\phi(t)$. The definitions are as follows:
\begin{equation*}
    \phi(t) =
    \begin{cases}
        1,           & \text{if } t=1,   \\
        2 - HV(t-1), & \text{otherwise,}
    \end{cases}
\end{equation*}
or
\begin{equation*}
    \phi(t) =
    \begin{cases}
        1,            & \text{if } t=1,   \\
        e^{IGD(t-1)}, & \text{otherwise.}
    \end{cases}
\end{equation*}
However, these indicators (especially HV) are computationally expensive, posing significant challenges in practical applications. Consequently, when selecting $\phi(t)$, the knee-point method (see Equation~\eqref{eq:gts_time_linkage}) is a primary choice, requiring a balanced consideration of computational efficiency, accuracy, and scalability. Future work should systematically compare these alternative time-linkage formulations to identify their respective advantages in modeling different dynamic systems.

\section{Generalized Test Suite}%
\label{sec:gts_gts_instances}

In this section, we provide a detailed formulation (see Table~\ref{tab:gts_benchmark}) of each benchmarking function in the GTS test suite, including its corresponding parameters as well as the associated PS and PF according to the above design principles. The Python code is openly available at our companion website \href{https://github.com/dynoptimization/pydmoo}{pydmoo}\footnote{\url{https://github.com/dynoptimization/pydmoo}}.

For convenience, the shared parameters used in the GTS benchmark problems are summarized in Table~\ref{tab:gts_symbols_gts}. The common control part of the PS, $g_{\text{base}}(\mathbf{x},t)$, is described as follows,
\begin{equation*}
    \begin{split}
        g_{\text{base}}(\mathbf{x},t) = 1
         & + \Bigl(\bigl(\mathbf{x}_{II,1} - h_1(\mathbf{x}_I)\bigr)^T \mathbf{R}_{II,1}(t) \bigl(\mathbf{x}_{II,1} - h_1(\mathbf{x}_I)\bigr)\Bigr)^{\frac{1}{p}} \\
         & + \Bigl(\bigl(\mathbf{x}_{II,2} - h_2(\mathbf{x}_I)\bigr)^T \mathbf{R}_{II,2}(t) \bigl(\mathbf{x}_{II,2} - h_2(\mathbf{x}_I)\bigr)\Bigr)^{\frac{1}{p}}
    \end{split}
\end{equation*}
where $p \geq 1$, the decision variable vector $\mathbf{x}$ is partitioned into three parts: the first part is $\mathbf{x}_I$, followed by $\mathbf{x}_{II,1}$ and $\mathbf{x}_{II,2}$. Here, $h_1(\mathbf{x}_I)$ and $h_2(\mathbf{x}_I)$ are functions of $\mathbf{x}_I$, which are problem-specific and designed to achieve different purposes. The PS is constructed by setting $g_{\text{base}}(\mathbf{x}, t) = 1$. This assignment implies that all components in $\mathbf{x}_{II,1}$ equal $h_1(\mathbf{x}_I)$, and all components in $\mathbf{x}_{II,2}$ equal $h_2(\mathbf{x}_I)$. Consequently, PF can be easily obtained by substituting the PS into the original objective functions.

\begin{table}[tb]
    \centering
    \caption{Parameters and functions used in the GTS}%
    \label{tab:gts_symbols_gts}
    \begin{tabular}{llll}
        \toprule
        $G(t) = \sin(0.5\pi   t)$        & $H(t) = 1.5 + G(t)$                 & $\alpha_t = 5\cos(0.5\pi t)$ & \\
        \midrule
        $\beta_t = 0.2 + 2.8\abs{G(t)}$  & $\omega_t = \lfloor 10G(t) \rfloor$ & $a_t = \sin(0.5\pi t)$       & \\
        \midrule
        $b_t = 1 + \abs{\cos(0.5\pi t)}$ & $y_t = 0.5 + G(t)(x_1 - 0.5)$       &                              & \\
        \bottomrule
    \end{tabular}
\end{table}

The test suite includes eight two-objective problems (GTS1-GTS8) and three three-objective problems (GTS9-GTS11), with GTS6 to GTS8 specifically designed as time-linkage cases. A key feature of this suite is its configurable dynamic behavior. The time variable $t$ can be set to either a regular pattern (see Equation~\eqref{eq:gts_time_regular}) or an irregular pattern (see Equation~\eqref{eq:gts_time_irregular}). Furthermore, the symmetric positive semidefinite matrices $\mathbf{R}_{II,1}$ and $\mathbf{R}_{II,2}$ are also adjustable, allowing the modification of the problem landscape. Notably, when $h_1(\mathbf{x}_I) = h_2(\mathbf{x}_I)$, the problem degenerates to an existing benchmark property where the PS changes on a hypersurface. The specific parameter configurations employed in this study are detailed in Section~\ref{sec:gts_benchmark_problems}.

Due to page limitations, the full set of reader-friendly, color versions of each problem in the GTS test suite, along with the associated figures depicting the PS and PF, are provided in the Supplementary Materials (see Section I). This section only presents a summarized overview (see Table~\ref{tab:gts_benchmark}) and a discussion of their key characteristics (see Table~\ref{tab:gts_characteristics}). $M$ represents the number of objective functions in Table~\ref{tab:gts_characteristics}.

\begin{table*}[htbp]
    \centering
    \caption{Definitions of the GTS Test Suite}%
    \label{tab:gts_benchmark}
    \resizebox*{!}{18.5cm}{
        \begin{tabular}{ll}
            \toprule
            Problem & Definition                                                                                                                                                                                                                                                      \\ \midrule
            GTS1    &
            \begin{tabular}[c]{@{}l@{}}
                $f_1(\mathbf{x},t) = x_1$, $f_2(\mathbf{x},t) = g(\mathbf{x},t)(1 -  (\frac{x_1}{g(\mathbf{x},t)})^{H(t)})$                                                                                                              \\
                $g(\mathbf{x},t) = g_{\text{base}}(\mathbf{x},t)$, $\mathbf{x}_I = (x_1)$, $\mathbf{x}_{II,1} = (x_2, \cdots, x_{\lfloor\frac{D}{2}\rfloor})$ and $\mathbf{x}_{II,2} = (x_{\lfloor\frac{D}{2}\rfloor + 1}, \cdots, x_D)$ \\
                $h_1(\mathbf{x}_I, t) = \cos(0.5\pi t)$ and $h_2(\mathbf{x}_I, t) = G(t) + x_1^{H(t)}$                                                                                                                                   \\
                Search space: $[0,1] \times [-1,1]^{\lfloor\frac{D}{2}\rfloor -1} \times  [-1, 2]^{\lceil\frac{D}{2}\rceil}$                                                                                                             \\
                PS(t): $0 \leq x_1 \leq 1, x_i = h_1(\mathbf{x}_I, t), x_i \in \mathbf{x}_{II,1},  x_j = h_2(\mathbf{x}_I, t), \in \mathbf{x}_{II,2}$,
                PF(t): a part of $f_2 = 1 - f_1^{H(t)}, 0 \leq f_1 \leq 1$
            \end{tabular}                                              \\ \midrule

            GTS2    &
            \begin{tabular}[c]{@{}l@{}}
                $f_1(\mathbf{x},t) = 0.5x_1+x_2$, $f_2(\mathbf{x},t) = g(\mathbf{x},t)(2.8 - (\frac{0.5x_1+x_2}{g(\mathbf{x},t)})^{H(t)})$                                                                                                        \\
                $g(\mathbf{x},t) = g_{\text{base}}(\mathbf{x},t)$, $\mathbf{x}_I = (x_1, x_2)$, $\mathbf{x}_{II,1} = (x_3, \cdots, x_{\lfloor\frac{D}{2}\rfloor + 1})$ and $\mathbf{x}_{II,2} = (x_{\lfloor\frac{D}{2}\rfloor + 2}, \cdots, x_D)$ \\
                $h_1(\mathbf{x}_I, t) = \frac{1}{\pi}\abs{\arctan(c)}$ and $h_2(\mathbf{x}_I, t) = G(t) + x_1^{H(t)}$                                                                                                                             \\
                $c = \cot(3\pi t^2), \text{when } t^2 \neq \frac{n}{3}, n \in \mathbb{Z}, c = 1e-32, \text{otherwise}$                                                                                                                            \\
                Search space: $[0,1]^2 \times [0,1]^{\lfloor\frac{D}{2}\rfloor -1} \times  [-1, 2]^{\lceil\frac{D}{2}\rceil-1}$                                                                                                                   \\
                PS(t): $0 \leq x_{1,2} \leq 1, x_i = h_1(\mathbf{x}_I, t), x_i \in \mathbf{x}_{II,1},  x_j = h_2(\mathbf{x}_I, t), \in \mathbf{x}_{II,2}$,
                PF(t): a part of $f_2 = 2.8 - f_1^{H(t)}, 0 \leq f_1 \leq 1.5$
            \end{tabular}                                     \\ \midrule

            GTS3    &
            \begin{tabular}[c]{@{}l@{}}
                $f_1(\mathbf{x},t) = g(x)(x_1 +  0.1\sin(3\pi x_1))^{\beta_t}$, $f_2(\mathbf{x},t) = g(\mathbf{x},t)(1 - x_1 + 0.1\sin(3\pi  x_1))^{\beta_t}$                                                                            \\
                $g(\mathbf{x},t) = g_{\text{base}}(\mathbf{x},t)$, $\mathbf{x}_I = (x_1)$, $\mathbf{x}_{II,1} = (x_2, \cdots, x_{\lfloor\frac{D}{2}\rfloor})$ and $\mathbf{x}_{II,2} = (x_{\lfloor\frac{D}{2}\rfloor + 1}, \cdots, x_D)$ \\
                $h_1(\mathbf{x}_I, t) = \frac{G(t)\sin(4\pi x_1)}{1 + \abs{G(t)}}$ and $h_2(\mathbf{x}_I, t) = G(t) + x_1^{H(t)}$                                                                                                        \\
                Search space: $[0,1] \times [-1,1]^{\lfloor\frac{D}{2}\rfloor - 1} \times  [-1, 2]^{\lceil\frac{D}{2}\rceil}$                                                                                                            \\
                PS(t): $0 \leq x_1 \leq 1, x_i = h_1(\mathbf{x}_I, t), x_i \in \mathbf{x}_{II,1},  x_j = h_2(\mathbf{x}_I, t), \in \mathbf{x}_{II,2}$,
                PF(t): a part of $f_1^{\frac{1}{\beta_t}} + f_2^{\frac{1}{\beta_t}} = 1 +  0.2\sin(3\pi \frac{f_1^{\frac{1}{\beta_t}} -f_2^{\frac{1}{\beta_t}} + 1}{2}),  0 \leq f_1 \leq 1$
            \end{tabular}                                              \\ \midrule

            GTS4    &
            \begin{tabular}[c]{@{}l@{}}
                $f_1(\mathbf{x},t) = g(\mathbf{x},t)\frac{1 +  t}{x_1 + 3}$, $f_2(\mathbf{x},t) = g(\mathbf{x},t)\frac{x_1 + 3}{1 + t}$                                                                                                                             \\
                $g(\mathbf{x},t) = g_{\text{base}}(\mathbf{x},t) - 0.5 + 0.25\sin(0.3\pi t)$, $\mathbf{x}_I = (x_1)$, $\mathbf{x}_{II,1} = (x_2, \cdots, x_{\lfloor\frac{D}{2}\rfloor})$ and $\mathbf{x}_{II,2} = (x_{\lfloor\frac{D}{2}\rfloor + 1}, \cdots, x_D)$ \\
                $h_1(\mathbf{x}_I, t) = \abs{G(t)}$ and $h_2(\mathbf{x}_I, t) = \frac{G(t)\sin(4\pi x_1)}{1 + \abs{G(t)}}$                                                                                                                                          \\
                Search space: $[0,1] \times [0,1]^{\lfloor\frac{D}{2}\rfloor -1} \times  [-1, 1]^{\lceil\frac{D}{2}\rceil}$                                                                                                                                         \\
                PS(t): $0 \leq x_1 \leq 1, x_i = h_1(\mathbf{x}_I, t), x_i \in \mathbf{x}_{II,1},  x_j = h_2(\mathbf{x}_I, t), \in \mathbf{x}_{II,2}$,
                PF(t): a part of $f_2 = \frac{1}{f_1}, \frac{1+t}{16} \leq f_1 \leq \frac{1+t}{4}$
            \end{tabular}                   \\ \midrule

            GTS5    &
            \begin{tabular}[c]{@{}l@{}}
                $f_1(\mathbf{x},t) =  g(\mathbf{x},t)((0.5x_1+x_2) + 0.02\sin(\omega_t\pi (0.5x_1+x_2)))$, $f_2(\mathbf{x},t) = g(\mathbf{x},t)(1.6 - (0.5x_1+x_2) + 0.02\sin(\omega_t\pi  (0.5x_1+x_2)))$                                                        \\
                $g(\mathbf{x},t) = g_{\text{base}}(\mathbf{x},t) + 0.5 + 0.5G(t)$, $\mathbf{x}_I = (x_1, x_2)$, $\mathbf{x}_{II,1} = (x_3, \cdots, x_{\lfloor\frac{D}{2}\rfloor + 1})$ and $\mathbf{x}_{II,2} = (x_{\lfloor\frac{D}{2}\rfloor + 2}, \cdots, x_D)$ \\
                $h_1(\mathbf{x}_I, t) = \cos(0.5\pi t)$ and $h_2(\mathbf{x}_I, t) = G(t) + x_1^{H(t)}$                                                                                                                                                            \\
                Search space: $[0,1]^2 \times [-1,1]^{\lfloor\frac{D}{2}\rfloor - 1} \times  [-1, 2]^{\lceil\frac{D}{2}\rceil-1}$                                                                                                                                 \\
                PS(t): $0 \leq x_{1,2} \leq 1, x_i = h_1(\mathbf{x}_I, t), x_i \in \mathbf{x}_{II,1},  x_j = h_2(\mathbf{x}_I, t), \in \mathbf{x}_{II,2}$                                                                                                         \\
                PF(t): $(f_1 + f_2) = (1.6 + 0.5G(t))(1 +  0.04\sin(\omega_t\pi\frac{\frac{1}{1.6 + 0.5G(t)}(f_1 - f_2) + 1.6}{2})), 0  \leq f_1 \leq 3$
            \end{tabular}                     \\ \midrule

            GTS6    &
            \begin{tabular}[c]{@{}l@{}}
                $f_1(\mathbf{x},t) = x_1$, $f_2(\mathbf{x},t) =  g(\mathbf{x},t)(1 - (\frac{x_1}{g(\mathbf{x},t)})^{H(t)})$                                                                                                                      \\
                $g(\mathbf{x},t) = g_{\text{time-linkage}}(\mathbf{x},t)$, $\mathbf{x}_I = (x_1)$, $\mathbf{x}_{II,1} = (x_2, \cdots, x_{\lfloor\frac{D}{2}\rfloor})$ and $\mathbf{x}_{II,2} = (x_{\lfloor\frac{D}{2}\rfloor + 1}, \cdots, x_D)$ \\
                $h_1(\mathbf{x}_I, t) = \cos(0.5\pi t)$ and $h_2(\mathbf{x}_I, t) = G(t) + x_1^{H(t)}$                                                                                                                                           \\
                Search space: $[0,1] \times [-1,1]^{\lfloor\frac{D}{2}\rfloor -1} \times  [-1, 2]^{\lceil\frac{D}{2}\rceil}$                                                                                                                     \\
                PS(t): $0 \leq x_1 \leq 1, x_i = h_1(\mathbf{x}_I, t), x_i \in \mathbf{x}_{II,1},  x_j = h_2(\mathbf{x}_I, t), \in \mathbf{x}_{II,2}$,
                PF(t): $f_2 = 1 - f_1^{H(t)}, 0 \leq f_1 \leq 1$
            \end{tabular}                                      \\ \midrule

            GTS7    &
            \begin{tabular}[c]{@{}l@{}}
                $f_1(\mathbf{x},t) =  g(\mathbf{x},t)\abs{x_1-a_t}^{H(t)}$, $f_2(\mathbf{x},t) = g(\mathbf{x},t)\abs{x_1-a_t-b_t}^{H(t)}$                                                                                                        \\
                $g(\mathbf{x},t) = g_{\text{time-linkage}}(\mathbf{x},t)$, $\mathbf{x}_I = (x_1)$, $\mathbf{x}_{II,1} = (x_2, \cdots, x_{\lfloor\frac{D}{2}\rfloor})$ and $\mathbf{x}_{II,2} = (x_{\lfloor\frac{D}{2}\rfloor + 1}, \cdots, x_D)$ \\
                $h_1(\mathbf{x}_I, t) = \cos(0.5\pi t)$ and $h_2(\mathbf{x}_I, t) = \frac{1}{1 + e^{\alpha_t(x_1 - 0.5)}}$                                                                                                                       \\
                Search space: $[-1,2.5] \times [-1,1]^{\lfloor\frac{D}{2}\rfloor - 1} \times  [0, 1]^{\lceil\frac{D}{2}\rceil}$                                                                                                                  \\
                PS(t): $a_t \leq x_1 \leq (a_t + b_t), x_i = h_1(\mathbf{x}_I, t), x_i \in \mathbf{x}_{II,1},  x_j = h_2(\mathbf{x}_I, t), \in \mathbf{x}_{II,2}$,
                PF(t): $f_2 = (b_t - f_1^{\frac{1}{H(t)}})^{H(t)}, 0 \leq f_1 \leq 3.5$
            \end{tabular}                                      \\ \midrule

            GTS8    &
            \begin{tabular}[c]{@{}l@{}}
                $f_1(\mathbf{x},t) = (0.5x_1+x_2)$, $f_2(\mathbf{x},t) = g(\mathbf{x},t)(2.8 - (\frac{(0.5x_1+x_2)}{g(\mathbf{x},t)})^{H(t)})$                                                                                                                                        \\
                $g(\mathbf{x},t) = g_{\text{time-linkage}}(\mathbf{x},t) + 0.25\abs{\cos(0.3 \pi t)}$, $\mathbf{x}_I = (x_1, x_2)$, $\mathbf{x}_{II,1} = (x_3, \cdots, x_{\lfloor\frac{D}{2}\rfloor + 1})$ and $\mathbf{x}_{II,2} = (x_{\lfloor\frac{D}{2}\rfloor + 2}, \cdots, x_D)$ \\
                $h_1(\mathbf{x}_I, t) = \frac{1}{1 + e^{\alpha _t(x_1 - 0.5)}}$ and $h_2(\mathbf{x}_I, t) = G(t) + x_1^{H(t)}$                                                                                                                                                        \\
                Search space: $[0,1]^2 \times [0,1]^{\lfloor\frac{D}{2}\rfloor -1} \times  [-1, 2]^{\lceil\frac{D}{2}\rceil-1}$                                                                                                                                                       \\
                PS(t): $0 \leq x_{1,2} \leq 1, x_i = h_1(\mathbf{x}_I, t), x_i \in \mathbf{x}_{II,1},  x_j = h_2(\mathbf{x}_I, t), \in \mathbf{x}_{II,2}$,
                PF(t): $\frac{f_2}{1 + 0.25\abs{\cos(0.3\pi t)}} = 2.8 - (\frac{f_1}{1 +  0.25\abs{\cos(0.3\pi t)}})^{H(t)}, 0 \leq f_1 \leq 1.5$
            \end{tabular} \\ \midrule

            GTS9    &
            \begin{tabular}[c]{@{}l@{}}
                $f_1(\mathbf{x},t) = g(\mathbf{x},t)\cos(0.5\pi  x_1)\cos(0.5\pi x_2)$, $f_2(\mathbf{x},t) = g(\mathbf{x},t)\cos(0.5\pi x_1)\sin(0.5\pi x_2)$, $f_3(\mathbf{x},t) = g(\mathbf{x},t)\sin(0.5\pi x_1)$                                                      \\
                $g(\mathbf{x},t) = g_{\text{base}}(\mathbf{x},t) + \abs{\cos(0.27\pi t)}$, $\mathbf{x}_I = (x_1, x_2)$, $\mathbf{x}_{II,1} = (x_3, \cdots, x_{\lfloor\frac{D}{2}\rfloor + 1})$ and $\mathbf{x}_{II,2} = (x_{\lfloor\frac{D}{2}\rfloor + 2}, \cdots, x_D)$ \\
                $h_1(\mathbf{x}_I, t) = \frac{1}{1+e^{\alpha_t(x_1 - 0.5)}}$ and $h_2(\mathbf{x}_I, t) = \sin(tx_1)$                                                                                                                                                      \\
                Search space: $[0,1]^2 \times [0,1]^{\lfloor\frac{D}{2}\rfloor - 1} \times  [-1, 1]^{\lceil\frac{D}{2}\rceil - 1}$                                                                                                                                        \\
                PS(t): $0 \leq x_{1,2} \leq 1, x_i = h_1(\mathbf{x}_I, t), x_i \in \mathbf{x}_{II,1},  x_j = h_2(\mathbf{x}_I, t), \in \mathbf{x}_{II,2}$,
                PF(t): $\sum_{i = 1}^{3}f_i^2 = 1 + \abs{\cos(0.27\pi t)}, 0 \leq f_{i =  1,2,3} \leq 2$
            \end{tabular}             \\ \midrule

            GTS10   &
            \begin{tabular}[c]{@{}l@{}}
                $f_1(\mathbf{x},t) =  g(\mathbf{x},t)\cos^2(0.5\pi x_1)$, $f_2(\mathbf{x},t) = g(\mathbf{x},t)\cos^2(0.5\pi x_2)$, $f_3(\mathbf{x},t) = g(\mathbf{x},t)\sum_{j = 1}^{2}(\sin^2(0.5\pi x_j) + \sin(0.5\pi  x_j)\cos^2(\lfloor6G(t)\rfloor \pi x_j))$ \\
                $g(\mathbf{x},t) = g_{\text{base}}(\mathbf{x},t)$, $\mathbf{x}_I = (x_1, x_2)$, $\mathbf{x}_{II,1} = (x_3, \cdots, x_{\lfloor\frac{D}{2}\rfloor + 1})$ and $\mathbf{x}_{II,2} = (x_{\lfloor\frac{D}{2}\rfloor + 2}, \cdots, x_D)$                   \\
                $h_1(\mathbf{x}_I, t) = \abs{G(t)}$ and $h_2(\mathbf{x}_I, t) = -0.5 + \frac{\abs{G(t)\sin(4\pi x_1)}}{0.5(1+\abs{G(t)})}$                                                                                                                          \\
                Search space: $[0,1]^2 \times [0,1]^{\lfloor\frac{D}{2}\rfloor - 1} \times  [-1, 1]^{\lceil\frac{D}{2}\rceil - 1}$                                                                                                                                  \\
                PS(t): $0 \leq x_{1,2} \leq 1, x_i = h_1(\mathbf{x}_I, t), x_i \in \mathbf{x}_{II,1},  x_j = h_2(\mathbf{x}_I, t), \in \mathbf{x}_{II,2}$
            \end{tabular}                   \\ \midrule

            GTS11   &
            \begin{tabular}[c]{@{}l@{}}
                $f_1(\mathbf{x},t) = g(\mathbf{x},t)(1.05 - y +  0.05\sin(6\pi y))$, $f_2(\mathbf{x},t) = g(\mathbf{x},t)(1.05 - x_2 + 0.05\sin(6\pi x_2))(y +  0.05\sin(6\pi y))$,                                                               \\
                $f_3(\mathbf{x},t) = g(\mathbf{x},t)(x_2 + 0.05\sin(6\pi x_2))(y + 0.05\sin(6\pi y))$                                                                                                                                             \\
                $g(\mathbf{x},t) = g_{\text{base}}(\mathbf{x},t)$, $\mathbf{x}_I = (x_1, x_2)$, $\mathbf{x}_{II,1} = (x_3, \cdots, x_{\lfloor\frac{D}{2}\rfloor + 1})$ and $\mathbf{x}_{II,2} = (x_{\lfloor\frac{D}{2}\rfloor + 2}, \cdots, x_D)$ \\
                $h_1(\mathbf{x}_I, t) = \abs{G(t)}$ and $h_2(\mathbf{x}_I, t) = G(t) + x_1^{H(t)}$                                                                                                                                                \\
                Search space: $[0,1]^2 \times [0,1]^{\lfloor\frac{D}{2}\rfloor - 1} \times  [-1, 2]^{\lceil\frac{D}{2}\rceil - 1}$                                                                                                                \\
                PS(t): $0 \leq x_{1,2} \leq 1, x_i = h_1(\mathbf{x}_I, t), x_i \in \mathbf{x}_{II,1},  x_j = h_2(\mathbf{x}_I, t), \in \mathbf{x}_{II,2}$
            \end{tabular}                                     \\ \bottomrule
        \end{tabular}
    }
\end{table*}

\begin{table*}[htbp]
    \centering
    \caption{Characteristics of the GTS test suite}%
    \label{tab:gts_characteristics}
    \begin{tabular}{p{0.09\textwidth}p{0.02\textwidth}p{0.45\textwidth}p{0.35\textwidth}} \toprule
        Problem & $M$ & PS                                                                                                     & PF                                                                 \\ \midrule
        GTS1    & 2   & disconnected PS, PS varying on the hypersurface, simple variable-linkage                               & mixed convexity-concavity, disconnected PF                         \\ \midrule
        GTS2    & 2   & PS moving on the hypersurface, simple variable-linkage                                                 & mixed convexity-concavity, fixed intersection point                \\ \midrule
        GTS3    & 2   & PS changing in decision space, simple variable-linkage, the length of PS changing                      & dynamic PF, knee point                                             \\ \midrule
        GTS4    & 2   & time-varying PS, simple variable-linkage, degeneration                                                 & dynamic PF, PF range changing                                      \\ \midrule
        GTS5    & 2   & dynamic PS, simple variable-linkage                                                                    & dynamic PF, knee point, degeneration                               \\ \midrule
        GTS6    & 2   & time-linkage PS, PS moving on the hyper-surface, simple variable-linkage                               & mixed convexity-concavity                                          \\ \midrule
        GTS7    & 2   & time-linkage PS, PS changing in the decision space, the length of PS changing, simple variable-linkage & mixed convexity-concavity, the length and curvature of PF changing \\ \midrule
        GTS8    & 2   & time-linkage PS, PS varying  in the decision space, simple variable-linkage                            & mixed convexity-concavity                                          \\ \midrule
        GTS9    & 3   & PS moving in the space, time-varying geometry shape of PS, simple variable-linkage                     & disconnected PF                                                    \\ \midrule
        GTS10   & 3   & PS varying on the hypersurface, simple variable-linkage, degeneration                                  & disconnected segments                                              \\ \midrule
        GTS11   & 3   & PS varying on the hypersurface, simple variable-linkage                                                & degeneration                                                       \\ \bottomrule
    \end{tabular}
\end{table*}

\section{Experiments}%
\label{sec:gts_experiments}

In this section, comprehensive numerical experiments on the DF test suite~\cite{R_jiang2018benchmark}, along with two additional functions FDA4 and FDA5~\cite{J_TEVC_farina2004dynamic}, as well as the GTS test suites, are conducted to demonstrate the advantages of the GTS test suite in enabling fair and effective comparisons of dynamic multi-objective optimization algorithms. For convenience, in this study, the term ``DF test suite'' herein represents the original DF test suite as well as the FDA4 and FDA5 functions.

\subsection{Benchmark Problems}%
\label{sec:gts_benchmark_problems}

For the GTS test suite\footnote{\url{https://github.com/dynoptimization/pydmoo}}, we define three distinct groups. The first group (Group 1) sets both $\mathbf{R}_{II,1}$ and $\mathbf{R}_{II,2}$ to identity matrices $\mathbf{I}$, representing simple, separable problems. In the second group (Group 2), both matrices are diagonal, with their diagonal elements set to the sequence of natural numbers starting from 1 (i.e., 1, 2, 3,~...), introducing imbalanced properties. The third group (Group 3) uses symmetric matrices for $\mathbf{R}_{II,1}$ and $\mathbf{R}_{II,2}$. Their diagonal elements form a sequence of natural numbers starting from the matrix dimension plus one, while all off-diagonal elements are set to 1, resulting in a positive semidefinite matrix (see Theorem~\ref{theorem:gts_pos_sem_matrix}) designed to test variable interactions.

The source code for the DF test suite was obtained from the well-known multi-objective Python library pymoo\footnote{https://github.com/anyoptimization/pymoo}~\cite{J_ACCESS_blank2020pymoo}, and executed using their default configuration without any modification.

\subsection{Benchmarking Algorithms}

Six state-of-the-art algorithms, i.e., PPS~\cite{J_TCYB_zhou2014population}, DPIM~\cite{J_ISAT_li2021dual}, AE~\cite{J_TCYB_feng2022solving}, IGP~\cite{J_TCYB_zhang2022inverse}, KGB~\cite{J_KBS_ye2022knowledge}, and KTMM~\cite{J_TEVC_zou2025knowledge} are employed for experimental study. Among them, PPS~\cite{J_TCYB_zhou2014population} and AE~\cite{J_TCYB_feng2022solving} are classic prediction-based approaches, and IGP~\cite{J_TCYB_zhang2022inverse} and DPIM~\cite{J_ISAT_li2021dual} are methods that predict the next time PF in objective space and then map the predicted PF to approximated PS in the decision space. KGB~\cite{J_KBS_ye2022knowledge} and KTMM~\cite{J_TEVC_zou2025knowledge} are two novel knowledge-based methods, using historical information to construct next time approximated PS. Parameters of each algorithm are set to the same values as in the original paper.

\subsection{Experimental Setup}

Dynamics for DF and GTS were controlled by Equation~\eqref{eq:gts_time_regular} and Equation~\eqref{eq:gts_time_irregular}, respectively. The severity of change was set to $n_t \in \{5, 10\}$ and the number of generations available in each time step was $\tau_t=\{5, 10\}$. $\tau_t=5$ represents fast change scenarios, while $\tau_t=10$ indicates a general changing speed. $T=50$ represents the number of time steps involved in each run. $50$ generations were executed for each algorithm before the first change to minimize the effect of static optimization~\cite{R_jiang2018benchmark}. Algorithms terminated after $N \times \tau_t \times T$ fitness evaluations, where $N$ is the population size. Each algorithm was run $21$ times on each test instance, with the problem dimension set to $10$.

\subsection{Performance Metrics}

In this study, three widely-used performance metrics, i.e.,  Mean Inverted Generational Distance (MIGD)~\cite{J_TCYB_zhou2014population, C_MICAI_coellocoello2004study}, Mean Hypervolume (MHV)~\cite{J_TEVC_zitzler1999multiobjective}, Mean Maximum Spread (MMS)~\cite{J_TEVC_goh2009competitivecooperative}, are employed to fairly and comprehensively evaluate algorithmic performance.

IGD assesses the convergence and diversity of the evolved PF which is an approximation to the true PF. A smaller IGD value reflects a small deviation between the evolved and true PF. MIGD refers to the average IGD value of all time steps in a run
\begin{equation}
    \mathrm{MIGD}=\frac{1}{T}\sum_{K=1}^T\frac{1}{N}\sum_{i=1}^{N} d_i^K,
\end{equation}
where $T$ is the number of time steps involved in a run, $N$ is the number of reference points, and $d_i^K$ is the Euclidean distance between the $i$th reference point and the evolved PF at time $K$. The reference points were uniformly sampled from the true PF at each time step, where $N=1500$ for the two-objective GTS1-8 and $N=2500$ for the three-objective GTS9-11 problems.

The hypervolume (HV) quantifies the size of the objective space that is dominated by a non-dominated solution set $\text{PF}^t$ and bounded above by a reference point $\mathbf{Z}_{ref}$. This metric comprehensively evaluates both the convergence and distribution of the set, with a larger value being preferable. MHV refers to the average HV value of all time steps in a run
\begin{equation}
    \mathrm{MHV} = \frac{1}{T}\sum_{K=1}^T \bigcup_{i=1}^{N^*} V_i
\end{equation}
where $V_i$ represents the volume of the hypercube spanned by the $i$-th solution in $\text{PF}^t$ and the reference point $\mathbf{Z}_{ref}$, and $N^* = |\text{PF}^t|$ is the cardinality of the solution set at time t.

The original MS indicator misjudges when all the obtained solutions are out of the bounds of the true PF. To avoid this unexpected situation, we propose a revised version called MS2, which measures the distribution of the obtained solutions, i.e., the coverage of the obtained solutions over the true PF. A larger MS2 result indicates a better solution distribution. Similar to the MIGD metric, we calculate the average MS2 value of all the $T$ time periods in a run
\begin{equation}
    \mathrm{MMS}=\frac{1}{T}\sum_{t=1}^T\mathrm{MS2}^t,
\end{equation}
where
\begin{equation}\label{eq:gts_ms2}
    \text{MS2}^t = \prod_{j=1}^{M}\Theta(\text{PF}_{\text{max}}^{t,j}- \text{PF}_{\text{min}}^{t,j,*}) * \Theta(\text{PF}_{\text{max}}^{t,j,*}- \text{PF}_{\text{min}}^{t,j}) * \text{MS}^t
\end{equation}
in Equation~\eqref{eq:gts_ms2}, $\Theta(x) = 1, \text{when}~x \geq 0, \Theta(x) = 0, \text{when}~x < 0$,
\begin{equation}\label{eq:gts_ms}
    \text{MS}^t = \sqrt{\frac{1}{M}\sum_{j = 1}^{M}\left[\frac{\min(\text{PF}_{\text{max}}^{t,j},\text{PF}_{\text{max}}^{t,j,*})-\max(\text{PF}_{\text{min}}^{t,j},\text{PF}_\text{{min}}^{t,j,*})}{\text{PF}_{\text{max}}^{t,j}-\text{PF}_{\text{min}}^{t,j}}\right]^2}
\end{equation}
in Equation~\eqref{eq:gts_ms2} and Equation~\eqref{eq:gts_ms}, $\text{PF}_{\text{max}}^{t,j,*}$ and $\text{PF}_{\text{min}}^{t,j,*}$ are the maximum and minimum values of the $j$th objective in the final population at time $t$, respectively; while $\text{PF}_{\text{max}}^{t,j}$ and $\text{PF}_{\text{min}}^{t,j}$ are the maximum and minimum values of the $j$th objective in the true $\text{PF}^t$.

\subsection{Experimental Results}

Due to page limitations, all tables for MIGD, MHV, MMS, and Runtime are provided in the Supplementary Materials (see Section II). The dynamic mean inverted generational distance (DMIGD), dynamic mean hypervolume (DMHV), and dynamic mean maximum spread (DMMS) are defined as follows:
\begin{align}
    DMIGD & = \frac{1}{|P|} \sum_{p \in P} \frac{1}{|C|} \sum_{c \in C} \frac{1}{|R|} \sum_{r \in R} MIGD(p, c, r) \\
    DMHV  & = \frac{1}{|P|} \sum_{p \in P} \frac{1}{|C|} \sum_{c \in C} \frac{1}{|R|} \sum_{r \in R} MHV(p, c, r)  \\
    DMMS  & = \frac{1}{|P|} \sum_{p \in P} \frac{1}{|C|} \sum_{c \in C} \frac{1}{|R|} \sum_{r \in R} MMS(p, c, r)
\end{align}
where $P$ is the set of benchmark problems, $C$ is the set of configurations for a given problem, and $R$ is the set of independent run repeats. In this study, we set $R = \{1, 2, \cdots, 20\}$ and the symbol $|\cdot|$ denotes the cardinality of a set.

\subsubsection{Performance Analysis}

\begin{figure*}[!t]
    \centering
    \subfloat[\scriptsize DF]{\includegraphics[width=0.24\textwidth,height=0.25\textwidth]{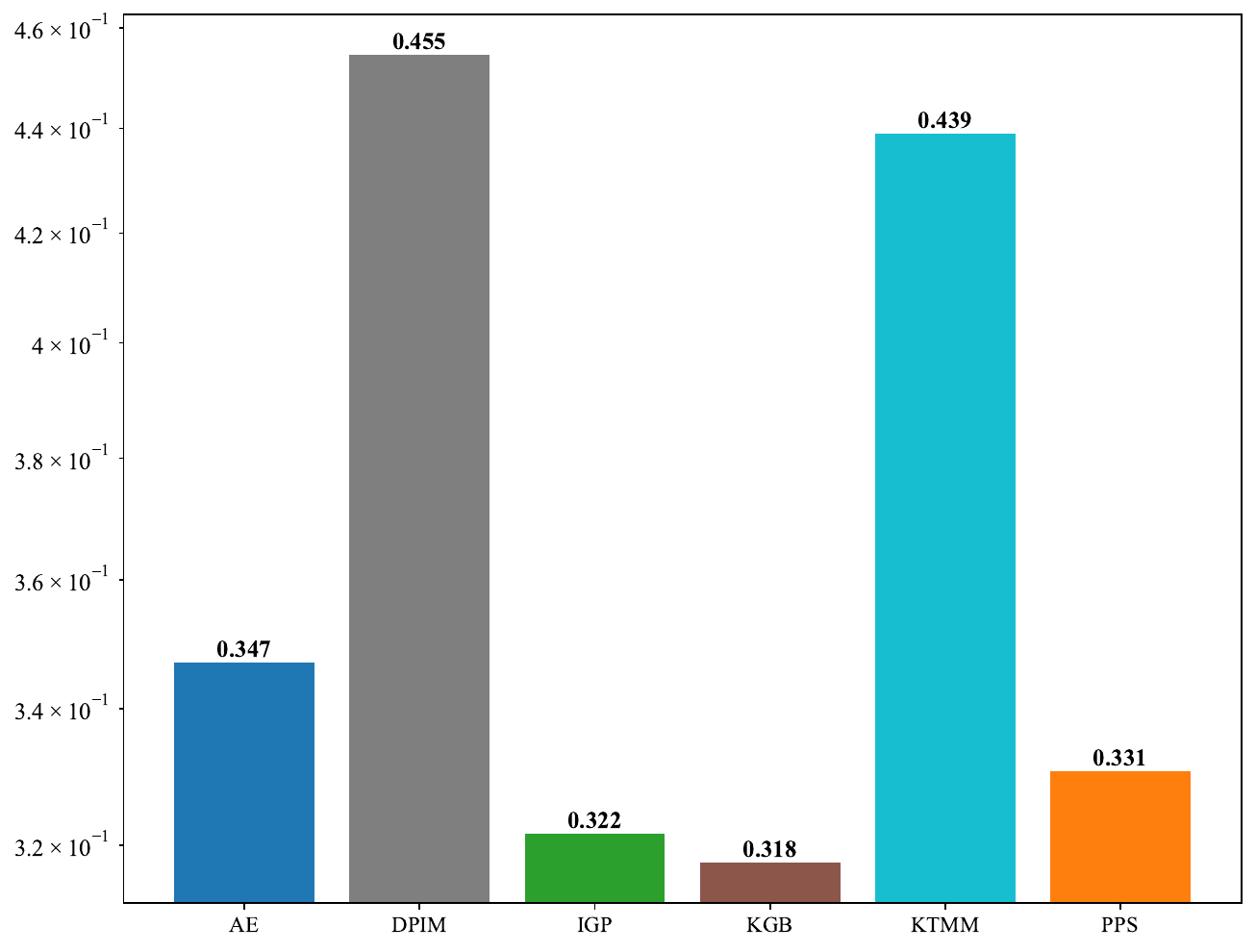}%
        \label{fig:gts_df_nsga2_stat_fast_505010_end_igd2_mean_bar}}
    \hfil
    \subfloat[\scriptsize GTS Group 1]{\includegraphics[width=0.24\textwidth,height=0.25\textwidth]{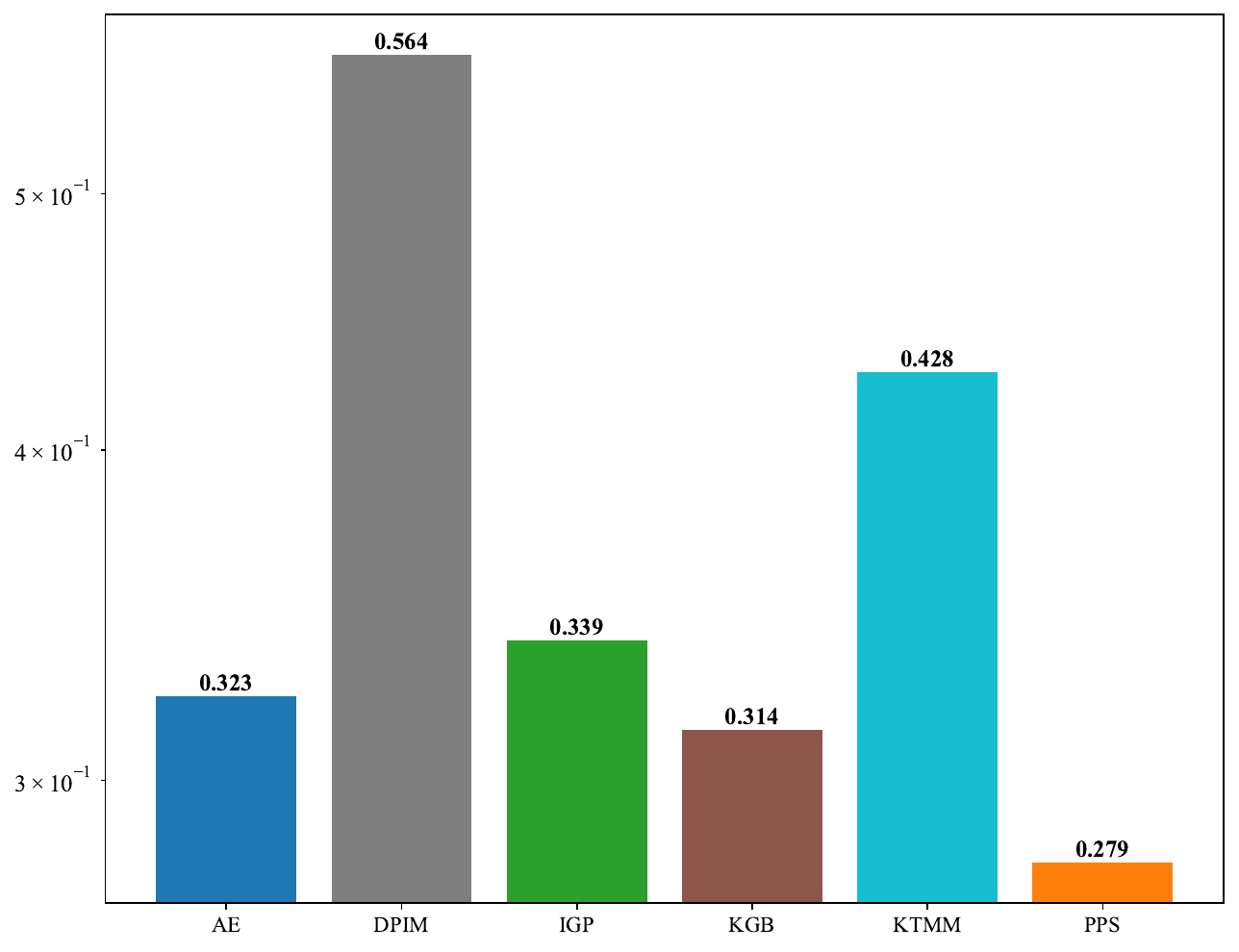}%
        \label{fig:gts_gts1_nsga2_stat_fast_505010_end_igd2_mean_bar}}
    \hfil
    \subfloat[\scriptsize GTS Group 2]{\includegraphics[width=0.24\textwidth,height=0.25\textwidth]{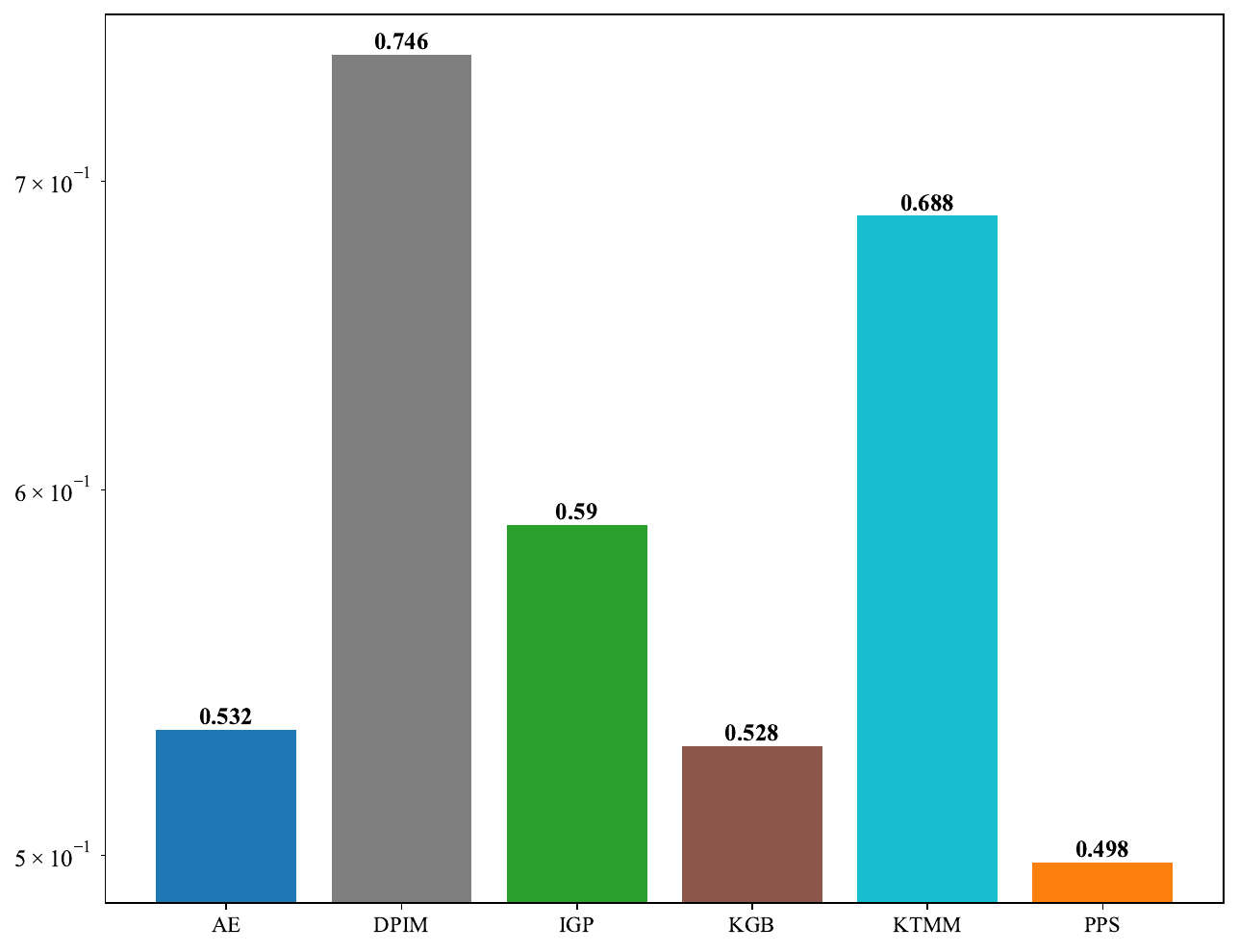}%
        \label{fig:gts_gts2_nsga2_stat_fast_505010_end_igd2_mean_bar}}
    \hfil
    \subfloat[\scriptsize GTS Group 3]{\includegraphics[width=0.24\textwidth,height=0.25\textwidth]{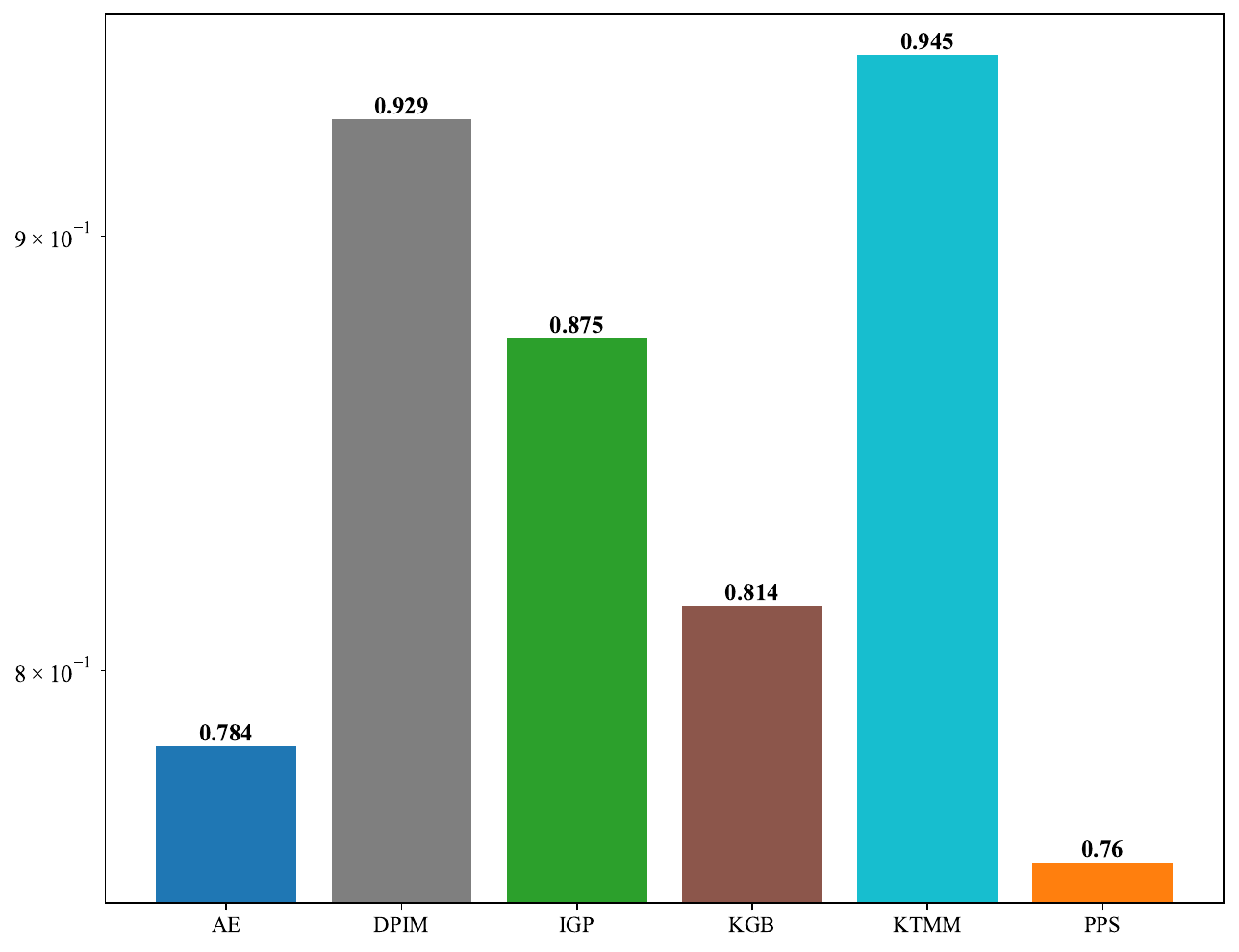}%
        \label{fig:gts_gts3_nsga2_stat_fast_505010_end_igd2_mean_bar}}

    \caption{DMIGD for DF, GTS Group 1, GTS Group 2, and GTS Group 3 test suites.}%
    \label{fig:gts_df_gts123_nsga2_stat_fast_505010_end_igd2_mean_bar}
\end{figure*}

\begin{figure*}[!t]
    \centering
    \subfloat[\scriptsize DF]{\includegraphics[width=0.24\textwidth,height=0.25\textwidth]{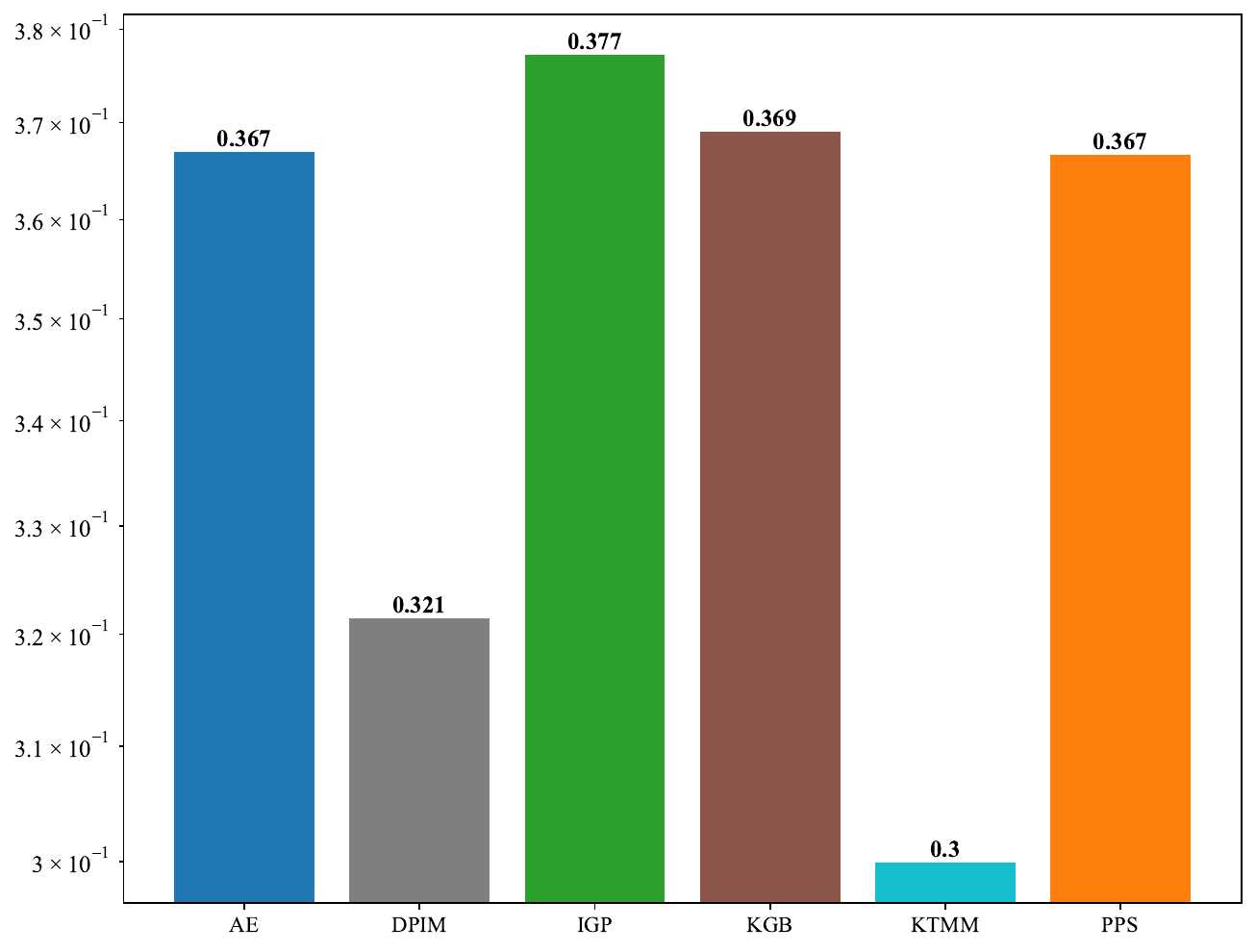}%
        \label{fig:gts_df_nsga2_stat_fast_505010_end_hv2_mean_bar}}
    \hfil
    \subfloat[\scriptsize GTS Group 1]{\includegraphics[width=0.24\textwidth,height=0.25\textwidth]{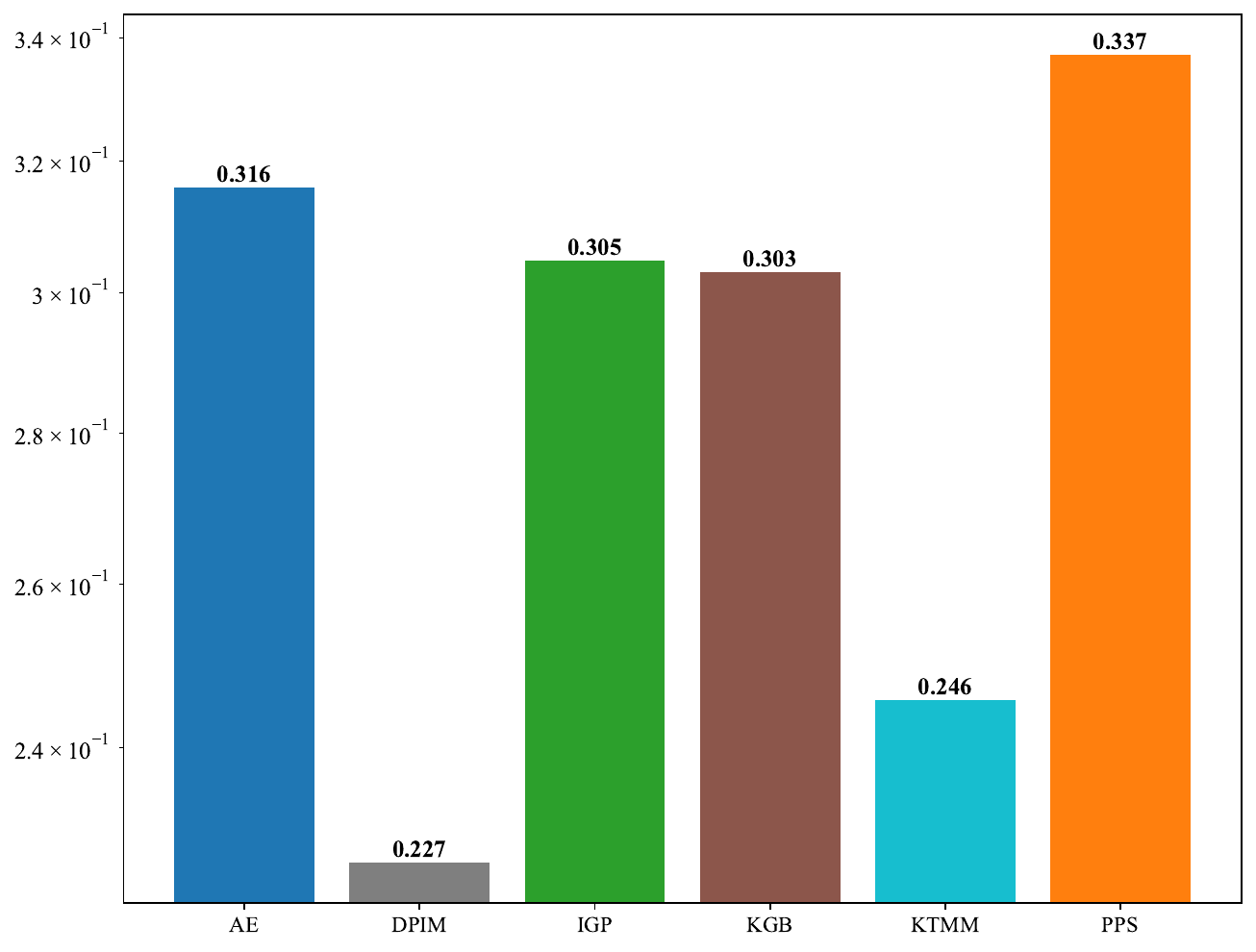}%
        \label{fig:gts_gts1_nsga2_stat_fast_505010_end_hv2_mean_bar}}
    \hfil
    \subfloat[\scriptsize GTS Group 2]{\includegraphics[width=0.24\textwidth,height=0.25\textwidth]{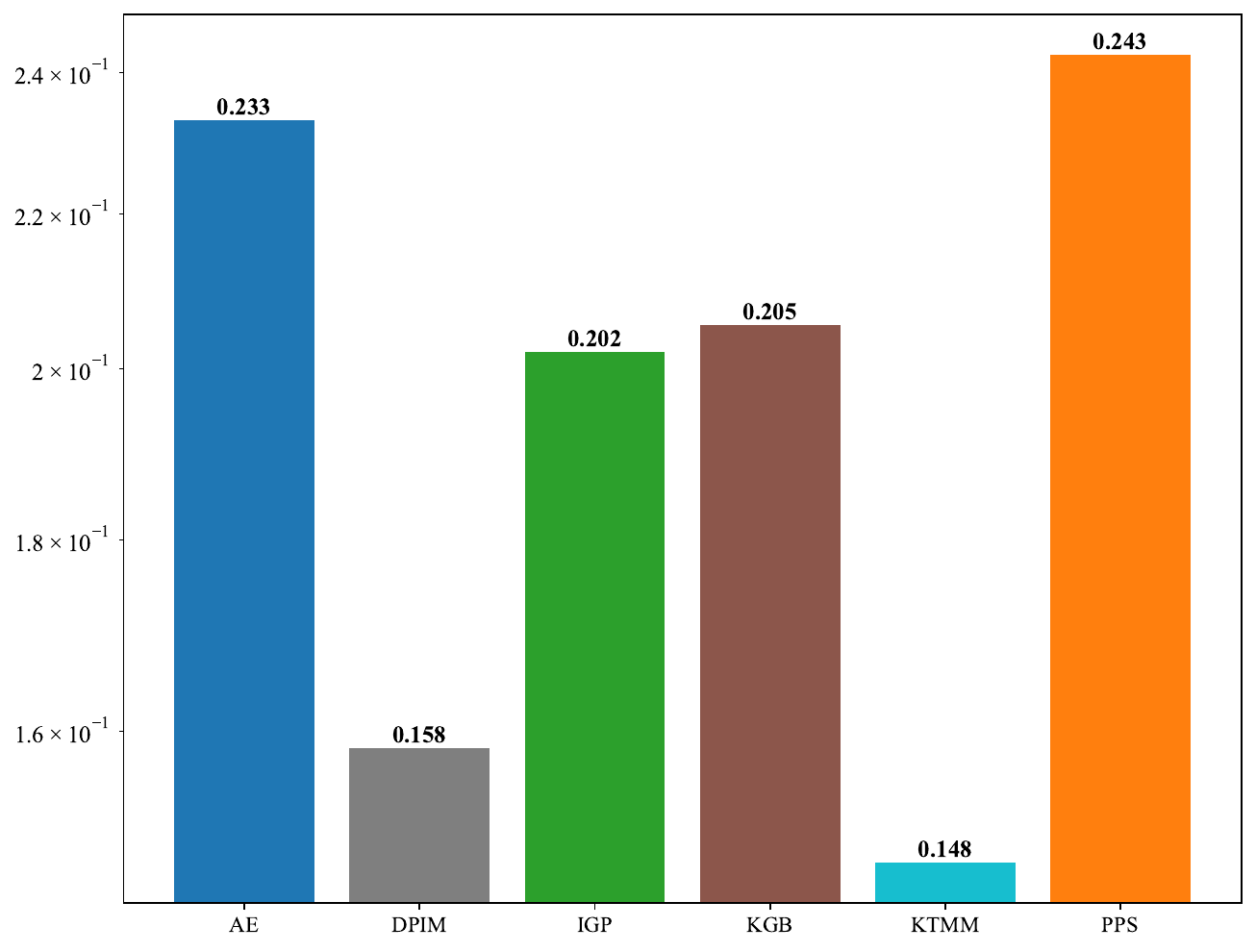}%
        \label{fig:gts_gts2_nsga2_stat_fast_505010_end_hv2_mean_bar}}
    \hfil
    \subfloat[\scriptsize GTS Group 3]{\includegraphics[width=0.24\textwidth,height=0.25\textwidth]{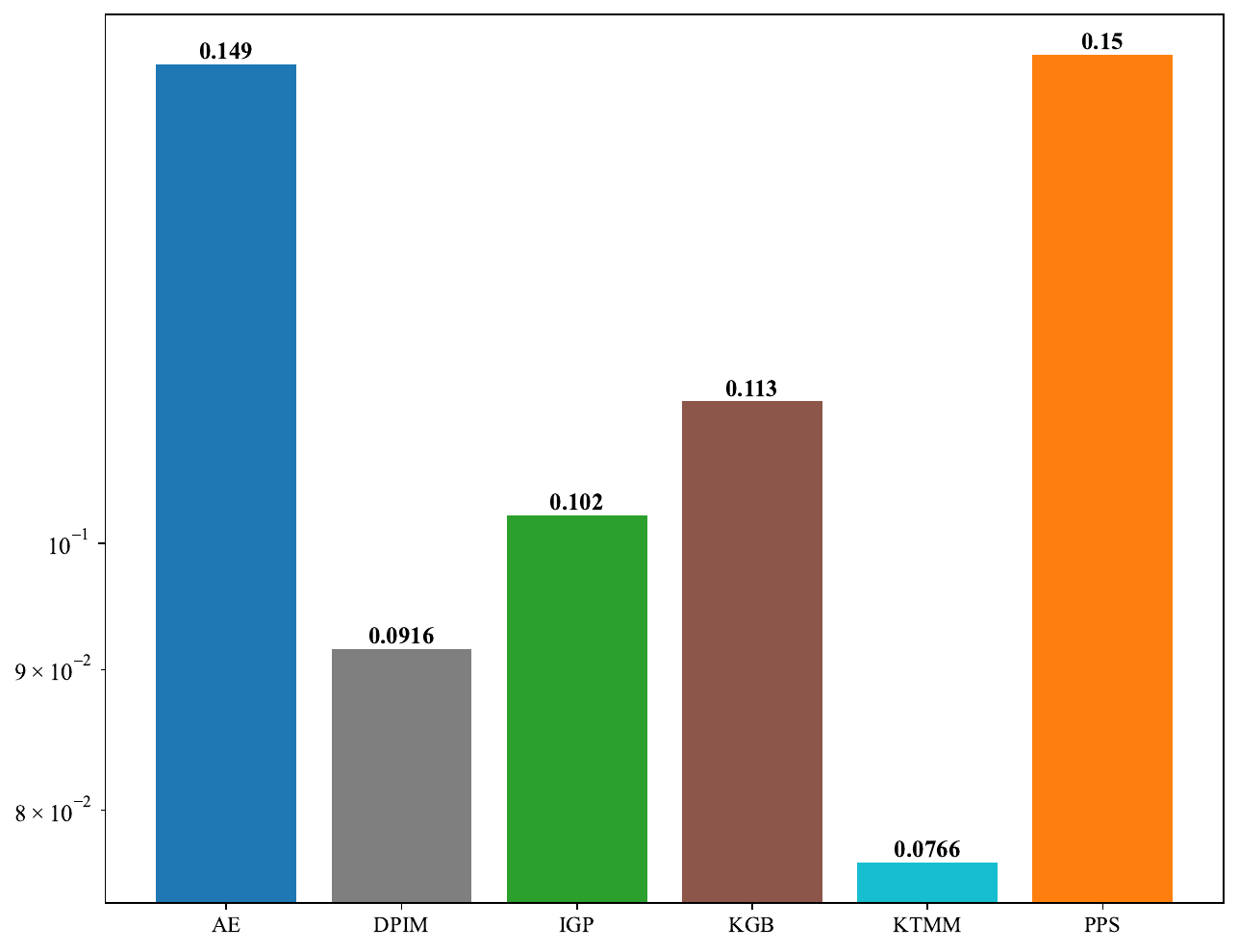}%
        \label{fig:gts_gts3_nsga2_stat_fast_505010_end_hv2_mean_bar}}

    \caption{DMHV for DF, GTS Group 1, GTS Group 2, and GTS Group 3 test suites.}%
    \label{fig:gts_df_gts123_nsga2_stat_fast_505010_end_hv2_mean_bar}
\end{figure*}

\begin{figure*}[!t]
    \centering
    \subfloat[\scriptsize DF]{\includegraphics[width=0.24\textwidth,height=0.25\textwidth]{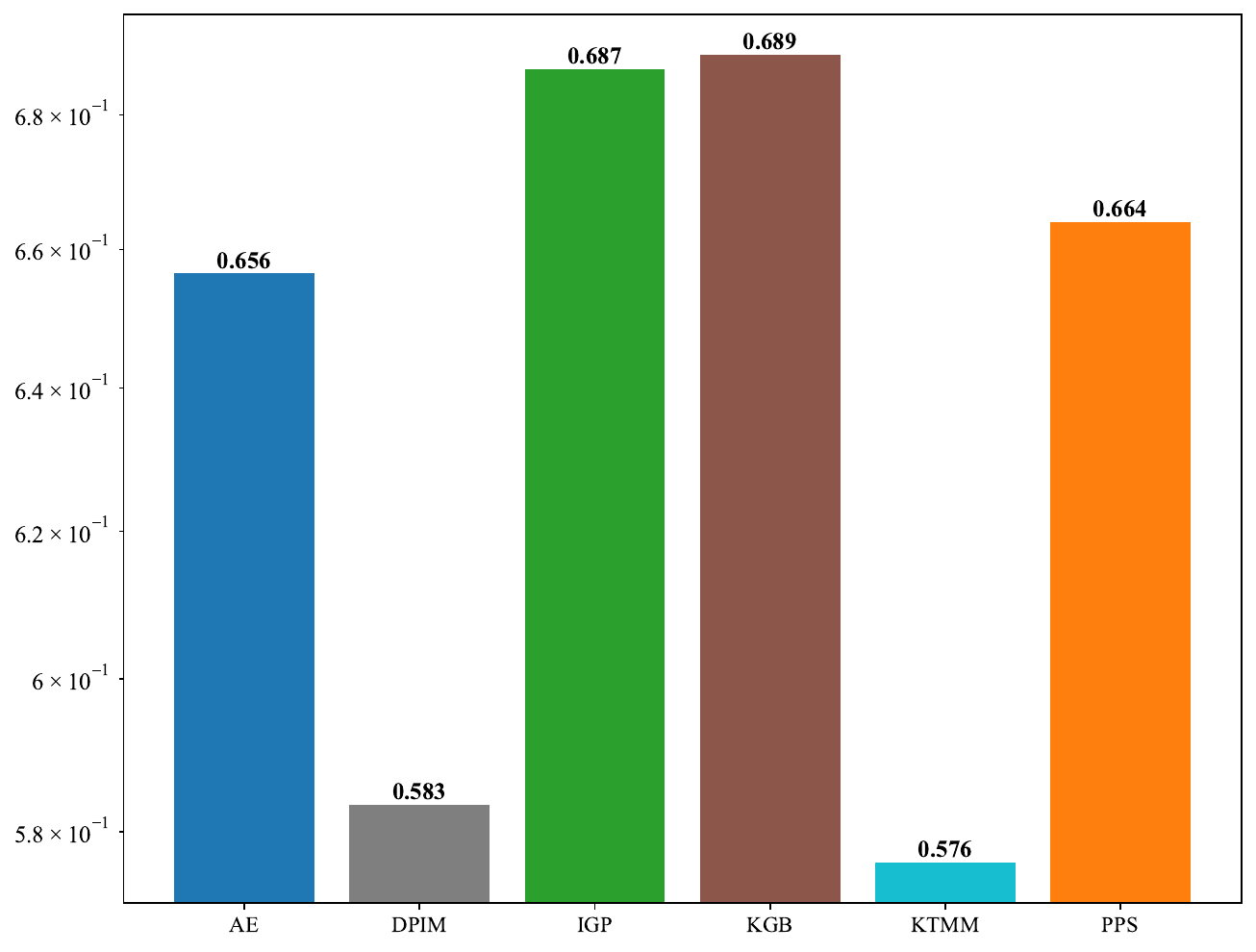}%
        \label{fig:gts_df_nsga2_stat_fast_505010_end_ms2_mean_bar}}
    \hfil
    \subfloat[\scriptsize GTS Group 1]{\includegraphics[width=0.24\textwidth,height=0.25\textwidth]{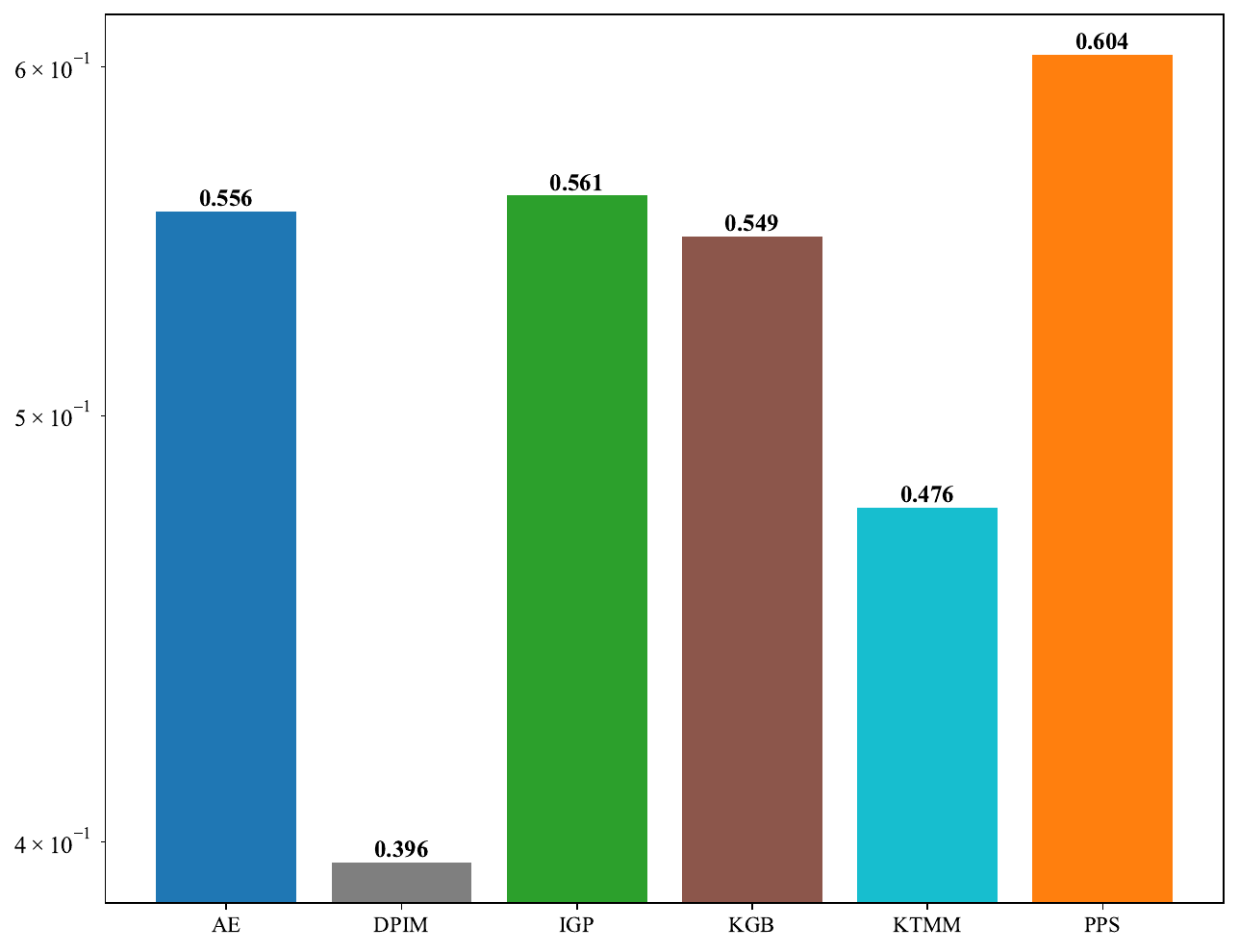}%
        \label{fig:gts_gts1_nsga2_stat_fast_505010_end_ms2_mean_bar}}
    \hfil
    \subfloat[\scriptsize GTS Group 2]{\includegraphics[width=0.24\textwidth,height=0.25\textwidth]{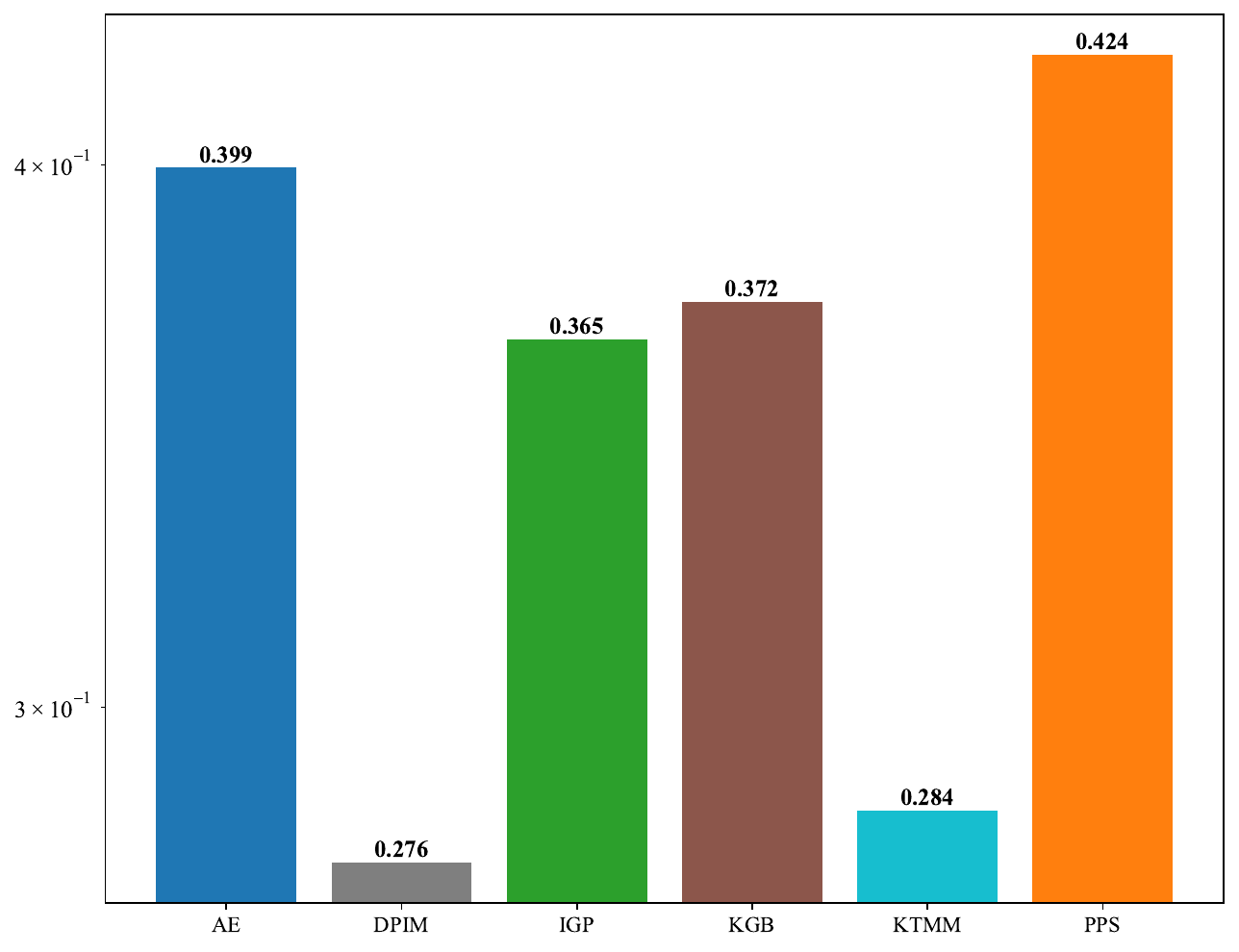}%
        \label{fig:gts_gts2_nsga2_stat_fast_505010_end_ms2_mean_bar}}
    \hfil
    \subfloat[\scriptsize GTS Group 3]{\includegraphics[width=0.24\textwidth,height=0.25\textwidth]{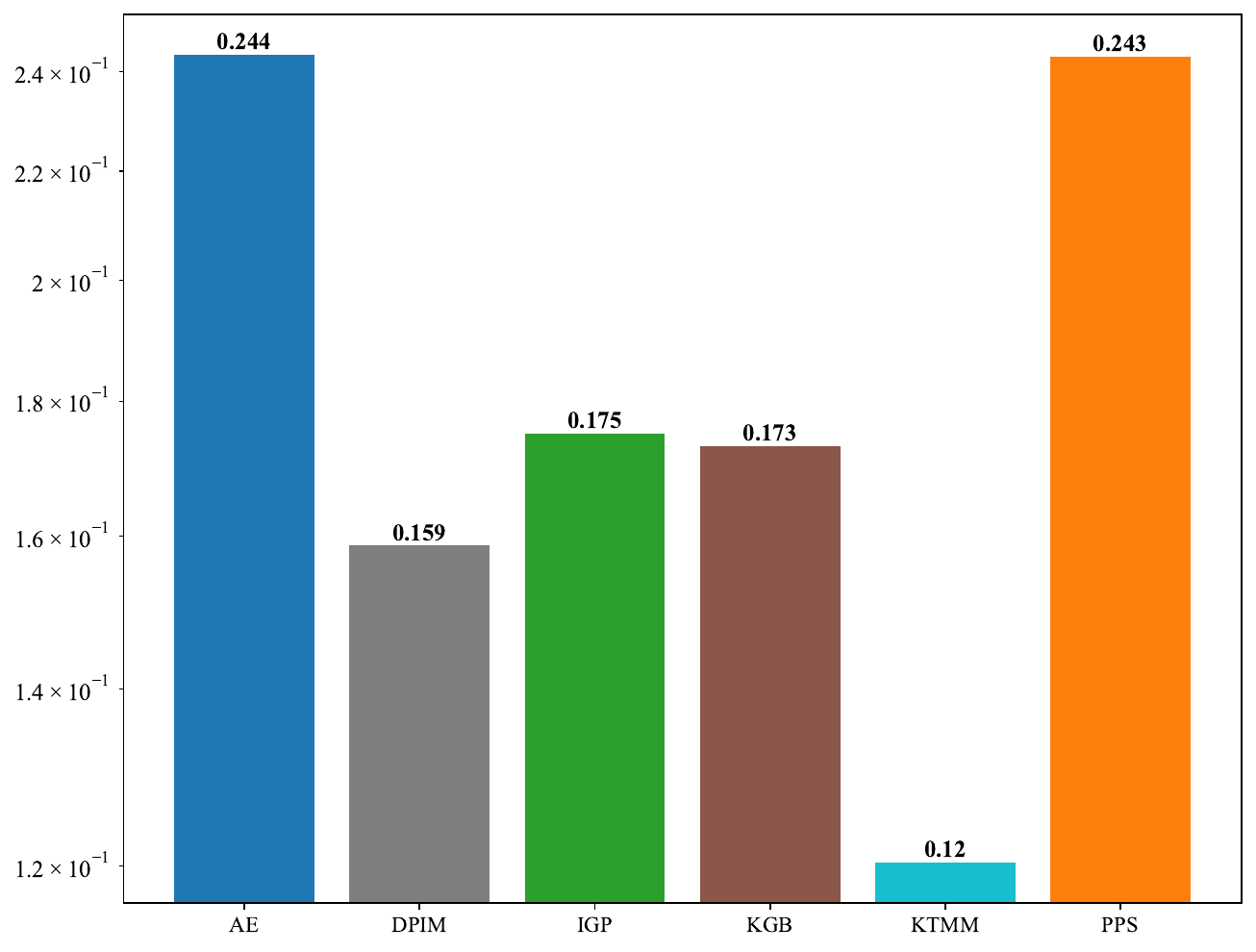}%
        \label{fig:gts_gts3_nsga2_stat_fast_505010_end_ms2_mean_bar}}

    \caption{DMMS for DF, GTS Group 1, GTS Group 2, and GTS Group 3 test suites.}%
    \label{fig:gts_df_gts123_nsga2_stat_fast_505010_end_ms2_mean_bar}
\end{figure*}

In terms of DMIGD, DMHV, and DMMS, the values of these metrics in the DF test suite and the GTS Group 1 test suite are within the same order of magnitude, and the differences between them remain acceptable (see Figures~\ref{fig:gts_df_gts123_nsga2_stat_fast_505010_end_igd2_mean_bar}, ~\ref{fig:gts_df_gts123_nsga2_stat_fast_505010_end_hv2_mean_bar}, and~\ref{fig:gts_df_gts123_nsga2_stat_fast_505010_end_ms2_mean_bar}). However, after introducing variable contribution imbalances, variable interactions, and temporally irregular perturbations, the values for GTS Group 2 and GTS Group 3 exhibit relatively significant changes. For instance, DMIGD and DMHV show a relative increase of about $50\%$ compared to their values before the modifications (see Figures~\ref{fig:gts_gts2_nsga2_stat_fast_505010_end_igd2_mean_bar} and ~\ref{fig:gts_gts3_nsga2_stat_fast_505010_end_igd2_mean_bar}, as well as~\ref{fig:gts_gts2_nsga2_stat_fast_505010_end_hv2_mean_bar} and ~\ref{fig:gts_gts3_nsga2_stat_fast_505010_end_hv2_mean_bar}). Notably, while the runtime of the algorithms does not increase significantly (see Figure~\ref{fig:gts_df_gts123_nsga2_stat_fast_505010_end_runtime_mean_bar}), their performance declines substantially. This reveals a limitation in current algorithm designs: they often struggle to achieve meaningful performance gains on existing benchmarks, leading to a scenario where algorithms excel on standardized tests yet underperform in practical applications. To foster the development of next‑generation intelligent dynamic multi‑objective optimization algorithms and enable equitable comparisons, the GTS test suite is thus both timely and valuable.

\subsubsection{Friedman Rank Test}%
\label{sec:gts_friedman_rank_test}

\begin{figure*}[!t]
    \centering
    \subfloat[\scriptsize DF]{\includegraphics[width=0.24\textwidth,height=0.25\textwidth]{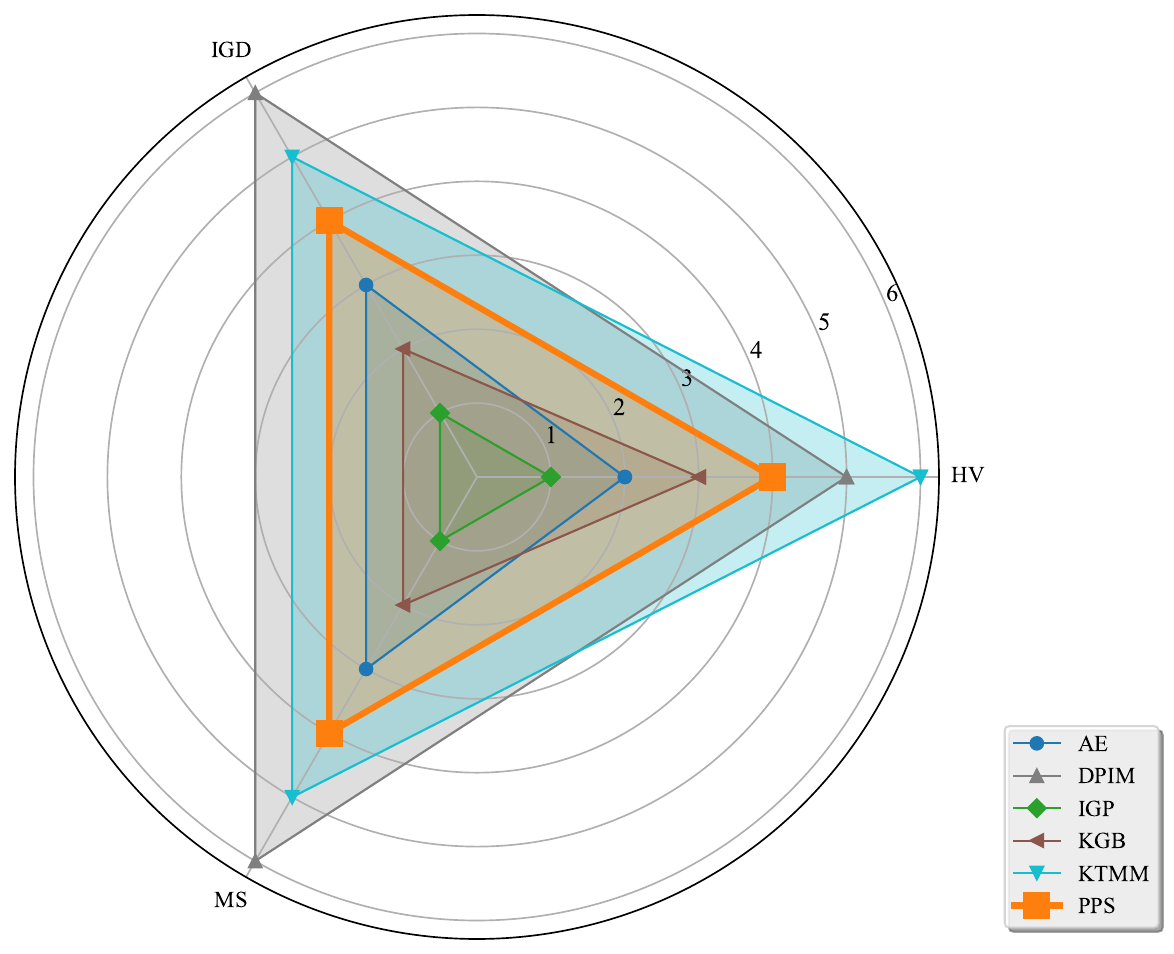}%
        \label{fig:gts_df_nsga2_stat_fast_505010_end_friedman_rank}}
    \hfil
    \subfloat[\scriptsize GTS Group 1]{\includegraphics[width=0.24\textwidth,height=0.25\textwidth]{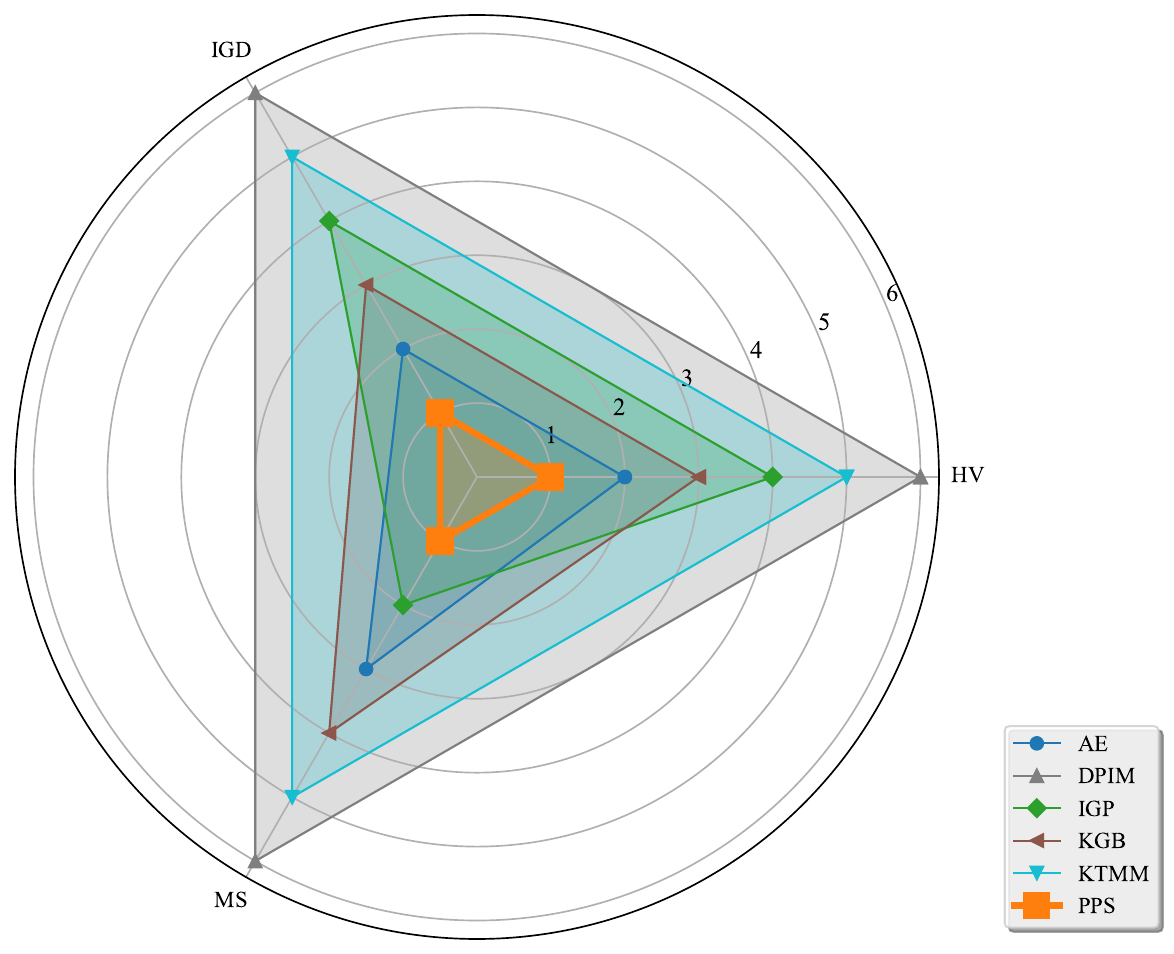}%
        \label{fig:gts_gts1_nsga2_stat_fast_505010_end_friedman_rank}}
    \hfil
    \subfloat[\scriptsize GTS Group 2]{\includegraphics[width=0.24\textwidth,height=0.25\textwidth]{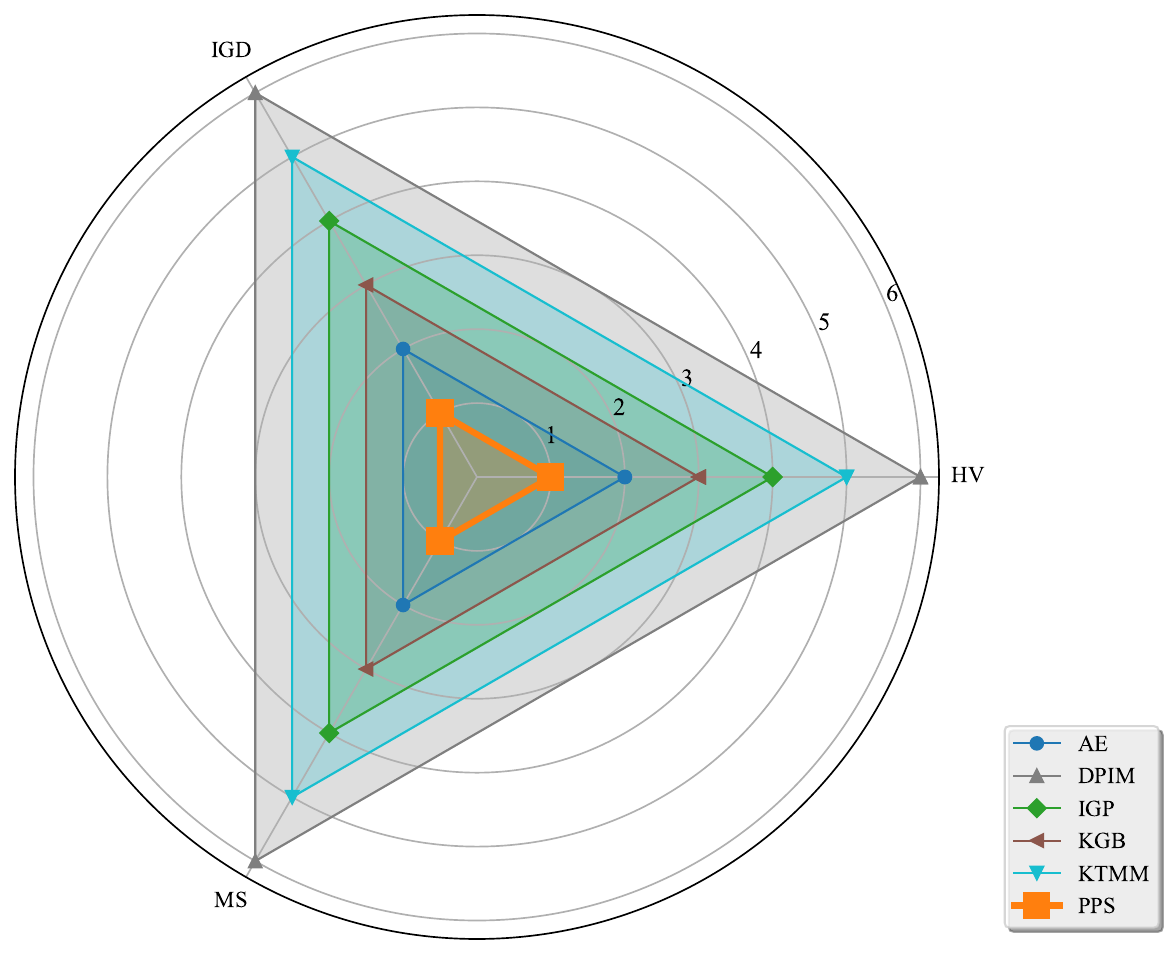}%
        \label{fig:gts_gts2_nsga2_stat_fast_505010_end_friedman_rank}}
    \hfil
    \subfloat[\scriptsize GTS Group 3]{\includegraphics[width=0.24\textwidth,height=0.25\textwidth]{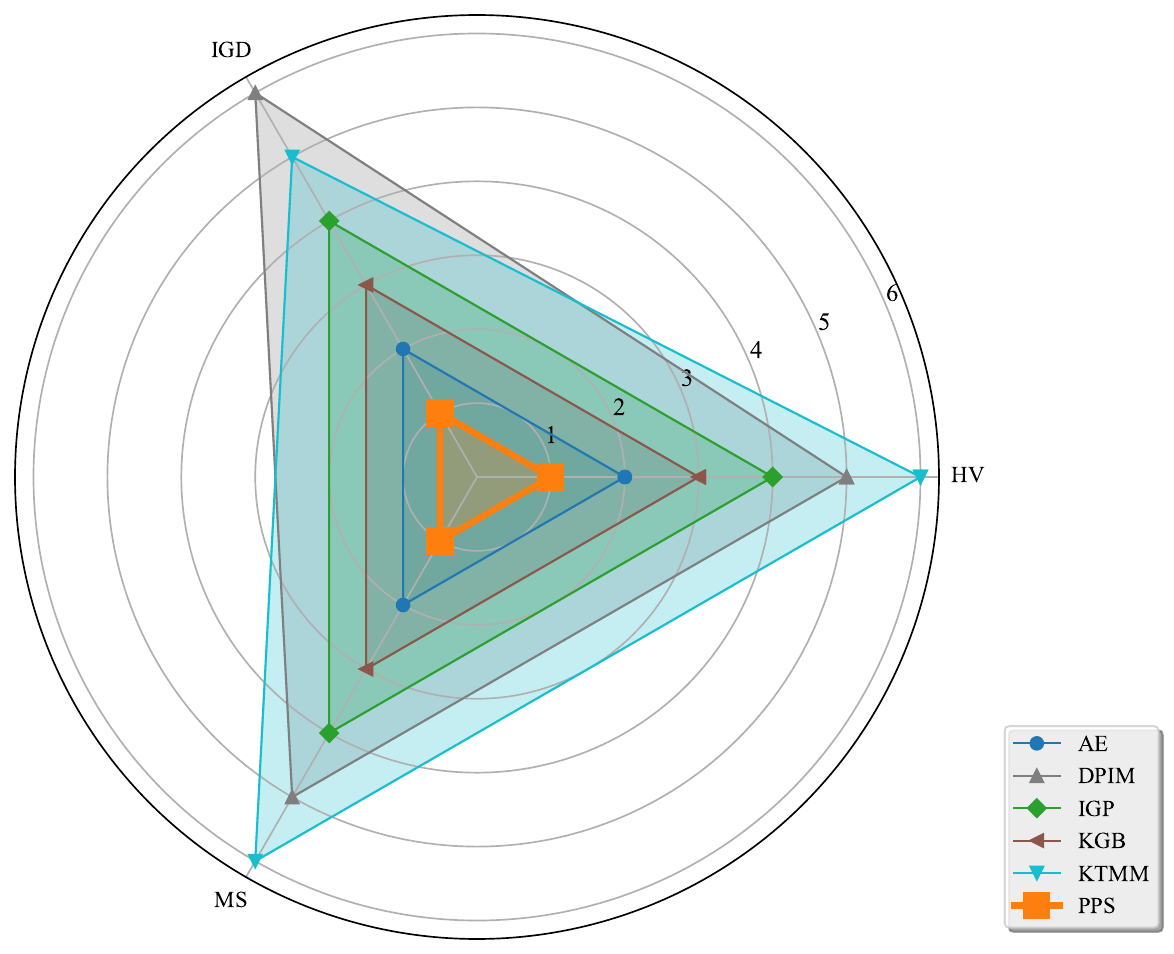}%
        \label{fig:gts_gts3_nsga2_stat_fast_505010_end_friedman_rank}}

    \caption{Friedman Ranking of Algorithms on DF, GTS Group 1, GTS Group 2, and GTS Group 3 test suites.}%
    \label{fig:gts_df_gts123_nsga2_stat_fast_505010_end_friedman_rank}
\end{figure*}

\textbf{DF test suite}: Through the Friedman rank test applied to MIGD, MHV, and MMS, we obtained the following results on the DF test suite (see Figure~\ref{fig:gts_df_nsga2_stat_fast_505010_end_friedman_rank}): the IGP algorithm achieved the best ranking (as shown in Figure~\ref{fig:gts_df_nsga2_stat_fast_505010_end_friedman_rank}), followed by KGB. Subsequently, AE and PPS ranked lower, while KTMM and DPIM obtained relatively poor rankings. In DF, since the majority of PS vary on hyperplanes, IGP effectively fits the inverse mapping from the objective space (2D or 3D) to the decision space through the given sampling mechanism to approximate the next-time PF. Similar trends are reflected in the results, with AE obtaining the third-best ranking.

\textbf{GTS test suites}: However, in the GTS test suites, PPS achieved the top ranking, while AE and KGB ranked second and third, respectively (see Figures~\ref{fig:gts_gts1_nsga2_stat_fast_505010_end_friedman_rank},~\ref{fig:gts_gts2_nsga2_stat_fast_505010_end_friedman_rank},~and~\ref{fig:gts_gts3_nsga2_stat_fast_505010_end_friedman_rank}). In contrast, IGP dropped from first place on DF to fourth place here. KTMM and DPIM again occupied the lowest rankings. PPS first predicts the central point for the next time using historical information, employing a univariate autoregression (AR) model for each decision variable. This enables it to effectively handle PS variations on hypersurfaces, and subsequently, it approximates the next time PS by incorporating manifold predictions. The decline in IGP’s ranking to fourth place is attributed to the temporally irregular perturbations introduced in GTS Groups, which challenge the simple sampling mechanism used in the IGP study for predicting the next-time PF. In other words, while IGP’s sampling mechanism performs well under regular changes, it struggles to adapt to the irregular variations incorporated in GTS.

\textbf{Conclusion}: Based on the experimental results across the three test suites, the GTS benchmark demonstrates significant advantages in comprehensively evaluating the capabilities of DMOAs. Firstly, by introducing temporally irregular perturbations in Group 1, it effectively reveals the limitations of algorithms like IGP that rely on simple sampling mechanisms under regular changes, while highlighting the superiority of prediction-based methods such as PPS in handling complex hypersurface variations. Secondly, the incorporation of controlled variable contribution imbalances and dynamic variable interactions in Groups 2 and 3 successfully distinguishes algorithms with specific methodological strengths, such as KGB's robust performance through knowledge-guided Bayesian classification in handling complex variable relationships. Most importantly, the hierarchical structure of GTS enables a more systematic assessment of algorithm adaptability, not only validating performance under different dynamic characteristics but also exposing trade-offs between solution quality and computational efficiency, thereby providing valuable guidance for both practical applications and methodological development.

\subsection{Running Time Analysis}

In this subsection, we first present a comparison of the running time between the DF and GTS test suites. This is followed by an internal analysis of the computational time across different groups in the GTS. Finally, we discuss a notable observation related to the KGB algorithm.

\subsubsection{Comparison of Runtime between DF and GTS}

Figures~\ref{fig:gts_df_nsga2_stat_fast_505010_end_runtime_mean_bar} illustrate that, for most algorithms, the mean runtime is less than $16$ seconds and consistent across different test suites, with a variation of less than $10\%$ between DF and GTS, for both two and three objective functions under the given experimental conditions. Due to the page limit, the detailed running time data can be found in the Supplementary Materials (see Section II).

\subsubsection{Comparison of Runtime among GTS Groups}

As demonstrated in Figures~\ref{fig:gts_gts1_nsga2_stat_fast_505010_end_runtime_mean_bar},~\ref{fig:gts_gts2_nsga2_stat_fast_505010_end_runtime_mean_bar}, and~\ref{fig:gts_gts3_nsga2_stat_fast_505010_end_runtime_mean_bar}, the choice of different symmetric positive semidefinite matrices $\mathbf{R}_{II,1}$ and $\mathbf{R}_{II,2}$ configurations exhibits negligible impact on the algorithm's runtime performance. Complementary evidence from Figures~\ref{fig:gts_df_gts123_nsga2_stat_fast_505010_end_igd2_mean_bar}, ~\ref{fig:gts_df_gts123_nsga2_stat_fast_505010_end_hv2_mean_bar}, and~\ref{fig:gts_df_gts123_nsga2_stat_fast_505010_end_ms2_mean_bar} further reveals that these matrix variations primarily influence PS/PF quality metrics rather than computational efficiency. Specifically, lower values of both MHV and MMS, as well as a higher MIGD value, are observed with increasing complexity of $\mathbf{R}_{II,1}$ and $\mathbf{R}_{II,2}$. The observed runtime stability (see Figure~\ref{fig:gts_gts1_nsga2_stat_fast_505010_end_runtime_mean_bar}) suggests that the algorithm's complexity is robust to the specified structural modifications in the $\mathbf{R}_{II, 1}$ and $\mathbf{R}_{II, 2}$ matrices.

\subsubsection{Notable Observation for KGB}

\begin{figure*}[!t]
    \centering
    \subfloat[\scriptsize DF]{\includegraphics[width=0.24\textwidth,height=0.25\textwidth]{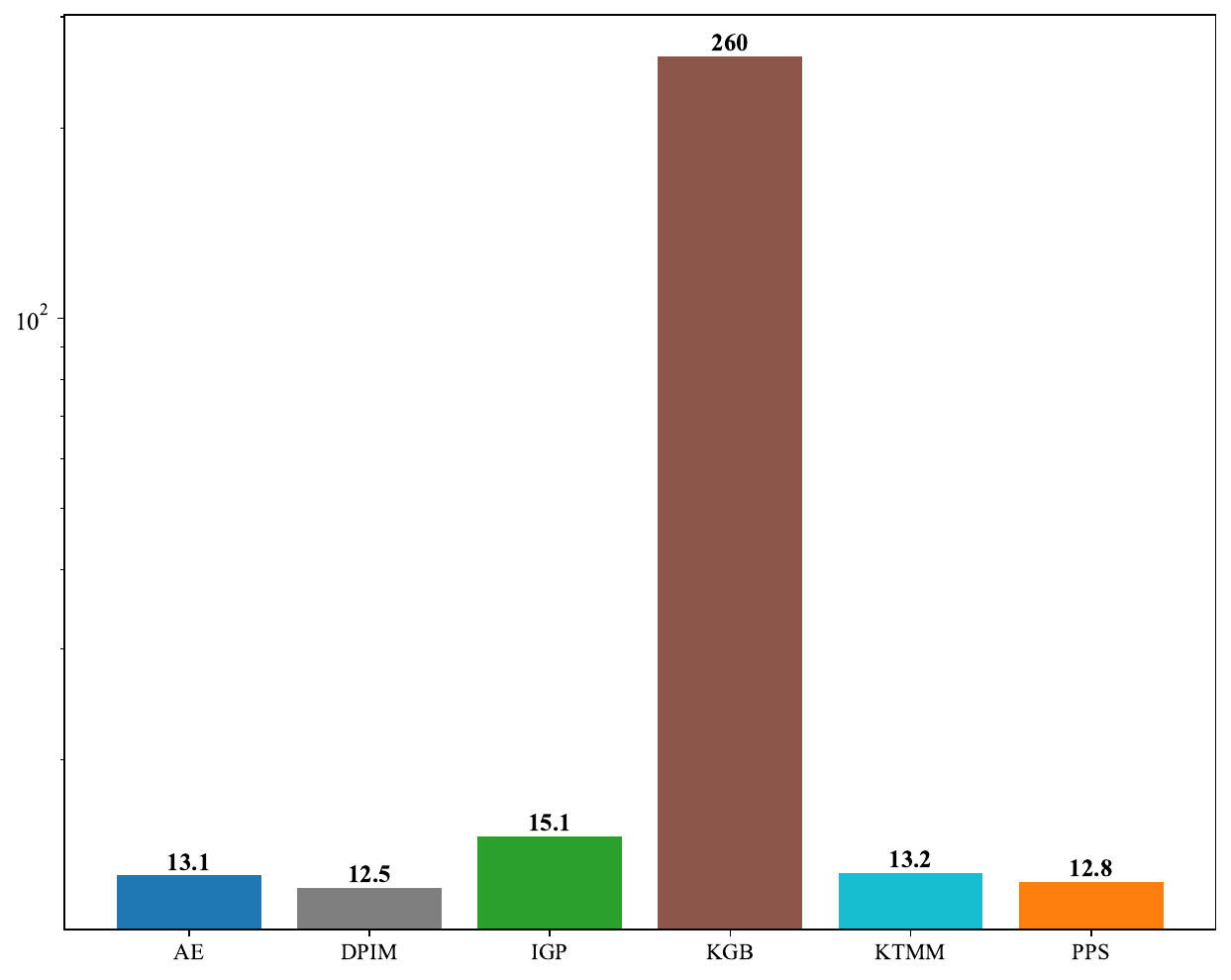}%
        \label{fig:gts_df_nsga2_stat_fast_505010_end_runtime_mean_bar}}
    \hfil
    \subfloat[\scriptsize GTS Group 1]{\includegraphics[width=0.24\textwidth,height=0.25\textwidth]{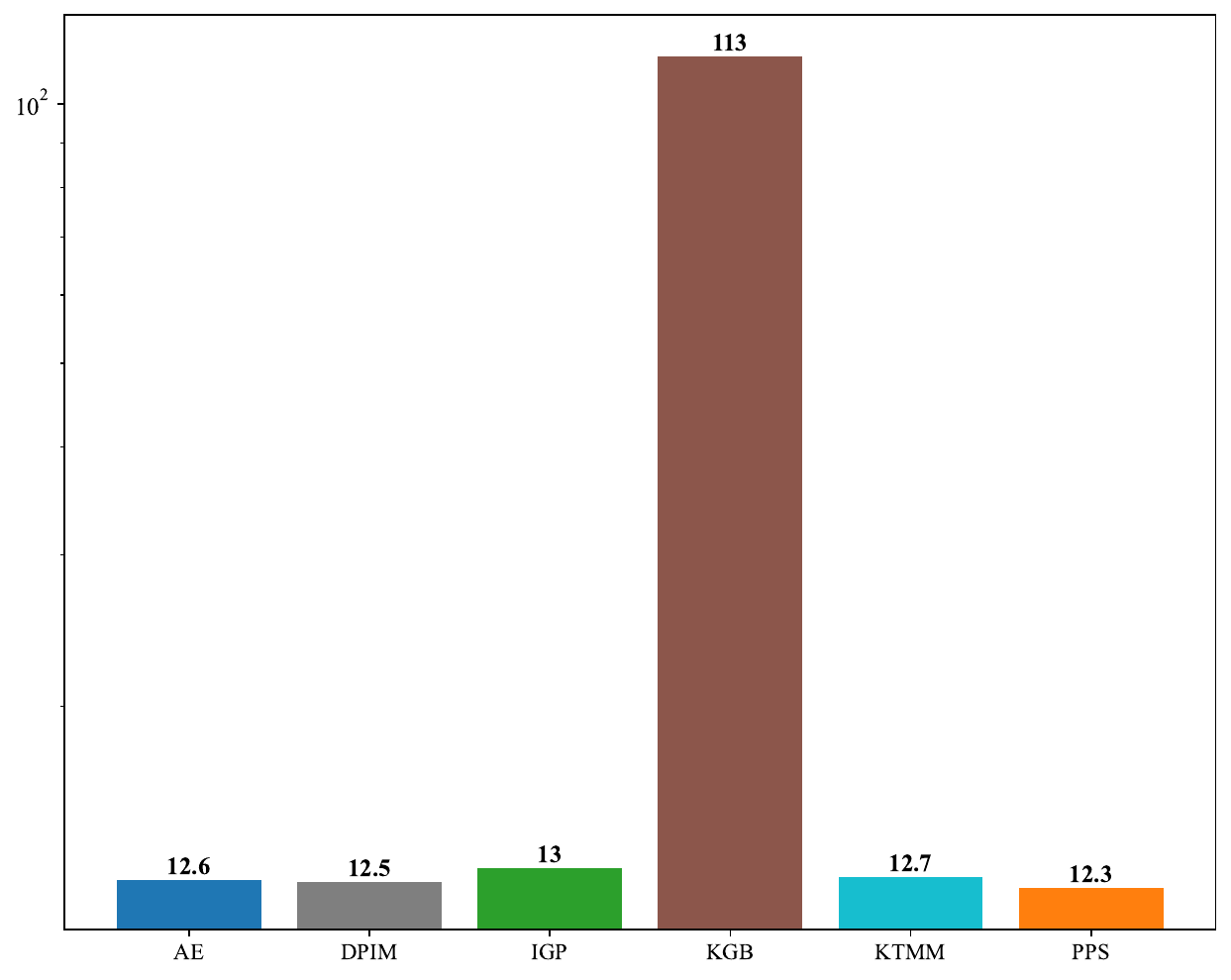}%
        \label{fig:gts_gts1_nsga2_stat_fast_505010_end_runtime_mean_bar}}
    \hfil
    \subfloat[\scriptsize GTS Group 2]{\includegraphics[width=0.24\textwidth,height=0.25\textwidth]{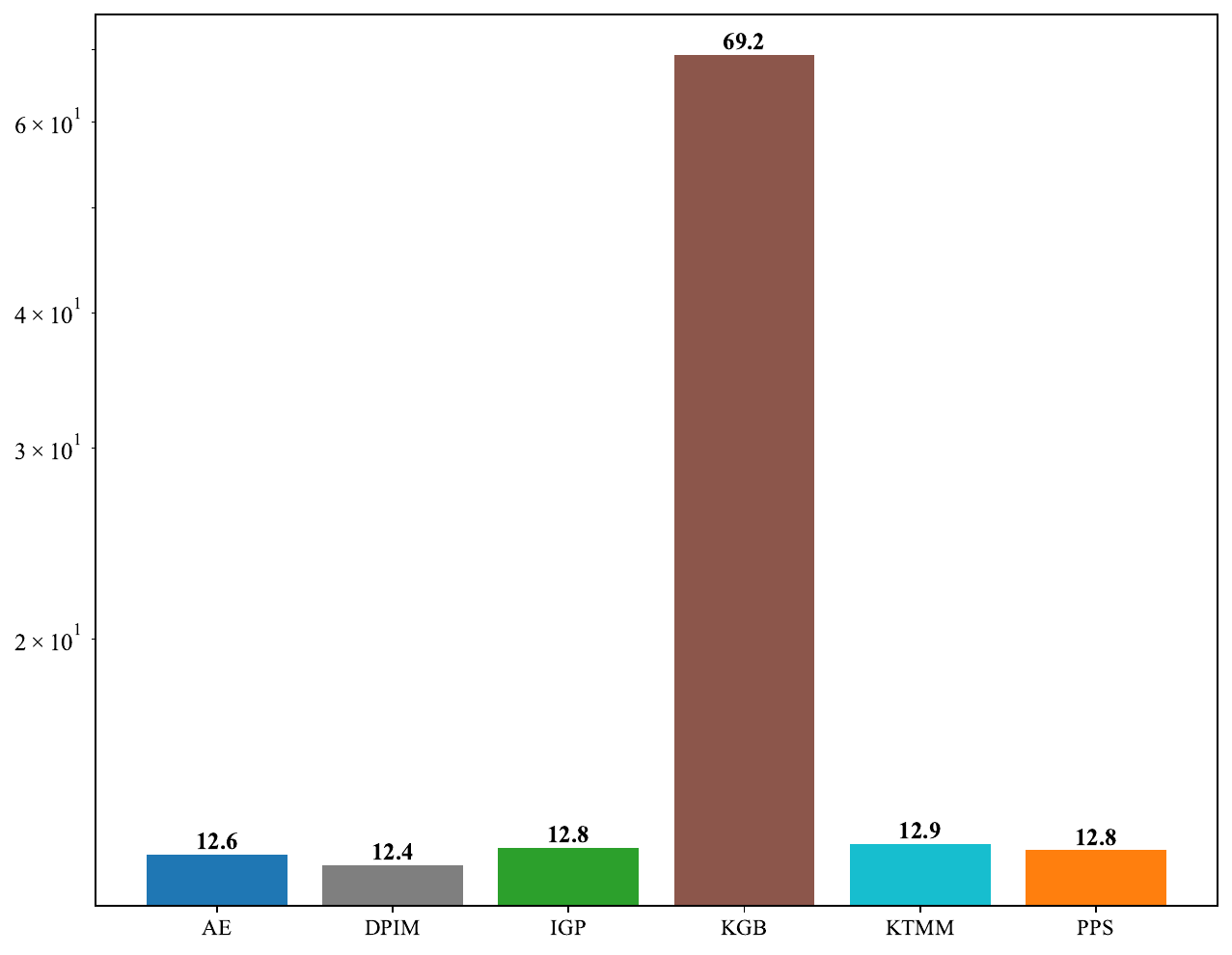}%
        \label{fig:gts_gts2_nsga2_stat_fast_505010_end_runtime_mean_bar}}
    \hfil
    \subfloat[\scriptsize GTS Group 3]{\includegraphics[width=0.24\textwidth,height=0.25\textwidth]{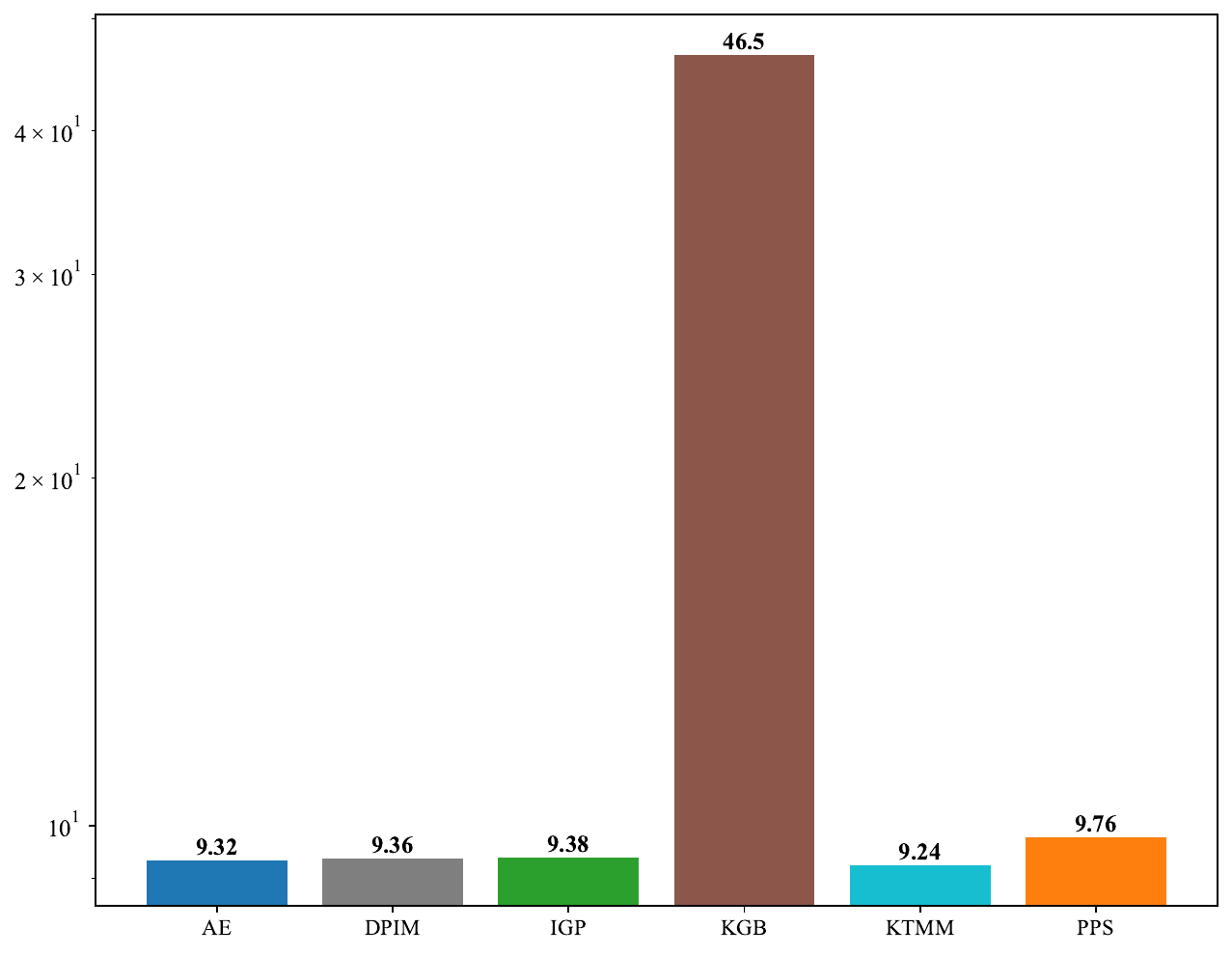}%
        \label{fig:gts_gts3_nsga2_stat_fast_505010_end_runtime_mean_bar}}

    \caption{Averaged runtime of all algorithms across all trials in the DF and GTS test suite.}%
    \label{fig:gts_df_gts123_nsga2_stat_fast_505010_end_runtime_mean_bar}
\end{figure*}

For the KGB algorithm, mean runtime varies significantly (Figures~\ref{fig:gts_df_nsga2_stat_fast_505010_end_runtime_mean_bar} and~\ref{fig:gts_df_gts123_nsga2_stat_fast_505010_end_runtime_mean_bar}), with DF taking approximately $260$ seconds, GTS group 1 about $110$ seconds, and both GTS group 2 and GTS group 3 around $70$ and $45$ seconds. This result arises primarily from variable independence: in DF and GTS group 1, most variables are independent, forcing the clustering algorithm to expend considerable effort in grouping them. In GTS2 and GTS3, however, variable interactions and imbalances are introduced, enabling faster and more efficient convergence. Consequently, KGB may be ill-suited for large-scale or real-world problems with predominantly independent variables. Moreover, on the DF and GTS test suites, KGB consistently demands more time for response and is generally slower than the other compared algorithms.

\section{Conclusion}%
\label{sec:gts_conclusion}

Based on the theoretical analysis and experimental results presented in this paper, we conclude that the proposed benchmark construction framework successfully addresses key limitations in existing DMOO test suites. It introduces novel and diverse components that advance theoretical understanding and enhance practical relevance. Through the integration of generalized PS/PF dynamics, controlled variable contribution imbalances, dynamic variable interactions, temporally irregular perturbations, and a time-linkage mechanism, the framework provides a more principled and systematic approach for generating DMOPs with configurable complexity. Experimental results demonstrate that these components collectively enable a more rigorous assessment of algorithm performance, particularly in tracking evolving PS/PF under heterogeneous landscapes and complex temporal dynamics.

The framework's modular architecture offers significant flexibility for constructing test problems that capture diverse real-world challenges while maintaining mathematical tractability. By enabling precise control over landscape characteristics, variable interactions, and temporal patterns, the proposed methodology establishes a new standard for dynamic multi-objective optimization benchmarking. The publicly available implementation\footnote{\url{https://github.com/dynoptimization/pydmoo}} facilitates reproducibility and further research advancement. Future work will focus on extending the framework to handle additional dynamic characteristics and exploring its application to more complex real-world optimization scenarios or high-dimensional cases.

\section*{Acknowledgement}

This work is supported by the National Natural Science Foundation of China under the Grant No.61761136008, the National Natural Science Foundation of China under Grant No.72401122, Guangdong Basic and Applied Basic Research Foundation under Grants No. 2024A1515012241 and 2021A1515110024.

\appendix

\section{Generalized Test Suite}%
\label{sec:gts_supp_detailed_problems}

In this section, we provide a detailed formulation of each benchmarking function in the GTS, including its corresponding parameters as well as the PS and PF. The Python code is publicly available at our companion website \href{https://github.com/dynoptimization/pydmoo}{https://github.com/dynoptimization/pydmoo}. The common/shared parameters used in the GTS are summarized in Table~\ref{tab:gts_supp_symboles_gts}.

\begin{table}[h]
    \centering
    \caption{Symbols used in the GTS}%
    \label{tab:gts_supp_symboles_gts}
    \begin{tabular}{llll}
        \toprule
        $G(t) = \sin(0.5\pi   t)$        & $H(t) = 1.5 + G(t)$                 & $\alpha_t = 5\cos(0.5\pi t)$ & \\
        \midrule
        $\beta_t = 0.2 + 2.8\abs{G(t)}$  & $\omega_t = \lfloor 10G(t) \rfloor$ & $a_t = \sin(0.5\pi t)$       & \\
        \midrule
        $b_t = 1 + \abs{\cos(0.5\pi t)}$ & $y_t = 0.5 + G(t)(x_1 - 0.5)$       &                              & \\
        \bottomrule
    \end{tabular}
\end{table}

\begin{figure}[htbp]
    \centering
    \begin{minipage}[t]{0.45\textwidth}
        \centering
        \includegraphics[width=\linewidth]{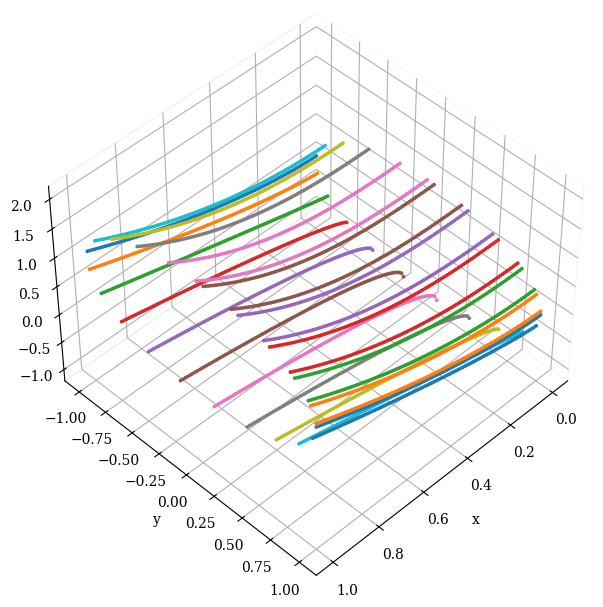}
        \caption{PS of GTS1 problem}%
        \label{fig:gts_supp_GTS1_ps}
    \end{minipage}
    \hfill
    \begin{minipage}[t]{0.45\textwidth}
        \centering
        \includegraphics[width=\linewidth]{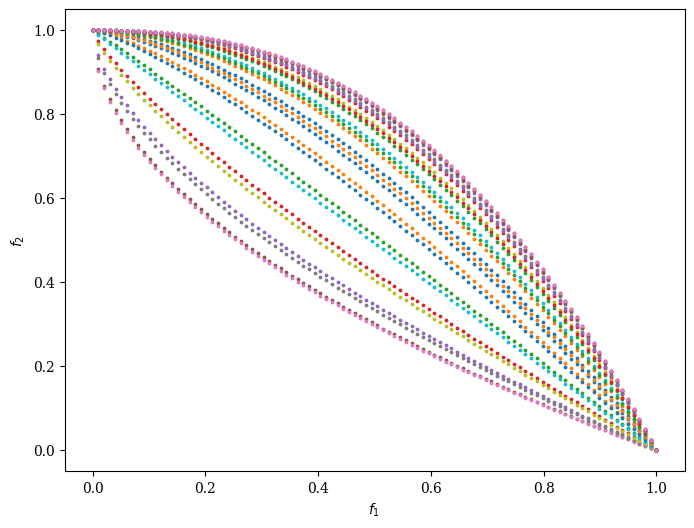}
        \caption{PF of GTS1 problem}%
        \label{fig:gts_supp_GTS1_pf}
    \end{minipage}
\end{figure}

\begin{figure}[htbp]
    \centering
    \begin{minipage}[t]{0.45\textwidth}
        \centering
        \includegraphics[width=\linewidth]{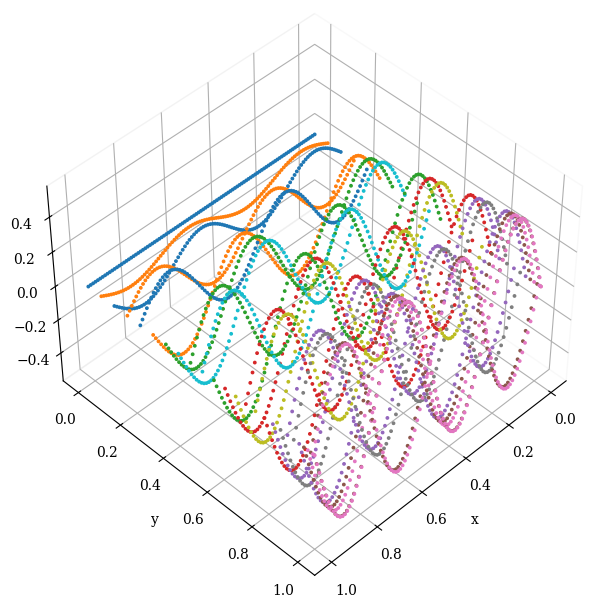}
        \caption{PS of GTS4 problem}%
        \label{fig:gts_supp_GTS4_ps}
    \end{minipage}
    \hfill
    \begin{minipage}[t]{0.45\textwidth}
        \centering
        \includegraphics[width=\linewidth]{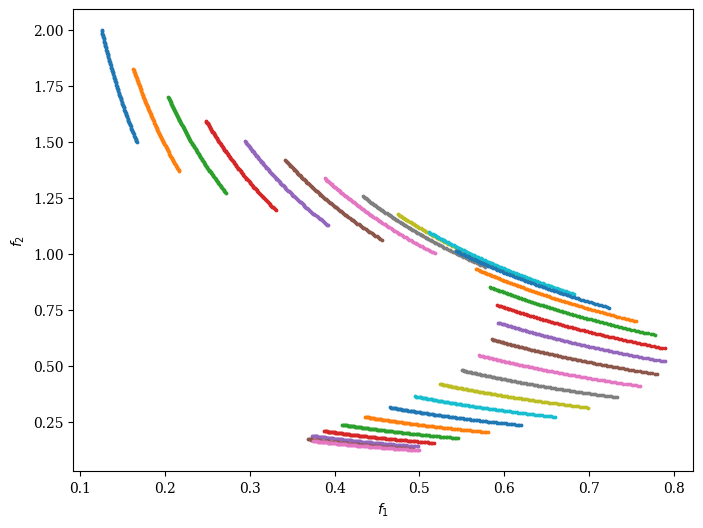}
        \caption{PF of GTS4 problem}%
        \label{fig:gts_supp_GTS4_pf}
    \end{minipage}
\end{figure}

\begin{figure}[htbp]
    \centering
    \begin{minipage}[t]{0.45\textwidth}
        \centering
        \includegraphics[width=\linewidth]{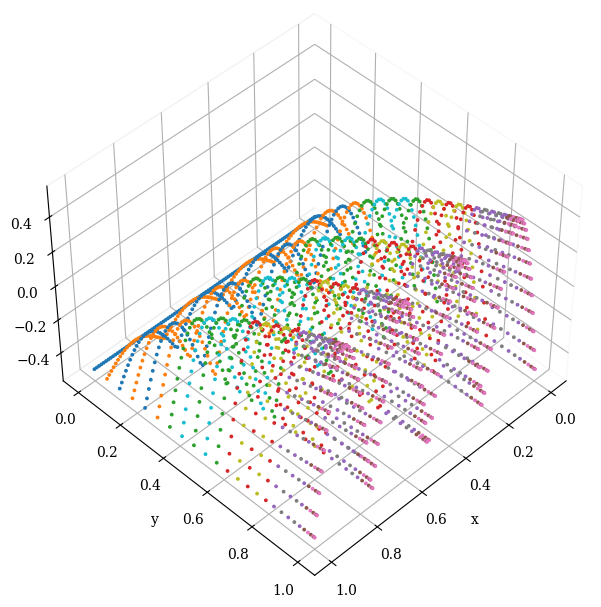}
        \caption{PS of GTS10 problem}%
        \label{fig:gts_supp_GTS10_ps}
    \end{minipage}
    \hfill
    \begin{minipage}[t]{0.45\textwidth}
        \centering
        \includegraphics[width=\linewidth]{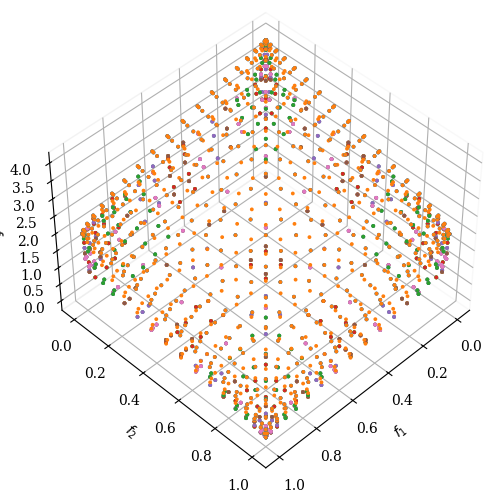}
        \caption{PF of GTS10 problem}%
        \label{fig:gts_supp_GTS10_pf}
    \end{minipage}
\end{figure}

In the provided figures (such as Figures~\ref{fig:gts_supp_GTS1_ps},~\ref{fig:gts_supp_GTS4_ps}, and~\ref{fig:gts_supp_GTS10_ps}) illustrating the PSs, the variables $x$, $y$, and $z$ represent components from the first, second, and third parts of the decision variable vector, respectively, following a specific decomposition. For instance, in the GTS1 problem, the variable vector is partitioned such that the first part is $\mathbf{x}_I = (x_1)$, the second part is $\mathbf{x}_{II,1}=(x_2, \cdots, x_{\lfloor\frac{D}{2}\rfloor})$, and the third part is $\mathbf{x}_{II,2} = (x_{\lfloor\frac{D}{2}\rfloor + 1}, \cdots, x_D)$, where $D$ is the dimensionality of decision variable. The PS is constructed under the condition that all variables within $\mathbf{x}_{II,1}$ share an identical value $h_1(\mathbf{x}_I)$, and similarly, all variables within $\mathbf{x}_{II,2}$ share another identical value $h_2(\mathbf{x}_I)$. Consequently, $x$ represents the variable $x_1$ from $\mathbf{x}_I$, while $y$ and $z$ denote arbitrary components selected from the variable groups $\mathbf{x}_{II,1}$ and $\mathbf{x}_{II,2}$, respectively. Note that the conventional approach constructs the PS by assigning identical values to $\mathbf{x}_{II,1}$ and $\mathbf{x}_{II,2}$, restricting it to a hyperplane. In contrast, our approach, based on a symmetric design, enables the PS to vary over a hypersurface, thereby providing a more realistic characterization of real-world problems.

In the provided figures (such as Figures~\ref{fig:gts_supp_GTS1_pf},~\ref{fig:gts_supp_GTS4_pf}, and~\ref{fig:gts_supp_GTS10_pf}) depicting the PFs, $f_1$, $f_2$, and $f_3$ represent the first, second, and third objective functions, respectively. Note that in (dynamic) multi-objective optimization community, the number of objective functions is typically limited to two or three. Problems with more than three objectives, known as (dynamic) many-objective optimization~\cite{C_CEC_hughes2005evolutionary}, are not considered in this study.

\subsection{GTS1}

\begin{equation}
    \text{min}
    \begin{cases}
        f_1(\mathbf{x},t) = x_1 \\
        f_2(\mathbf{x},t) = g(\mathbf{x},t)(1 - (\frac{x_1}{g(\mathbf{x},t)})^{H(t)})
    \end{cases}
\end{equation}
with
\begin{equation*}
    \begin{split}
        g(\mathbf{x},t) = 1
         & + \Bigl(\bigl(\mathbf{x}_{II,1} - h_1(\mathbf{x}_I)\bigr)^T \mathbf{R}_{II,1}(t) \bigl(\mathbf{x}_{II,1} - h_1(\mathbf{x}_I)\bigr)\Bigr)^{\frac{1}{p}} \\
         & + \Bigl(\bigl(\mathbf{x}_{II,2} - h_2(\mathbf{x}_I)\bigr)^T \mathbf{R}_{II,2}(t) \bigl(\mathbf{x}_{II,2} - h_2(\mathbf{x}_I)\bigr)\Bigr)^{\frac{1}{p}}
    \end{split}
\end{equation*}
where $p \geq 1$, $\mathbf{x}_I = (x_1)$, $\mathbf{x}_{II,1} = (x_2, \cdots, x_{\lfloor\frac{D}{2}\rfloor})$ and $\mathbf{x}_{II,2} = (x_{\lfloor\frac{D}{2}\rfloor + 1}, \cdots, x_D)$,
$h_1(\mathbf{x}_I, t) = \cos(0.5\pi t)$ and $h_2(\mathbf{x}_I, t) = G(t) + x_1^{H(t)}$,
$\mathbf{R}_{II,1}(t)$ and $\mathbf{R}_{II,2}(t)$ are symmetric positive semidefinite matrices in the $t$-th environment,
the search space is $[0,1] \times [-1,1]^{\lfloor\frac{D}{2}\rfloor -1} \times [-1, 2]^{\lceil\frac{D}{2}\rceil}$.

The PS and PF at time t can be described as:
\begin{equation*}
    \begin{aligned}
         & \text{PS(t): }0 \leq x_1 \leq 1, x_i = h_1(\mathbf{x}_I, t) \in \mathbf{x}_{II,1}, x_j = h_2(\mathbf{x}_I, t) \in \mathbf{x}_{II,2} \\
         & \text{PF(t): a part of }f_2 = 1 - f_1^{H(t)}, 0 \leq f_1 \leq 1
    \end{aligned}
\end{equation*}

\subsection{GTS2}

\begin{equation}
    \text{min}
    \begin{cases}
        f_1(\mathbf{x},t) = 0.5x_1+x_2 \\
        f_2(\mathbf{x},t) = g(\mathbf{x},t)(2.8 - (\frac{0.5x_1+x_2}{g(\mathbf{x},t)})^{H(t)})
    \end{cases}
\end{equation}
with
\begin{equation*}
    \begin{split}
        g(\mathbf{x},t) = 1
         & + \Bigl(\bigl(\mathbf{x}_{II,1} - h_1(\mathbf{x}_I)\bigr)^T \mathbf{R}_{II,1}(t) \bigl(\mathbf{x}_{II,1} - h_1(\mathbf{x}_I)\bigr)\Bigr)^{\frac{1}{p}} \\
         & + \Bigl(\bigl(\mathbf{x}_{II,2} - h_2(\mathbf{x}_I)\bigr)^T \mathbf{R}_{II,2}(t) \bigl(\mathbf{x}_{II,2} - h_2(\mathbf{x}_I)\bigr)\Bigr)^{\frac{1}{p}}
    \end{split}
\end{equation*}
where $p \geq 1$, $\mathbf{x}_I = (x_1, x_2)$, $\mathbf{x}_{II,1} = (x_3, \cdots, x_{\lfloor\frac{D}{2}\rfloor + 1})$ and $\mathbf{x}_{II,2} = (x_{\lfloor\frac{D}{2}\rfloor + 2}, \cdots, x_D)$, $c = \cot(3\pi t^2), \text{when~} t^2 \neq \frac{n}{3}, n \in \mathbb{Z}, c = 10^{-32}, \text{otherwise}$,
$h_1(\mathbf{x}_I, t) = \frac{1}{\pi}\abs{\arctan(c)}$ and $h_2(\mathbf{x}_I, t) = G(t) + x_1^{H(t)}$,
$\mathbf{R}_{II,1}(t)$ and $\mathbf{R}_{II,2}(t)$ are symmetric positive semidefinite matrices in the $t$-th environment,
the search space is $[0,1]^2 \times [0,1]^{\lfloor\frac{D}{2}\rfloor -1} \times  [-1, 2]^{\lceil\frac{D}{2}\rceil-1}$.

The PS and PF at time t can be described as:
\begin{equation*}
    \begin{aligned}
         & \text{PS(t): }0 \leq x_{1,2} \leq 1, x_i = h_1(\mathbf{x}_I, t) \in \mathbf{x}_{II,1}, x_j = h_2(\mathbf{x}_I, t) \in \mathbf{x}_{II,2} \\
         & \text{PF(t): }f_2 = 2.8 - f_1^{H(t)}, 0 \leq f_1 \leq 1.5
    \end{aligned}
\end{equation*}

\begin{figure}[htbp]
    \centering
    \begin{minipage}[t]{0.45\textwidth}
        \centering
        \includegraphics[width=\linewidth]{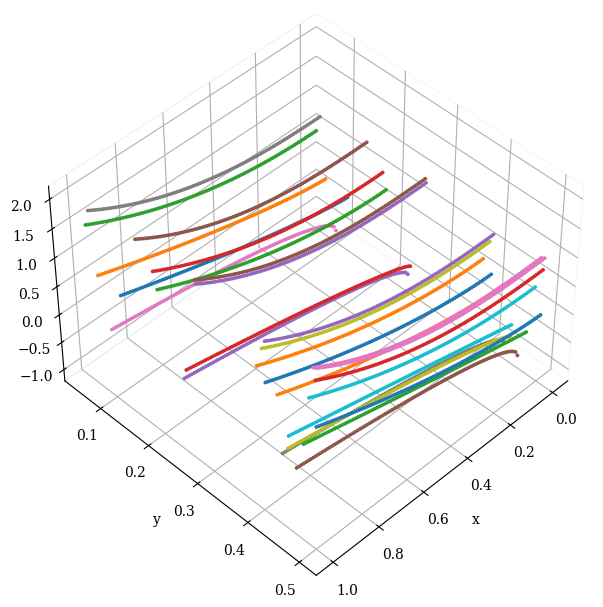}
        \caption{PS of GTS2 problem}%
        \label{fig:gts_supp_GTS2_ps}
    \end{minipage}
    \hfill
    \begin{minipage}[t]{0.45\textwidth}
        \centering
        \includegraphics[width=\linewidth]{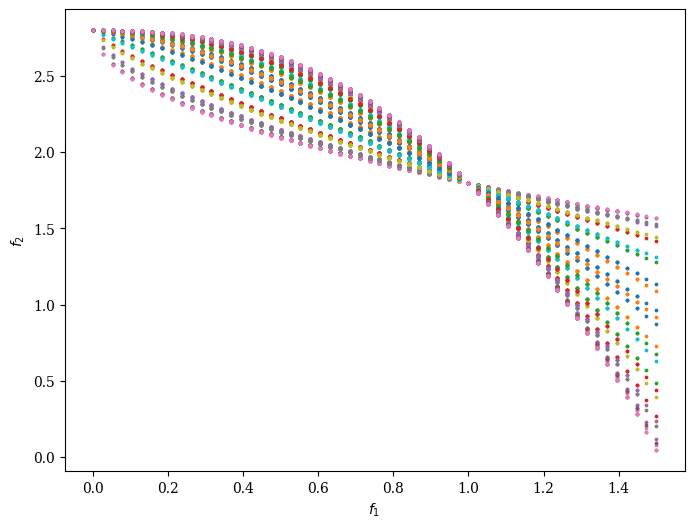}
        \caption{PF of GTS2 problem}%
        \label{fig:gts_supp_GTS2_pf}
    \end{minipage}
\end{figure}

\subsection{GTS3}

\begin{equation}
    \text{min}
    \begin{cases}
        f_1(\mathbf{x},t) = g(\mathbf{x},t)(x_1 + 0.1\sin(3\pi x_1))^{\beta_t} \\
        f_2(\mathbf{x},t) = g(\mathbf{x},t)(1 - x_1 + 0.1\sin(3\pi x_1))^{\beta_t}
    \end{cases}
\end{equation}
with
\begin{equation*}
    \begin{split}
        g(\mathbf{x},t) = 1
         & + \Bigl(\bigl(\mathbf{x}_{II,1} - h_1(\mathbf{x}_I)\bigr)^T \mathbf{R}_{II,1}(t) \bigl(\mathbf{x}_{II,1} - h_1(\mathbf{x}_I)\bigr)\Bigr)^{\frac{1}{p}} \\
         & + \Bigl(\bigl(\mathbf{x}_{II,2} - h_2(\mathbf{x}_I)\bigr)^T \mathbf{R}_{II,2}(t) \bigl(\mathbf{x}_{II,2} - h_2(\mathbf{x}_I)\bigr)\Bigr)^{\frac{1}{p}}
    \end{split}
\end{equation*}
where $p \geq 1$, $\mathbf{x}_I = (x_1)$, $\mathbf{x}_{II,1} = (x_2, \cdots, x_{\lfloor\frac{D}{2}\rfloor})$ and $\mathbf{x}_{II,2} = (x_{\lfloor\frac{D}{2}\rfloor + 1}, \cdots, x_D)$,
$h_1(\mathbf{x}_I, t) = \frac{G(t)\sin(4\pi x_1)}{1 + \abs{G(t)}}$ and $h_2(\mathbf{x}_I, t) = G(t) + x_1^{H(t)}$,
$\mathbf{R}_{II,1}(t)$ and $\mathbf{R}_{II,2}(t)$ are symmetric positive semidefinite matrices in the $t$-th environment,
the search space is $[0,1] \times [-1,1]^{\lfloor\frac{D}{2}\rfloor - 1} \times [-1, 2]^{\lceil\frac{D}{2}\rceil}$.

The PS and PF at time t can be described as:
\begin{equation*}
    \begin{aligned}
         & \text{PS(t): }0 \leq x_1 \leq 1, x_i = h_1(\mathbf{x}_I, t) \in \mathbf{x}_{II,1}, x_j = h_2(\mathbf{x}_I, t) \in \mathbf{x}_{II,2}                              \\
         & \text{PF(t): }f_1^{\frac{1}{\beta_t}} + f_2^{\frac{1}{\beta_t}} = 1 + 0.2\sin(3\pi \frac{f_1^{\frac{1}{\beta_t}} -f_2^{\frac{1}{\beta_t}} + 1}{2}), 0 \leq f_1 \leq 1
    \end{aligned}
\end{equation*}

\begin{figure}[htbp]
    \centering
    \begin{minipage}[t]{0.45\textwidth}
        \centering
        \includegraphics[width=\linewidth]{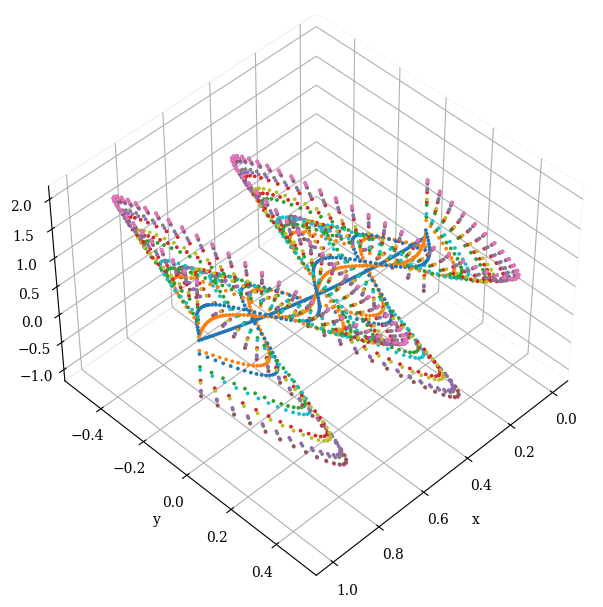}
        \caption{PS of GTS3 problem}%
        \label{fig:gts_supp_GTS3_ps}
    \end{minipage}
    \hfill
    \begin{minipage}[t]{0.45\textwidth}
        \centering
        \includegraphics[width=\linewidth]{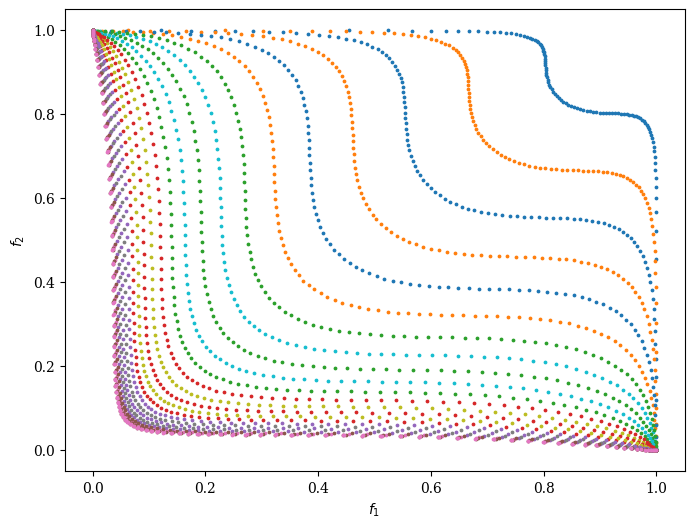}
        \caption{PF of GTS3 problem}%
        \label{fig:gts_supp_GTS3_pf}
    \end{minipage}
\end{figure}

\subsection{GTS4}

\begin{equation}
    \text{min}
    \begin{cases}
        f_1(\mathbf{x},t) = g(\mathbf{x},t)\frac{1 + t}{x_1 + 3} \\
        f_2(\mathbf{x},t) = g(\mathbf{x},t)\frac{x_1 + 3}{1 + t}
    \end{cases}
\end{equation}
with
\begin{equation*}
    \begin{split}
        g(\mathbf{x},t) = 1
         & + \Bigl(\bigl(\mathbf{x}_{II,1} - h_1(\mathbf{x}_I)\bigr)^T \mathbf{R}_{II,1}(t) \bigl(\mathbf{x}_{II,1} - h_1(\mathbf{x}_I)\bigr)\Bigr)^{\frac{1}{p}} \\
         & + \Bigl(\bigl(\mathbf{x}_{II,2} - h_2(\mathbf{x}_I)\bigr)^T \mathbf{R}_{II,2}(t) \bigl(\mathbf{x}_{II,2} - h_2(\mathbf{x}_I)\bigr)\Bigr)^{\frac{1}{p}} \\
         & - 0.5 + 0.25\sin(0.3\pi t)
    \end{split}
\end{equation*}
where $p \geq 1$, $\mathbf{x}_I = (x_1)$, $\mathbf{x}_{II,1} = (x_2, \cdots, x_{\lfloor\frac{D}{2}\rfloor})$ and $\mathbf{x}_{II,2} = (x_{\lfloor\frac{D}{2}\rfloor + 1}, \cdots, x_D)$,
$h_1(\mathbf{x}_I, t) = \abs{G(t)}$ and $h_2(\mathbf{x}_I, t) = \frac{G(t)\sin(4\pi x_1)}{1 + \abs{G(t)}}$,
$\mathbf{R}_{II,1}(t)$ and $\mathbf{R}_{II,2}(t)$ are symmetric positive semidefinite matrices in the $t$-th environment,
the search space is $[0,1] \times [0,1]^{\lfloor\frac{D}{2}\rfloor -1} \times [-1, 1]^{\lceil\frac{D}{2}\rceil}$.

The PS and PF at time t can be described as:
\begin{equation*}
    \begin{aligned}
         & \text{PS(t): }0 \leq x_1 \leq 1, x_i = h_1(\mathbf{x}_I, t) \in \mathbf{x}_{II,1}, x_j = h_2(\mathbf{x}_I, t) \in \mathbf{x}_{II,2} \\
         & \text{PF(t): }f_2 = \frac{1}{f_1}, \frac{1+t}{16} \leq f_1 \leq \frac{1+t}{4}
    \end{aligned}
\end{equation*}

\subsection{GTS5}

\begin{equation}
    \text{min}
    \begin{cases}
        f_1(\mathbf{x},t) = g(\mathbf{x},t)((0.5x_1+x_2) + 0.02\sin(\omega_t\pi (0.5x_1+x_2))) \\
        f_2(\mathbf{x},t) = g(\mathbf{x},t)(1.6 - (0.5x_1+x_2) + 0.02\sin(\omega_t\pi (0.5x_1+x_2)))
    \end{cases}
\end{equation}
with
\begin{equation*}
    \begin{split}
        g(\mathbf{x},t) = 1
         & + \Bigl(\bigl(\mathbf{x}_{II,1} - h_1(\mathbf{x}_I)\bigr)^T \mathbf{R}_{II,1}(t) \bigl(\mathbf{x}_{II,1} - h_1(\mathbf{x}_I)\bigr)\Bigr)^{\frac{1}{p}} \\
         & + \Bigl(\bigl(\mathbf{x}_{II,2} - h_2(\mathbf{x}_I)\bigr)^T \mathbf{R}_{II,2}(t) \bigl(\mathbf{x}_{II,2} - h_2(\mathbf{x}_I)\bigr)\Bigr)^{\frac{1}{p}} \\
         & + 0.5 + 0.5G(t)
    \end{split}
\end{equation*}
where $p \geq 1$, $\mathbf{x}_I = (x_1, x_2)$, $\mathbf{x}_{II,1} = (x_3, \cdots, x_{\lfloor\frac{D}{2}\rfloor + 1})$ and $\mathbf{x}_{II,2} = (x_{\lfloor\frac{D}{2}\rfloor + 2}, \cdots, x_D)$,
$h_1(\mathbf{x}_I, t) = \cos(0.5\pi t)$ and $h_2(\mathbf{x}_I, t) = G(t) + x_1^{H(t)}$,
$\mathbf{R}_{II,1}(t)$ and $\mathbf{R}_{II,2}(t)$ are symmetric positive semidefinite matrices in the $t$-th environment,
the search space is $[0,1]^2 \times [-1,1]^{\lfloor\frac{D}{2}\rfloor - 1} \times [-1, 2]^{\lceil\frac{D}{2}\rceil-1}$.

The PS and PF at time t can be described as:
\begin{equation*}
    \begin{aligned}
         & \text{PS(t): }0 \leq x_{1,2} \leq 1, x_i = h_1(\mathbf{x}_I, t) \in \mathbf{x}_{II,1}, x_j = h_2(\mathbf{x}_I, t) \in \mathbf{x}_{II,2} \\
         & \text{PF(t): }(f_1 + f_2) = (1.6 + 0.5G(t))(1 + 0.04\sin(\omega_t\pi\frac{\frac{1}{1.6 + 0.5G(t)}(f_1 - f_2) + 1.6}{2})), 0 \leq f_1 \leq 3
    \end{aligned}
\end{equation*}

\begin{figure}[htbp]
    \centering
    \begin{minipage}[t]{0.45\textwidth}
        \centering
        \includegraphics[width=\linewidth]{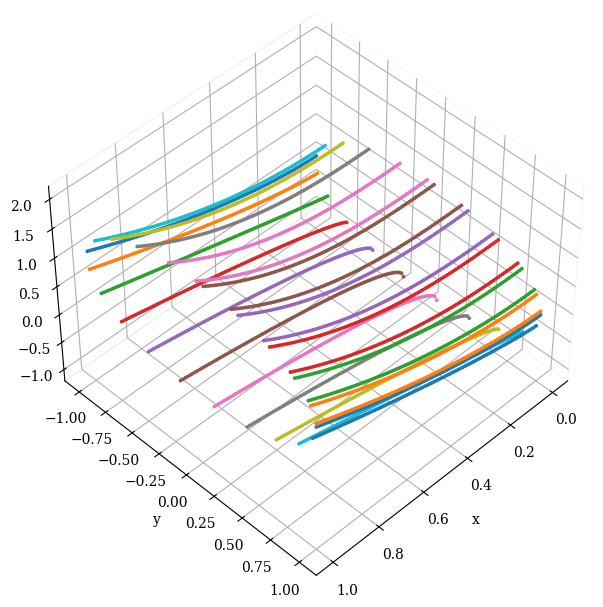}
        \caption{PS of GTS5 problem}%
        \label{fig:gts_supp_GTS5_ps}
    \end{minipage}
    \hfill
    \begin{minipage}[t]{0.45\textwidth}
        \centering
        \includegraphics[width=\linewidth]{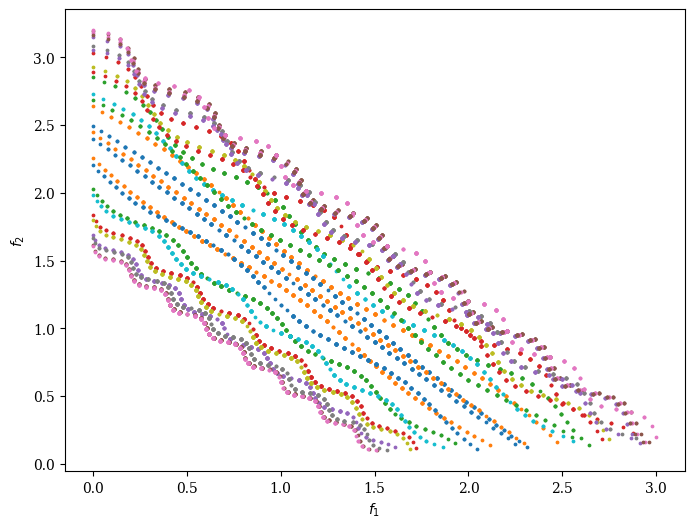}
        \caption{PF of GTS5 problem}%
        \label{fig:gts_supp_GTS5_pf}
    \end{minipage}
\end{figure}

\subsection{GTS6}

\begin{equation}
    \text{min}
    \begin{cases}
        f_1(\mathbf{x},t) = x_1 \\
        f_2(\mathbf{x},t) = g(\mathbf{x},t)(1 - (\frac{x_1}{g(\mathbf{x},t)})^{H(t)})
    \end{cases}
\end{equation}
with
\begin{equation*}
    \begin{split}
        g(\mathbf{x},t) = 1
         & + \Bigl(\bigl(\mathbf{x}_{II,1} - \phi(t)h_1(\mathbf{x}_I)\bigr)^T \mathbf{R}_{II,1}(t) \bigl(\mathbf{x}_{II,1} - \phi(t)h_1(\mathbf{x}_I)\bigr)\Bigr)^{\frac{1}{p}} \\
         & + \Bigl(\bigl(\mathbf{x}_{II,2} - \phi(t)h_2(\mathbf{x}_I)\bigr)^T \mathbf{R}_{II,2}(t) \bigl(\mathbf{x}_{II,2} - \phi(t)h_2(\mathbf{x}_I)\bigr)\Bigr)^{\frac{1}{p}}
    \end{split}
\end{equation*}
where $p \geq 1$, $\mathbf{x}_I = (x_1)$, $\mathbf{x}_{II,1} = (x_2, \cdots, x_{\lfloor\frac{D}{2}\rfloor})$ and $\mathbf{x}_{II,2} = (x_{\lfloor\frac{D}{2}\rfloor + 1}, \cdots, x_D)$,
$h_1(\mathbf{x}_I, t) = \cos(0.5\pi t)$ and $h_2(\mathbf{x}_I, t) = G(t) + x_1^{H(t)}$,
$\mathbf{R}_{II,1}(t)$ and $\mathbf{R}_{II,2}(t)$ are symmetric positive semidefinite matrices in the $t$-th environment,
the search space is $[0,1] \times [-1,1]^{\lfloor\frac{D}{2}\rfloor -1} \times [-1, 2]^{\lceil\frac{D}{2}\rceil}$.

\begin{equation}\label{eq:gts_supp_time_linkage}
    \phi(t) =
    \begin{cases}
        1,                                                                                                  & \text{if } t=1,   \\
        1 + \left\Vert f(\mathbf{x}_{\text{knee}},t-1) - f(\hat{\mathbf{x}}_{\text{knee}},t-1) \right\Vert, & \text{otherwise,}
    \end{cases}
\end{equation}
where $f(\mathbf{x}_{knee},t-1)$ and $f(\hat{\mathbf{x}_{knee}},t-1)$ are the objective values of the knee point of the true PF and estimated PF at time $t-1$, respectively.

The PS and PF at time t can be described as:
\begin{equation*}
    \begin{aligned}
         & \text{PS(t): }0 \leq x_1 \leq 1, x_i = h_1(\mathbf{x}_I, t) \in \mathbf{x}_{II,1}, x_j = h_2(\mathbf{x}_I, t) \in \mathbf{x}_{II,2} \\
         & \text{PF(t): }f_2 = 1 - f_1^{H(t)}, 0 \leq f_1 \leq 1
    \end{aligned}
\end{equation*}

\begin{figure}[htbp]
    \centering
    \begin{minipage}[t]{0.45\textwidth}
        \centering
        \includegraphics[width=\linewidth]{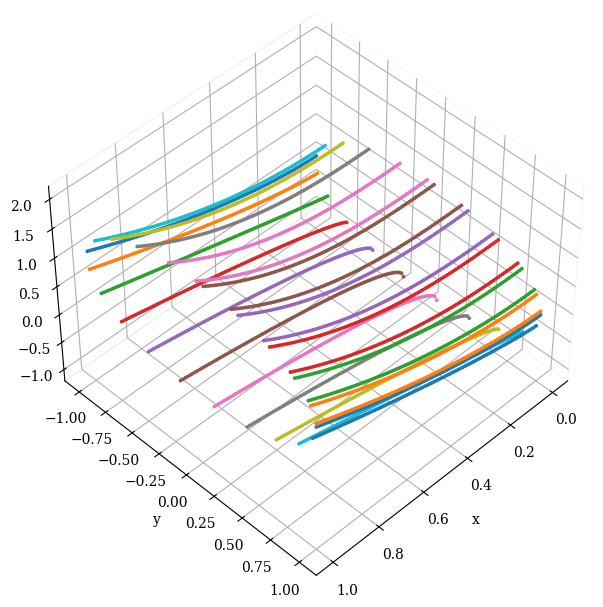}
        \caption{PS of GTS6 problem}%
        \label{fig:gts_supp_GTS6_ps}
    \end{minipage}
    \hfill
    \begin{minipage}[t]{0.45\textwidth}
        \centering
        \includegraphics[width=\linewidth]{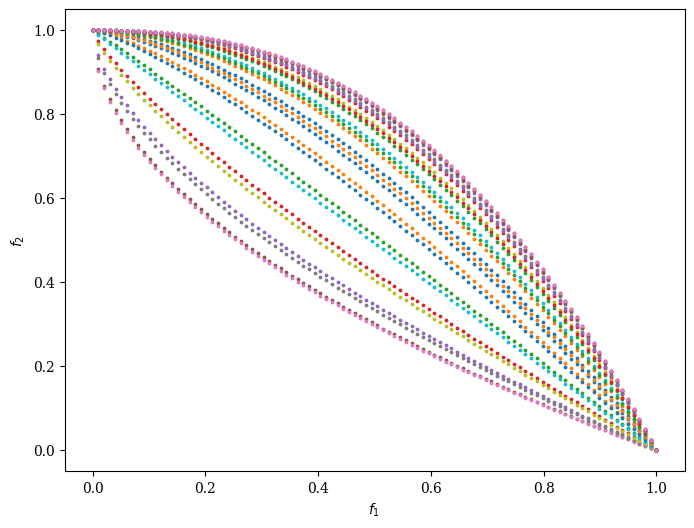}
        \caption{PF of GTS6 problem}%
        \label{fig:gts_supp_GTS6_pf}
    \end{minipage}
\end{figure}

\subsection{GTS7}

\begin{equation}
    \text{min}
    \begin{cases}
        f_1(\mathbf{x},t) =  g(\mathbf{x},t)\abs{x_1-a_t}^{H(t)} \\
        f_2(\mathbf{x},t) = g(\mathbf{x},t)\abs{x_1-a_t-b_t}^{H(t)}
    \end{cases}
\end{equation}
with
\begin{equation*}
    \begin{split}
        g(\mathbf{x},t) = 1
         & + \Bigl(\bigl(\mathbf{x}_{II,1} - \phi(t)h_1(\mathbf{x}_I)\bigr)^T \mathbf{R}_{II,1}(t) \bigl(\mathbf{x}_{II,1} - \phi(t)h_1(\mathbf{x}_I)\bigr)\Bigr)^{\frac{1}{p}} \\
         & + \Bigl(\bigl(\mathbf{x}_{II,2} - \phi(t)h_2(\mathbf{x}_I)\bigr)^T \mathbf{R}_{II,2}(t) \bigl(\mathbf{x}_{II,2} - \phi(t)h_2(\mathbf{x}_I)\bigr)\Bigr)^{\frac{1}{p}}
    \end{split}
\end{equation*}
where $p \geq 1$, $\mathbf{x}_I = (x_1)$, $\mathbf{x}_{II,1} = (x_2, \cdots, x_{\lfloor\frac{D}{2}\rfloor})$ and $\mathbf{x}_{II,2} = (x_{\lfloor\frac{D}{2}\rfloor + 1}, \cdots, x_D)$,
$h_1(\mathbf{x}_I, t) = \cos(0.5\pi t)$ and $h_2(\mathbf{x}_I, t) = \frac{1}{1 + e^{\alpha_t(x_1 - 0.5)}}$,
$\mathbf{R}_{II,1}(t)$ and $\mathbf{R}_{II,2}(t)$ are symmetric positive semidefinite matrices in the $t$-th environment,
the search space is $[-1,2.5] \times [-1,1]^{\lfloor\frac{D}{2}\rfloor - 1} \times [0, 1]^{\lceil\frac{D}{2}\rceil}$. $\phi(t)$ is the same as defined in~\eqref{eq:gts_time_linkage}.

The PS and PF at time t can be described as:
\begin{equation*}
    \begin{aligned}
         & \text{PS(t): }a_t \leq x_1 \leq (a_t + b_t), x_i = h_1(\mathbf{x}_I, t) \in \mathbf{x}_{II,1}, x_j = h_2(\mathbf{x}_I, t) \in \mathbf{x}_{II,2} \\
         & \text{PF(t): }f_2 = (b_t - f_1^{\frac{1}{H(t)}})^{H(t)}, 0 \leq f_1 \leq 3.5
    \end{aligned}
\end{equation*}

\begin{figure}[htbp]
    \centering
    \begin{minipage}[t]{0.45\textwidth}
        \centering
        \includegraphics[width=\linewidth]{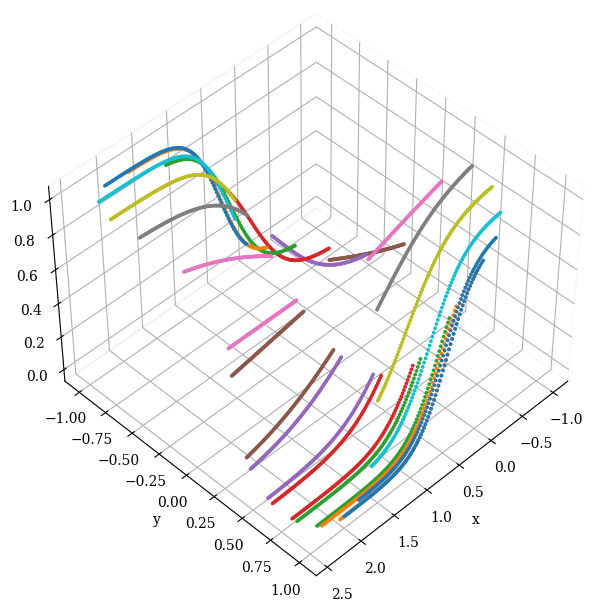}
        \caption{PS of GTS7 problem}%
        \label{fig:gts_supp_GTS7_ps}
    \end{minipage}
    \hfill
    \begin{minipage}[t]{0.45\textwidth}
        \centering
        \includegraphics[width=\linewidth]{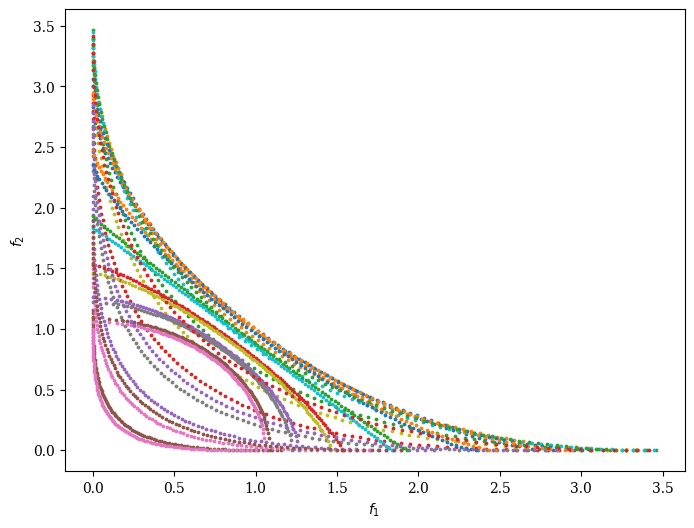}
        \caption{PF of GTS7 problem}%
        \label{fig:gts_supp_GTS7_pf}
    \end{minipage}
\end{figure}

\subsection{GTS8}

\begin{equation}
    \text{min}
    \begin{cases}
        f_1(\mathbf{x},t) = (0.5x_1+x_2)                                                         \\
        f_2(\mathbf{x},t) = g(\mathbf{x},t)(2.8 - (\frac{(0.5x_1+x_2)}{g(\mathbf{x},t)})^{H(t)}) \\
    \end{cases}
\end{equation}
with
\begin{equation*}
    \begin{split}
        g(\mathbf{x},t) = 1
         & + \Bigl(\bigl(\mathbf{x}_{II,1} - \phi(t)h_1(\mathbf{x}_I)\bigr)^T \mathbf{R}_{II,1}(t) \bigl(\mathbf{x}_{II,1} - \phi(t)h_1(\mathbf{x}_I)\bigr)\Bigr)^{\frac{1}{p}} \\
         & + \Bigl(\bigl(\mathbf{x}_{II,2} - \phi(t)h_2(\mathbf{x}_I)\bigr)^T \mathbf{R}_{II,2}(t) \bigl(\mathbf{x}_{II,2} - \phi(t)h_2(\mathbf{x}_I)\bigr)\Bigr)^{\frac{1}{p}} \\
         & + 0.25\abs{\cos(0.3 \pi t)}
    \end{split}
\end{equation*}
where $p \geq 1$, $\mathbf{x}_I = (x_1, x_2)$, $\mathbf{x}_{II,1} = (x_3, \cdots, x_{\lfloor\frac{D}{2}\rfloor + 1})$ and $\mathbf{x}_{II,2} = (x_{\lfloor\frac{D}{2}\rfloor + 2}, \cdots, x_D)$,
$h_1(\mathbf{x}_I, t) = \frac{1}{1 + e^{\alpha _t(x_1 - 0.5)}}$ and $h_2(\mathbf{x}_I, t) = G(t) + x_1^{H(t)}$,
$\mathbf{R}_{II,1}(t)$ and $\mathbf{R}_{II,2}(t)$ are symmetric positive semidefinite matrices in the $t$-th environment,
the search space is $[0,1]^2 \times [0,1]^{\lfloor\frac{D}{2}\rfloor -1} \times [-1, 2]^{\lceil\frac{D}{2}\rceil-1}$. $\phi(t)$ is the same as defined in~\eqref{eq:gts_time_linkage}.

The PS and PF at time t can be described as:
\begin{equation*}
    \begin{aligned}
         & \text{PS(t): }0 \leq x_{1,2} \leq 1, x_i = h_1(\mathbf{x}_I, t) \in \mathbf{x}_{II,1}, x_j = h_2(\mathbf{x}_I, t) \in \mathbf{x}_{II,2} \\
         & \text{PF(t): }\frac{f_2}{1 + 0.25\abs{\cos(0.3\pi t)}} = 2.8 - (\frac{f_1}{1 + 0.25\abs{\cos(0.3\pi t)}})^{H(t)}, 0 \leq f_1 \leq 1.5
    \end{aligned}
\end{equation*}

\begin{figure}[htbp]
    \centering
    \begin{minipage}[t]{0.45\textwidth}
        \centering
        \includegraphics[width=\linewidth]{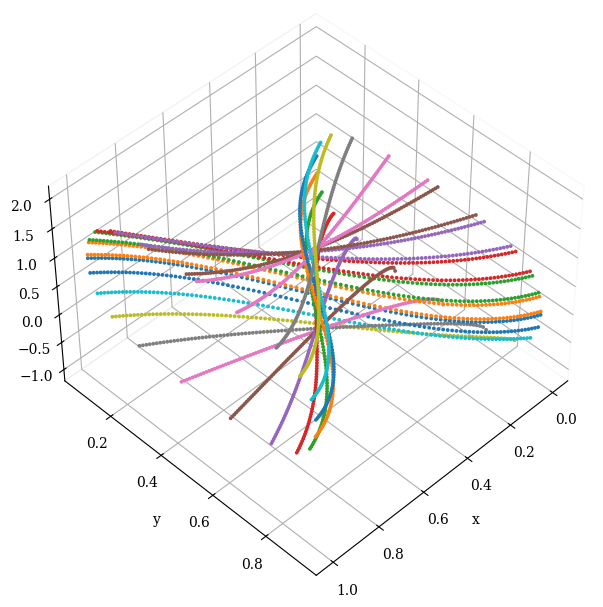}
        \caption{PS of GTS8 problem}%
        \label{fig:gts_supp_GTS8_ps}
    \end{minipage}
    \hfill
    \begin{minipage}[t]{0.45\textwidth}
        \centering
        \includegraphics[width=\linewidth]{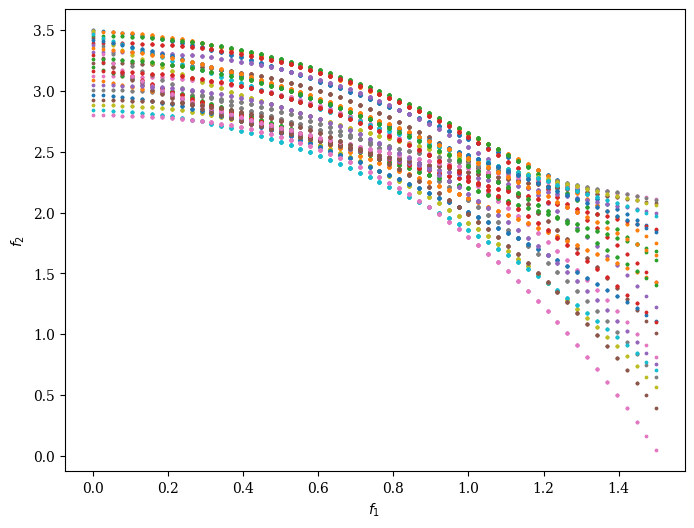}
        \caption{PF of GTS8 problem}%
        \label{fig:gts_supp_GTS8_pf}
    \end{minipage}
\end{figure}

\subsection{GTS9}

\begin{equation}
    \text{min}
    \begin{cases}
        f_1(\mathbf{x},t) = g(\mathbf{x},t)\cos(0.5\pi x_1)\cos(0.5\pi x_2) \\
        f_2(\mathbf{x},t) = g(\mathbf{x},t)\cos(0.5\pi x_1)\sin(0.5\pi x_2) \\
        f_3(\mathbf{x},t) = g(\mathbf{x},t)\sin(0.5\pi x_1)
    \end{cases}
\end{equation}
with
\begin{equation*}
    \begin{split}
        g(\mathbf{x},t) = 1
         & + \Bigl(\bigl(\mathbf{x}_{II,1} - h_1(\mathbf{x}_I)\bigr)^T \mathbf{R}_{II,1}(t) \bigl(\mathbf{x}_{II,1} - h_1(\mathbf{x}_I)\bigr)\Bigr)^{\frac{1}{p}} \\
         & + \Bigl(\bigl(\mathbf{x}_{II,2} - h_2(\mathbf{x}_I)\bigr)^T \mathbf{R}_{II,2}(t) \bigl(\mathbf{x}_{II,2} - h_2(\mathbf{x}_I)\bigr)\Bigr)^{\frac{1}{p}} \\
         & + \abs{\cos(0.27\pi t)}
    \end{split}
\end{equation*}
where $p \geq 1$, $\mathbf{x}_I = (x_1, x_2)$, $\mathbf{x}_{II,1} = (x_3, \cdots, x_{\lfloor\frac{D}{2}\rfloor + 1})$ and $\mathbf{x}_{II,2} = (x_{\lfloor\frac{D}{2}\rfloor + 2}, \cdots, x_D)$,
$h_1(\mathbf{x}_I, t) = \frac{1}{1+e^{\alpha_t(x_1 - 0.5)}}$ and $h_2(\mathbf{x}_I, t) = \sin(tx_1)$,
$\mathbf{R}_{II,1}(t)$ and $\mathbf{R}_{II,2}(t)$ are symmetric positive semidefinite matrices in the $t$-th environment,
the search space is $[0,1]^2 \times [0,1]^{\lfloor\frac{D}{2}\rfloor - 1} \times [-1, 1]^{\lceil\frac{D}{2}\rceil - 1}$.

The PS and PF at time t can be described as:
\begin{equation*}
    \begin{aligned}
         & \text{PS(t): }0 \leq x_{1,2} \leq 1, x_i = h_1(\mathbf{x}_I, t) \in \mathbf{x}_{II,1}, x_j = h_2(\mathbf{x}_I, t) \in \mathbf{x}_{II,2} \\
         & \text{PF(t): }\sum_{i = 1}^{3}f_i^2 = 1 + \abs{\cos(0.27\pi t)}, 0 \leq f_{i = 1,2,3} \leq 2
    \end{aligned}
\end{equation*}

\begin{figure}[htbp]
    \centering
    \begin{minipage}[t]{0.45\textwidth}
        \centering
        \includegraphics[width=\linewidth]{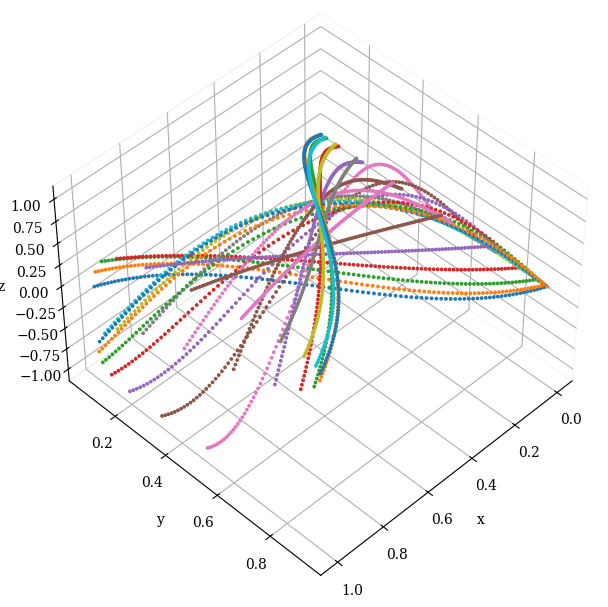}
        \caption{PS of GTS9 problem}%
        \label{fig:gts_supp_GTS9_ps}
    \end{minipage}
    \hfill
    \begin{minipage}[t]{0.45\textwidth}
        \centering
        \includegraphics[width=\linewidth]{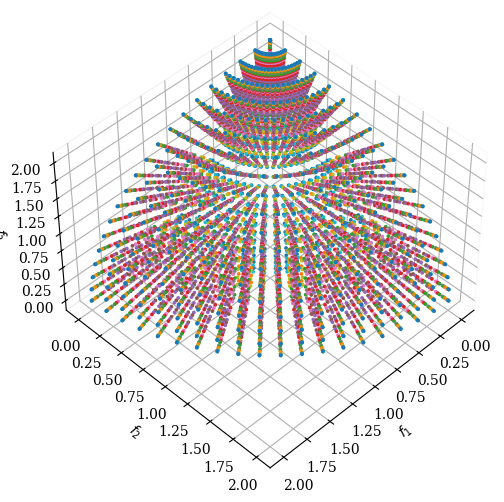}
        \caption{PF of GTS9 problem}%
        \label{fig:gts_supp_GTS9_pf}
    \end{minipage}
\end{figure}

\subsection{GTS10}

\begin{equation}
    \text{min}
    \begin{cases}
        f_1(\mathbf{x},t) =  g(\mathbf{x},t)\cos^2(0.5\pi x_1) \\
        f_2(\mathbf{x},t) = g(\mathbf{x},t)\cos^2(0.5\pi x_2)  \\
        f_3(\mathbf{x},t) = g(\mathbf{x},t)\sum_{j = 1}^{2}(\sin^2(0.5\pi x_j) + \sin(0.5\pi x_j)\cos^2(\lfloor6G(t)\rfloor \pi x_j))
    \end{cases}
\end{equation}
with
\begin{equation*}
    \begin{split}
        g(\mathbf{x},t) = 1
         & + \Bigl(\bigl(\mathbf{x}_{II,1} - h_1(\mathbf{x}_I)\bigr)^T \mathbf{R}_{II,1}(t) \bigl(\mathbf{x}_{II,1} - h_1(\mathbf{x}_I)\bigr)\Bigr)^{\frac{1}{p}} \\
         & + \Bigl(\bigl(\mathbf{x}_{II,2} - h_2(\mathbf{x}_I)\bigr)^T \mathbf{R}_{II,2}(t) \bigl(\mathbf{x}_{II,2} - h_2(\mathbf{x}_I)\bigr)\Bigr)^{\frac{1}{p}}
    \end{split}
\end{equation*}
where $p \geq 1$, $\mathbf{x}_I = (x_1, x_2)$, $\mathbf{x}_{II,1} = (x_3, \cdots, x_{\lfloor\frac{D}{2}\rfloor + 1})$ and $\mathbf{x}_{II,2} = (x_{\lfloor\frac{D}{2}\rfloor + 2}, \cdots, x_D)$,
$h_1(\mathbf{x}_I, t) = \abs{G(t)}$ and $h_2(\mathbf{x}_I, t) = -0.5 + \frac{\abs{G(t)\sin(4\pi x_1)}}{0.5(1+\abs{G(t)})}$,
$\mathbf{R}_{II,1}(t)$ and $\mathbf{R}_{II,2}(t)$ are symmetric positive semidefinite matrices in the $t$-th environment,%
the search space is $[0,1]^2 \times [0,1]^{\lfloor\frac{D}{2}\rfloor - 1} \times [-1, 1]^{\lceil\frac{D}{2}\rceil - 1}$.

The PS at time t can be described as:
\begin{equation*}
    \begin{aligned}
         & \text{PS(t): }0 \leq x_{1,2} \leq 1, x_i = h_1(\mathbf{x}_I, t) \in \mathbf{x}_{II,1}, x_j = h_2(\mathbf{x}_I, t) \in \mathbf{x}_{II,2}
    \end{aligned}
\end{equation*}

\subsection{GTS11}

\begin{equation}
    \text{min}
    \begin{cases}
        f_1(\mathbf{x},t) = g(\mathbf{x},t)(1.05 - y + 0.05\sin(6\pi y))                            \\
        f_2(\mathbf{x},t) = g(\mathbf{x},t)(1.05 - x_2 + 0.05\sin(6\pi x_2))(y +  0.05\sin(6\pi y)) \\
        f_3(\mathbf{x},t) = g(\mathbf{x},t)(x_2 + 0.05\sin(6\pi x_2))(y + 0.05\sin(6\pi y))
    \end{cases}
\end{equation}
with
\begin{equation*}
    \begin{split}
        g(\mathbf{x},t) = 1
         & + \Bigl(\bigl(\mathbf{x}_{II,1} - h_1(\mathbf{x}_I)\bigr)^T \mathbf{R}_{II,1}(t) \bigl(\mathbf{x}_{II,1} - h_1(\mathbf{x}_I)\bigr)\Bigr)^{\frac{1}{p}} \\
         & + \Bigl(\bigl(\mathbf{x}_{II,2} - h_2(\mathbf{x}_I)\bigr)^T \mathbf{R}_{II,2}(t) \bigl(\mathbf{x}_{II,2} - h_2(\mathbf{x}_I)\bigr)\Bigr)^{\frac{1}{p}}
    \end{split}
\end{equation*}
where $p \geq 1$, $\mathbf{x}_I = (x_1, x_2)$, $\mathbf{x}_{II,1} = (x_3, \cdots, x_{\lfloor\frac{D}{2}\rfloor + 1})$ and $\mathbf{x}_{II,2} = (x_{\lfloor\frac{D}{2}\rfloor + 2}, \cdots, x_D)$,
$h_1(\mathbf{x}_I, t) = \abs{G(t)}$ and $h_2(\mathbf{x}_I, t) = G(t) + x_1^{H(t)}$,
$\mathbf{R}_{II,1}(t)$ and $\mathbf{R}_{II,2}(t)$ are symmetric positive semidefinite matrices in the $t$-th environment,
the search space is $[0,1]^2 \times [0,1]^{\lfloor\frac{D}{2}\rfloor - 1} \times [-1, 2]^{\lceil\frac{D}{2}\rceil - 1}$.

The PS at time t can be described as:
\begin{equation*}
    \begin{aligned}
         & \text{PS(t): }0 \leq x_{1,2} \leq 1, x_i = h_1(\mathbf{x}_I, t) \in \mathbf{x}_{II,1}, x_j = h_2(\mathbf{x}_I, t) \in \mathbf{x}_{II,2}
    \end{aligned}
\end{equation*}

\begin{figure}[htbp]
    \centering
    \begin{minipage}[t]{0.45\textwidth}
        \centering
        \includegraphics[width=\linewidth]{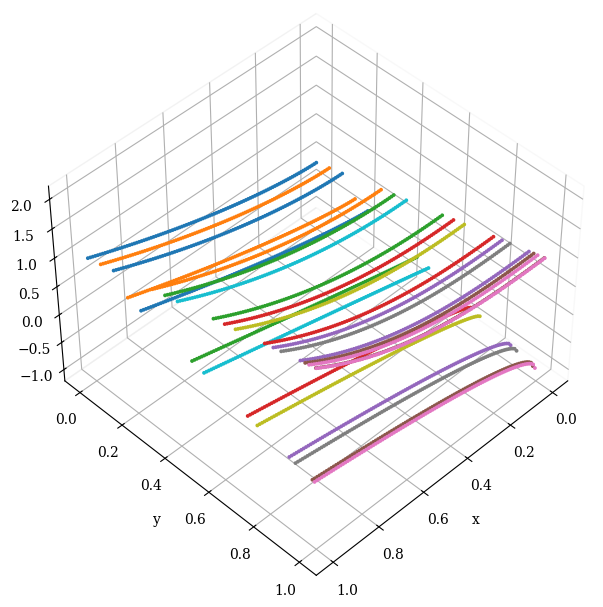}
        \caption{PS of GTS11 problem}%
        \label{fig:gts_supp_GTS11_ps}
    \end{minipage}
    \hfill
    \begin{minipage}[t]{0.45\textwidth}
        \centering
        \includegraphics[width=\linewidth]{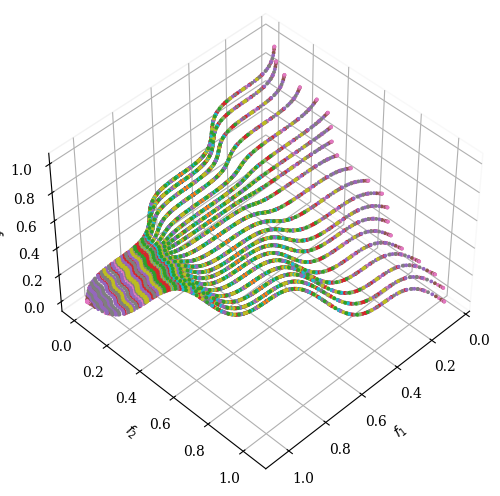}
        \caption{PF of GTS11 problem}%
        \label{fig:gts_supp_GTS11_pf}
    \end{minipage}
\end{figure}
\newpage

\bibliographystyle{elsarticle-num}
\bibliography{bibs/abbr}

\end{document}